\documentclass{article} 
\usepackage{template/arxiv,times}

\preprintcopy 

\usepackage{template/packages}
\usepackage{template/macros}

\title{\centering Probing Transformer Representations \\ Layer by Layer, Module by Module}

\def\authspace{1.5}
\def\authbspace{1}
\author{\\
  \makebox[\textwidth][c]{ 
    \begin{tabular}{cc}
      \textbf{Ambroise Odonnat}$^*$ \hspace{\authspace cm} & \textbf{Th\'eo Gnassounou} \\
      Noah's Ark Lab, Inria \hspace{\authspace cm} & EPFL
      \\[1ex] 
    \end{tabular}
  }
  \\ \\
  \makebox[\textwidth][c]{
    \begin{tabular}{ccc}
      \textbf{Laetitia Chapel} \hspace{\authbspace cm} & \textbf{Romain Tavenard} \hspace{\authbspace cm} & \textbf{Ievgen Redko} \\
      L'Institut Agro, IRISA \hspace{\authbspace cm} & Univ. Rennes 2, IRISA \hspace{\authbspace cm} & Schneider Electric$^\dagger$
    \end{tabular}
  }
}

\hypersetup{
  pdftitle={Probing Transformer Representations Layer by Layer, Module by Module},
  pdfauthor={Ambroise Odonnat, Théo Gnassounou, Laetitia Chapel, Romain Tavenard, Ievgen Redko}
}

\begin{document}
\addtocontents{toc}{\protect\setcounter{tocdepth}{0}}

\maketitle
\begingroup
\renewcommand{\thefootnote}{}
\footnotetext{$^{*}$Correspondence to \href{mailto:ambroise.odonnat@gmail.com}{\texttt{ambroise.odonnat@gmail.com}}. \quad $^{\dagger}$Work done while at Noah's Ark Lab.}
\endgroup

\begin{abstract}
Which representation should be probed in a transformer-based foundation model? Recent work shows that the standard practice of probing the final layer is often suboptimal, as intermediate layers can provide better features. Yet, representations are still largely studied along depth alone, treating each transformer block as an atomic unit. In this work, we analyze transformer representations \emph{layer by layer, module by module}. Through more than $144\,000$ linear probing evaluations of vision foundation models (VFMs) and large language models (LLMs), we show that layer and module choices should be considered jointly: the best layer-module pair yields substantially larger gains over the final-layer residual stream than the best layer with the residual stream fixed. Moreover, module performance can vary substantially across layers, showing that useful representations cannot be characterized by depth alone. Building on these findings, we evaluate simple fixed probing strategies: the final-layer feedforward activation ($\vfm$) for vision and the second normalization layer at $60\%$ depth ($\llm$) for language. These strategies improve over the probing baseline in a majority of evaluations and recover, on average, $42\%$ and $86\%$ of the gains from probing the best layer-module pair for vision and language, respectively. For VFMs, these gains increase with the transfer difficulty between pretraining and downstream data. 
\begin{center}
\faGithub \quad \href{https://github.com/ambroiseodt/waldo}{\texttt{github.com/ambroiseodt/waldo}}
\end{center}
\vspace{2pt}
\begin{figure}[!h]
    \centering
    \includegraphics[width=\linewidth]{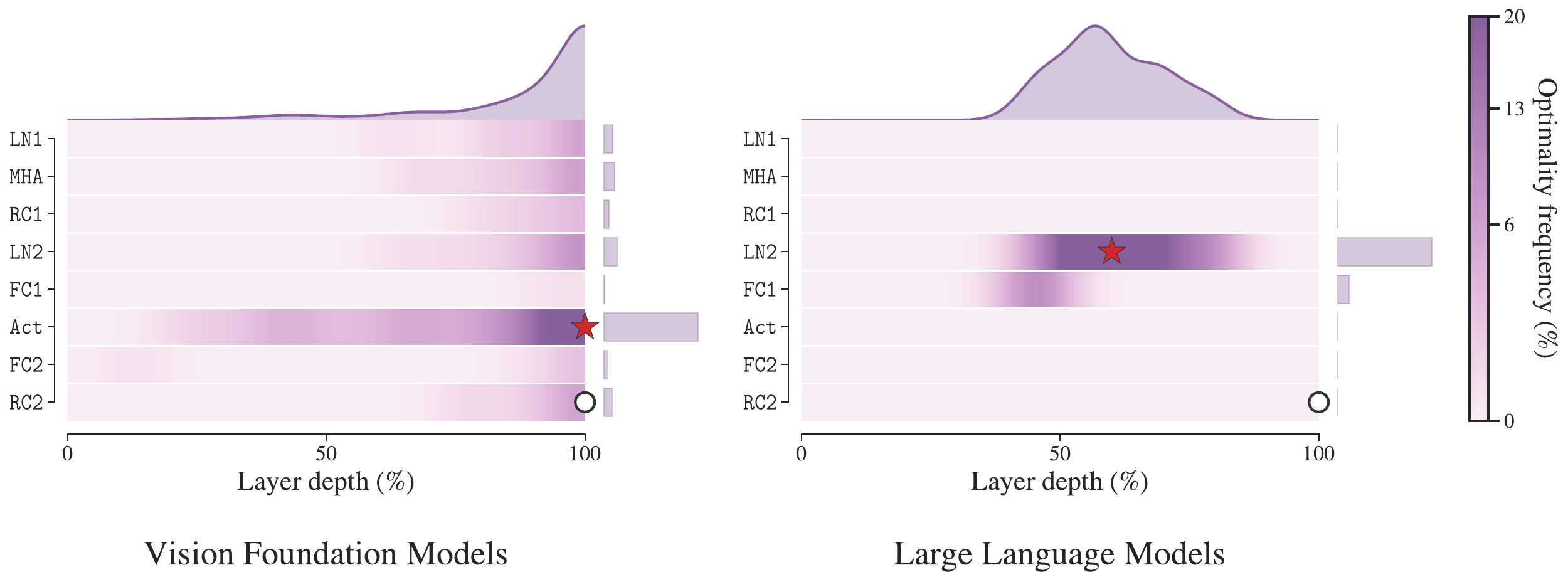}
    \caption{\textbf{Layer by layer, module by module.} Heatmaps showing the optimality frequency of each layer--module pair, i.e., how often it achieves the highest probing accuracy, with marginal distributions of optimal layers (top) and modules (right). For both vision foundation models (\textbf{left}) and large language models ($\textbf{right})$, performance depends jointly on layer and module choices. The probing baseline on the final-layer residual stream ($\circ$) is largely suboptimal, while the fixed strategies $\vfm$ and $\llm$, for vision and language (${\color{red}\star}$), respectively, often achieve the best performance.}
    \label{fig:layer_module_interaction}
    \vspace{-1em}
\end{figure}
\end{abstract}

\section{Introduction}
\label{sec:intro}
Foundation models, which rely primarily on the transformer architecture~\citep{vaswani2017attention}, have achieved impressive performance in a wide range of areas such as natural language processing~\citep[NLP,][]{touvron2023llama,brown2020gpt3}, computer vision~\citep{simeoni2025dinov3}, time series forecasting~\citep{nie2023patchtst,ilbert2024samformer}, and mathematical reasoning~\citep{comanici2025gemini25pushingfrontier}. These models are typically pretrained on massive amounts of diverse data~\citep{kimiteam2025kimivltechnicalreport,shukor2025scaling} to encode general knowledge in their weights. A common use of foundation models is as frozen feature extractors, which provide rich representations for downstream tasks without retraining or finetuning the model~\citep{oquab2024dinov2,lee2025geminiembedding,roschmann2026tivit}.

\paragraph{Probing intermediate representations.} \looseness=-1
A standard way to evaluate representations is through linear probing: the model is frozen, and a linear classifier is trained on top of the representation. \citet{alain2017probe} introduced probing as a diagnostic tool for neural networks and observed that classification accuracy increases with depth. Their experiments, however, focused on supervised, task-specific convolutional neural networks~\citep[CNN,][]{lecun1998cnn}. While the final layer may be optimal for a model trained for a single task, it need not be for a model trained to generalize across tasks. Recent work shows that intermediate layers of LLMs can outperform the final layer across diverse downstream tasks~\citep{skean2025layer}. For VFMs, however, the picture is less clear. \citet{skean2025layer} report that intermediate layers are beneficial in autoregressive VFMs~\citep{fini2025aimv2,elnouby2024aim}, while final representations remain optimal in non-autoregressive models such as ViT~\citep{dosovitskiy2021vit} and DINOv2~\citep{oquab2024dinov2}. In contrast, \citet{bolya2025perception} find that intermediate layers provide the best embeddings in contrastive VFMs, and~\citet{uselis2025intermediate} observe a similar trend in ViT, particularly under distribution shift. This motivates further investigation.

\paragraph{Layer by layer, module by module.}
\looseness=-1
The layer-wise analyses discussed above share a common limitation: they all treat the transformer block as an atomic unit. This perspective overlooks the intermediate representations produced within each block, illustrated in~\cref{fig:transformer_block}, namely the outputs of normalization layers, attention modules, feedforward linear layers, activation functions, and residual connections.\footnote{In what follows, we use the terms block and layer interchangeably, as well as module and component, referring to normalization layers, multi-head attention modules, feedforward layers, activation functions, and residual connections throughout the network.} These components can exhibit distinct properties, notably in terms of smoothness and adaptability during finetuning~\citep{odonnat2026vision}. We therefore conduct a large-scale, systematic study of transformer representations in VFMs and LLMs, layer by layer, module by module. We ask two natural questions: \emph{1) How much can optimal representations improve over standard probing?} and \emph{2) Can a simple fixed heuristic recover most of the gain at no extra cost?}

\paragraph{Our contributions.} 
\looseness=-1
We conduct extensive experiments on VFMs and LLMs. \cref{fig:layer_module_interaction} illustrates our findings, showcasing the suboptimality of probing the final-layer residual stream and the joint importance of both layer and module choices. Our main contributions can be summarized as:
\begin{enumerate}[leftmargin=*]
\item \textbf{Layer and module choices matter jointly.} Probing the best layer-module pair yields substantially larger gains over the final-layer residual stream than probing the best layer with the residual stream fixed. Moreover, module performance can vary substantially across depth, showing that layer and module choices should not be considered independently.
\item \textbf{Practical benefits.} We evaluate simple fixed probing strategies: final-layer feedforward activation ($\vfm$) for VFMs and the second normalization layer at $60\%$ depth ($\llm$) for LLMs. Both strategies improve performance in a majority of evaluations and recover a substantial fraction of the gains from probing the best layer-module pair.
\item \textbf{Impact of transfer difficulty.} When the discrepancy between pretraining and downstream data is large, transfer becomes more difficult. For VFMs, we find that the benefits of optimal probing locations increase in such settings, with $\vfm$ yielding larger gains. Moreover, the impact of depth grows under distribution shift, allowing further improvement through the best layer-wise $\act$ representations.
\end{enumerate}

\section{Probing transformers layer by layer, module by module}
\begin{wrapfigure}[15]{r}{0.5\linewidth}
\centering
\begin{adjustbox}{height=.5\linewidth}
    \begin{tikzpicture}
        \begin{scope}[xshift=-3cm]

\draw[fill=\background, line width=0.046875cm, rounded corners=0.300000cm] (-0.55000, -0.45) -- (2.8000, -0.45) -- (2.8000, 6.12) -- (-0.55000, 6.12) -- cycle;

\draw[line width=0.046875cm, -latex] (1.250000, -0.950000) -- (1.250000, 0.300000);

\draw[line width=0.046875cm, fill=ln1_color!80, rounded corners=0.100000cm] (0.450000, 0.800000) -- (2.0000, 0.800000) -- (2.0000, 0.300000) -- (0.450000, 0.300000) -- cycle;
\node[text width=2.500000cm, align=center] at (1.250000,0.550000) {$\mhanorm$};
\draw[line width=0.046875cm, -latex] (1.250000, 0.800000) -- (1.250000, 1.60000);

\draw[-latex, line width=0.046875cm, rounded corners=0.200000cm] (1.250000, -0.20000) -- (-0.25000, -0.2000) -- (-0.250000, 2.720000) -- (1.05000, 2.720000);
\draw[-latex, line width=0.046875cm, rounded corners=0.200000cm] (1.250000, 1.230000) -- (0.7000, 1.230000) -- (0.7000, 1.63000);
\draw[-latex, line width=0.046875cm, rounded corners=0.200000cm] (1.250000, 1.230000) -- (1.80, 1.230000) -- (1.80, 1.63000);

\draw[line width=0.046875cm, fill=mha_color!90, rounded corners=0.100000cm] (0.450000, 2.130000) -- (2.0000, 2.130000) -- (2.0000, 1.630000) -- (0.450000, 1.630000) -- cycle;
\node[text width=2.500000cm, align=center] at (1.250000,1.880000) {$\mha$};
\draw[line width=0.046875cm] (1.250000, 2.130000) -- (1.250000, 2.5250000);
\draw[line width=0.046875cm, -latex] (1.250000, 2.940000) -- (1.250000, 3.430000);

\node[circle, draw, minimum size=0.25em, fill=rc1_color!80, inner sep=0pt, line width=0.046875cm, align=center, label=right:{$\mhares$}] at (1.250000,2.720000) {$\bm{\mathbf{+}}$};

\draw[line width=0.046875cm, fill=ln2_color!70, rounded corners=0.100000cm] (0.450000, 3.930000) -- (2.00000, 3.930000) -- (2.00000, 3.430000) -- (0.450000, 3.430000) -- cycle;
\node[text width=2.500000cm, align=center] at (1.250000,3.68000) {$\ffnnorm$};
\draw[line width=0.046875cm, -latex] (1.250000, 3.93000) -- (1.250000, 4.68000);

\draw[-latex, line width=0.046875cm, rounded corners=0.200000cm] (1.250000, 3.050000) -- (-0.250000, 3.050000) -- (-0.250000,5.770000) -- (1.05000, 5.770000);

\draw[line width=0.046875cm, fill=white, rounded corners=0.100000cm] (0.450000, 5.18000) -- (2.00000, 5.18000) -- (2.00000, 4.68000) -- (0.450000, 4.68000) -- cycle;
\node[text width=2.500000cm, align=center] at (1.250000,4.930000) {\texttt{FFN}};
\draw[line width=0.046875cm] (1.250000, 5.180000) -- (1.250000, 5.5750000);
\draw[line width=0.046875cm, -latex] (1.250000, 5.990000) -- (1.250000, 6.480000);

\node[circle, draw, minimum size=0.25em, fill=rc2_color!60, inner sep=0pt, line width=0.046875cm, align=center, label=right:{$\ffnres$}] at (1.250000,5.770000) {$\bm{\mathbf{+}}$};

\draw[thick][dashed](2.0, 4.68) -- (3.05, 3.87);
\draw[thick][dashed] (2.0, 5.18) -- (3.05, 6.1);

\draw[line width=0.046875cm, fill=fc1_color!80, rounded corners=0.100000cm] (3.050000, 4.34) -- (4.6000, 4.34) -- (4.6000, 3.84) -- (3.050000, 3.84) -- cycle;
\node[text width=2.500000cm, align=center] at (3.850000,4.090000) {$\linact$};
\draw[line width=0.046875cm] (3.850000, 4.34) -- (3.850000, 4.685000);
\draw[line width=0.046875cm, -latex] (3.850000, 5.19000) -- (3.850000, 5.64000);

\node[circle, draw, minimum size=1.2em, fill=act_color!60, inner sep=0pt, line width=0.046875cm, align=center, label=right:{$\act$}] at (3.850000,4.930000) {$\bm{\sigma}$};

\draw[line width=0.046875cm, fill=fc2_color!70, rounded corners=0.100000cm] (3.050000, 6.1400) -- (4.60000, 6.14000) -- (4.60000, 5.64000) -- (3.050000, 5.64000) -- cycle;
\node[text width=2.500000cm, align=center] at (3.850000,5.890000) {$\actlin$};
\end{scope}
    \end{tikzpicture}
\end{adjustbox}
\begin{adjustbox}{height=.5\linewidth}
    \begin{tikzpicture}
        \begin{scope}[xshift=-3cm]
\draw[fill=\background, line width=0.046875cm, rounded corners=0.300000cm] (-0.55000, -0.45) -- (2.8000, -0.45) -- (2.8000, 6.12) -- (-0.55000, 6.12) -- cycle;

\draw[line width=0.046875cm, -latex] (1.250000, -0.950000) -- (1.250000, 0.300000);

\draw[line width=0.046875cm, fill=ln1_color!80, rounded corners=0.100000cm] (0.450000, 0.800000) -- (2.0000, 0.800000) -- (2.0000, 0.300000) -- (0.450000, 0.300000) -- cycle;
\node[text width=2.500000cm, align=center] at (1.250000,0.550000) {$\mhanorm$};
\draw[line width=0.046875cm, -latex] (1.250000, 0.800000) -- (1.250000, 1.60000);

\draw[-latex, line width=0.046875cm, rounded corners=0.200000cm] (1.250000, -0.20000) -- (-0.25000, -0.2000) -- (-0.250000, 2.720000) -- (1.05000, 2.720000);
\draw[-latex, line width=0.046875cm, rounded corners=0.200000cm] (1.250000, 1.230000) -- (0.7000, 1.230000) -- (0.7000, 1.63000);
\draw[-latex, line width=0.046875cm, rounded corners=0.200000cm] (1.250000, 1.230000) -- (1.80, 1.230000) -- (1.80, 1.63000);

\draw[line width=0.046875cm, fill=mha_color!90, rounded corners=0.100000cm] (0.450000, 2.130000) -- (2.0000, 2.130000) -- (2.0000, 1.630000) -- (0.450000, 1.630000) -- cycle;
\node[text width=2.500000cm, align=center] at (1.250000,1.880000) {$\mha$};
\draw[line width=0.046875cm] (1.250000, 2.130000) -- (1.250000, 2.5250000);
\draw[line width=0.046875cm, -latex] (1.250000, 2.940000) -- (1.250000, 3.430000);

\node[circle, draw, minimum size=0.25em, fill=rc1_color!80, inner sep=0pt, line width=0.046875cm, align=center, label=right:{$\mhares$}] at (1.250000,2.720000) {$\bm{\mathbf{+}}$};

\draw[line width=0.046875cm, fill=ln2_color!70, rounded corners=0.100000cm] (0.450000, 3.930000) -- (2.00000, 3.930000) -- (2.00000, 3.430000) -- (0.450000, 3.430000) -- cycle;
\node[text width=2.500000cm, align=center] at (1.250000,3.68000) {$\ffnnorm$};
\draw[line width=0.046875cm, -latex] (1.250000, 3.93000) -- (1.250000, 4.68000);

\draw[-latex, line width=0.046875cm, rounded corners=0.200000cm] (1.250000, 3.050000) -- (-0.250000, 3.050000) -- (-0.250000,5.770000) -- (1.05000, 5.770000);

\draw[line width=0.046875cm, fill=white, rounded corners=0.100000cm] (0.450000, 5.18000) -- (2.00000, 5.18000) -- (2.00000, 4.68000) -- (0.450000, 4.68000) -- cycle;
\node[text width=2.500000cm, align=center] at (1.250000,4.930000) {\texttt{FFN}};
\draw[line width=0.046875cm] (1.250000, 5.180000) -- (1.250000, 5.5750000);
\draw[line width=0.046875cm, -latex] (1.250000, 5.990000) -- (1.250000, 6.480000);

\node[circle, draw, minimum size=0.25em, fill=rc2_color!60, inner sep=0pt, line width=0.046875cm, align=center, label=right:{$\ffnres$}] at (1.250000,5.770000) {$\bm{\mathbf{+}}$};

\draw[thick][dashed](2.0, 4.68) -- (3.05, 3.38);
\draw[thick][dashed] (2.0, 5.18) -- (3.05, 6.38);

\draw[line width=0.046875cm, fill=fc1_color!80, rounded corners=0.100000cm] (3.050000, 3.845) -- (4.6000, 3.845) -- (4.6000, 3.345) -- (3.050000, 3.345) -- cycle;
\node[text width=2.500000cm, align=center] at (3.850000,3.5950000) {$\linact$};
\draw[line width=0.046875cm] (3.850000, 3.845) -- (3.850000, 4.610000);
\draw[line width=0.046875cm, -latex] (3.850000, 4.55500) -- (3.850000, 4.7100);

\draw[line width=0.046875cm, fill=white, rounded corners=0.100000cm] (4.850000, 3.845) -- (6.4000, 3.845) -- (6.400, 3.345) -- (4.850000, 3.345) -- cycle;
\node[text width=2.500000cm, align=center] at (5.625000,3.5950000) {\texttt{FC3}};
\draw[line width=0.046875cm, -latex]
(4.86000,3.825000) --
(3.990000,4.75000);

\node[
    fill=white,
    minimum size=0.6em,
    inner sep=0.05em,
    align=center
] at (3.850000,4.2275) {$\bm{\sigma}$};

\node[circle, draw, minimum size=1.2em, fill=act_color!60, inner sep=0pt, line width=0.046875cm, align=center, label={[yshift=0.0cm]right:{$\act$}}] at (3.850000,4.93) {$\bm{\odot}$};
\draw[line width=0.046875cm, -latex] (3.850000, 5.1500) -- (3.850000, 5.9150);

\draw[line width=0.046875cm, fill=fc2_color!70, rounded corners=0.100000cm] (3.050000, 6.4150) -- (4.60000, 6.41500) -- (4.60000, 5.91500) -- (3.050000, 5.91500) -- cycle;
\node[text width=2.500000cm, align=center] at (3.850000,6.165000) {$\actlin$};
\end{scope}
    \end{tikzpicture}
\end{adjustbox}
\caption{\textbf{Transformer representations.} We probe transformer representations (colored) across layers in VFMs and LLMs, covering models with standard (\textbf{left}) and gated FFN (\textbf{right}).}
\label{fig:transformer_block}
\end{wrapfigure}
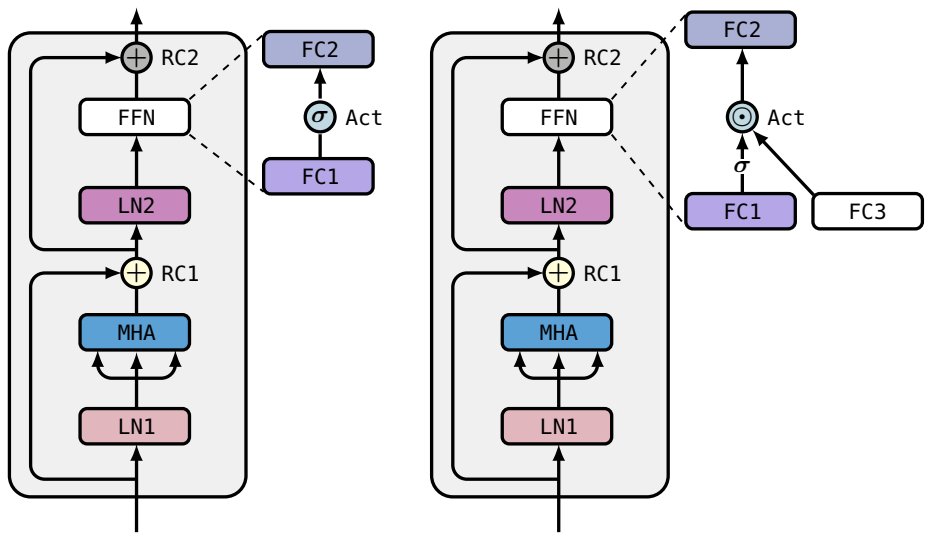

The transformer architecture~\citep{vaswani2017attention} consists of successive transformer blocks, each taking as input a sequence of tokens $(x_1, \ldots, x_n)$. As shown in~\cref{fig:transformer_block} (left), each block is made of alternating multi-head attention modules ($\mha$), with self-attention for encoders and causal attention for decoders, and feedforward networks (FFN). A standard FFN consists of two linear transformations,
denoted $\linact$ and $\actlin$, separated by a non-linear activation function $\act$, such as GELU~\citep{hendrycks2016gelu}. Assuming the pre-norm standard~\citep{xiong2020prenorm}, two normalization layers (LayerNorms~\citep{ba2016layernormalization} or RMSNorm~\citep{zhang2019rms}), respectively $\mhanorm$ and $\ffnnorm$, precede the attention and feedforward blocks, and two residual connections, respectively $\mhares$ and $\ffnres$, follow them. We will refer to $\ffnres$ as the residual stream, i.e., the output of the transformer block. In this work, we track the outputs of these eight sequence-to-sequence operations at each layer, indexing them by a layer-module pair $(\ell,m)$. Some large-scale transformers replace the standard two-layer FFN with a gated FFN. It combines the non-linear projection with an additional linear projection through element-wise multiplication. To ensure consistency between standard and gated FFNs, we recover the FFN representations as follows: $\linact$ is the linear layer fed as input to the non-linear operation $\sigma$, $\act$ is the non-linear transformation with a gated mechanism (thus absorbing the additional linear projection), fed as input to the second linear function, and $\actlin$ is this second linear projection. This is illustrated in~\cref{fig:transformer_block}, with exact tensor definitions deferred to~\cref{app:gated_ffn}.

\paragraph{Evaluating transformer representations.} \looseness=-1
Following~\citet{oquab2024dinov2} and~\citet{muennighoff2023mteb}, we evaluate transformer representations via linear probing for classification~\citep{alain2017probe}, which provides a common evaluation for visual~\citep{chen2020simclr} and textual~\citep{lee2025geminiembedding} embeddings. Each probing location $(\ell,m)$ produces a sequence of tokens $(x_1, \ldots, x_n) \in (\RR^d)^n$, where $\ell$ denotes the relative depth expressed as a percentage\footnote{We use zero-based layer indexing, such that for a model with $L$ layers, probing at relative depth $\ell$ corresponds to layer $\operatorname{int}(\frac{\ell}{100} \cdot (L-1))$.}, and $m$ is the transformer module being probed. By abuse of notation, $d$ denotes either the embedding dimension or the FFN hidden dimension, which is generally larger. As in~\citet{joseph2026interpreting}, we embed this sequence in $\mathbb{R}^{d}$ by global average pooling over content tokens, namely image patches or text tokens; that is, special tokens, including \texttt{CLS}, register tokens~\citep{darcet2024vision}, \texttt{EOS}, and \texttt{BOS}, as well as padding positions (\texttt{PAD}), are masked before pooling and do not contribute to the representation. This allows us to focus on representations produced within transformer blocks, with a consistent feature extraction procedure across models (e.g., AIMv2 models~\citep{fini2025aimv2} do not use a \texttt{CLS} token). Implementation details to ensure a scalable evaluation across layers and modules are provided in~\cref{app:implem_details}.

\paragraph{Measuring improvement with oracle gains.} \looseness=-1
The performance of the representation at location $(\ell,m)$ is the probing test accuracy on a given dataset or benchmark, denoted by $\mathrm{acc}(\ell,m)$. We use the final-layer residual stream, i.e., $\ffnres$ at $100\%$ relative depth, as the probing baseline, denoted by $\texttt{RC2@100}$. We define the gain at location $(\ell,m)$ as the accuracy difference in percentage points (pp) relative to the baseline, i.e., $g(\ell,m) = \mathrm{acc}(\ell,m) - \mathrm{acc}(100,\mathrm{RC2})$. To disentangle the impact of depth, module, and their interaction, we analyze the following oracle gains:
\begin{equation}
\label{def:gains}
\layer = \max_{\ell} \ g(\ell, \mathrm{RC2}), \quad
\module = \max_{m} \ g(100, m), \quad
\oracle = \max_{\ell, m} \ g(\ell, m).
\end{equation}
The term oracle emphasizes that these quantities are a posteriori upper bounds among the evaluated probing locations, since they require exhaustive evaluation and access to test performance. In particular, $\oracle$ defines the optimal performance among the evaluated probing locations.

\section{Experiments}
\label{sec:exp}
In this section, we address the questions raised in~\cref{sec:intro} through large-scale evaluations of vision (\cref{sec:vfm_exp}) and language models (\cref{sec:llm_exp}); see~\cref{app:implem_details} for the full implementation details. 

\subsection{Probing vision foundation models}
\label{sec:vfm_exp}
We consider models that cover different pretraining paradigms and scales: supervised ViT models~\citep{dosovitskiy2021vit}, discriminative self-supervised DINOv2~\citep{oquab2024dinov2} and DINOv3 models~\citep{simeoni2025dinov3}, and autoregressive AIMv2 models~\citep {fini2025aimv2}. This results in a total of $15$ publicly available models, ranging from $22$M to $7$B parameters. Experiments are conducted on a diverse set of $18$ image classification datasets, commonly studied in the literature~\citep{elnouby2024aim, oquab2024dinov2}. \cref{tab:image_classification} provides an overview of the datasets. Following~\citet{oquab2024dinov2}, linear probing is done via logistic regression, with features preprocessed using a standard scaler. The performance of a given representation is its top-1 test accuracy at the dataset level ($\%$). Overall, VFM experiments comprise nearly $90\,000$ probing evaluations.

\begin{wrapfigure}[21]{r}{0.4\linewidth}
    \centering
    \includegraphics[width=\linewidth]{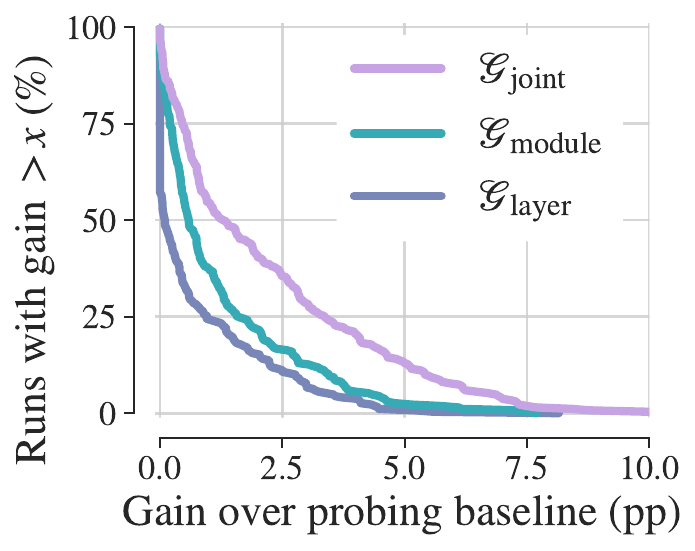}
    \caption{\textbf{Substantial gains from optimal probing locations in VFMs.} Complementary cumulative distribution of gains over the probing baseline across $270$ model-dataset pairs (each defining a run) when using the best representation across layers ($\layer$), modules ($\module$), or both ($\oracle$).}
    \label{fig:cdf_vision}
\end{wrapfigure}

\paragraph{Optimal representations yield substantial gains.} 
\looseness=-1
We first analyze the oracle gains provided by the best probing locations across depth and modules. \cref{fig:cdf_vision} displays the complementary 
cumulative distribution of gains over the probing baseline. Each curve shows, for a given gain threshold $x$, the percentage of model-dataset pairs with gain greater than $x$ when probing the best representation over layers alone ($\layer$), modules alone ($\module$), or both jointly ($\oracle$). We observe consistent gains, reaching up to $10$ pp with $\oracle \geq \module \geq \layer$ across the distribution. The first inequality follows directly from~\cref{def:gains}, while the second highlights the importance of module choice, which is often overlooked in prior work. Moreover, the gap between $\oracle$ and $\module$ demonstrates the additional benefit of jointly selecting layers and modules.
\begin{figure}[!b]
    \centering
    \includegraphics[width=\linewidth]{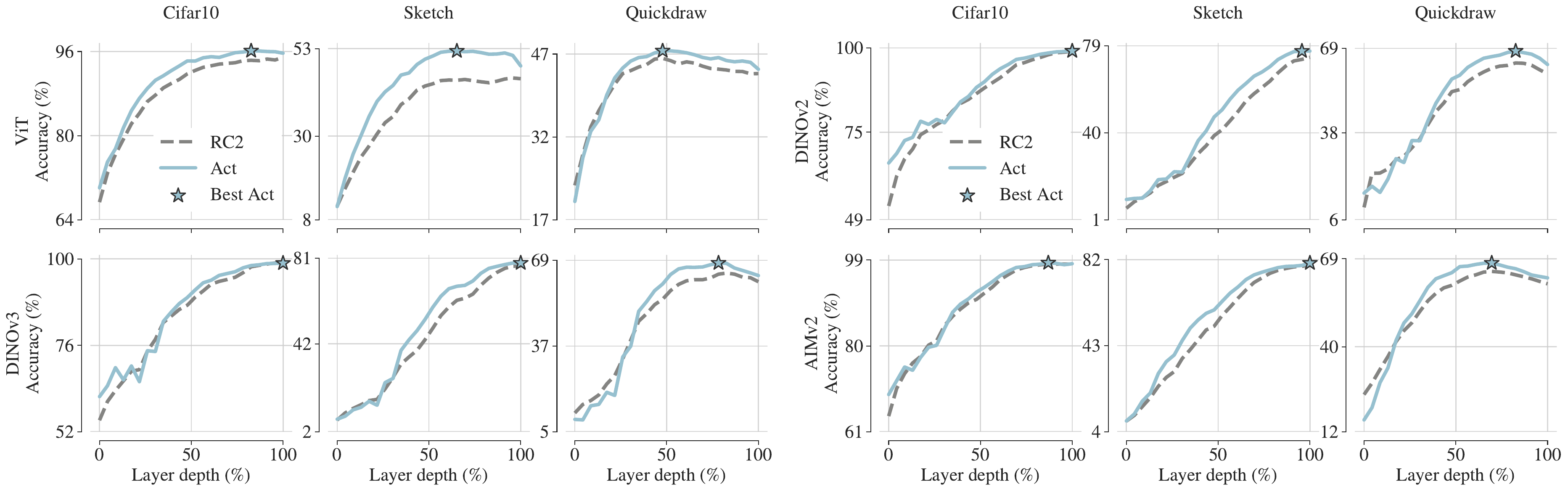}
    \caption{\textbf{Impact of $\act$ and intermediate layers in VFMs.} Probing accuracy across layers for $\act$ and $\ffnres$ on Cifar10, Sketch, and Quickdraw. Results are reported for the $307$M-parameter variant, common to each model family. The star indicates the best $\act$ performance over depth.}
    \label{fig:probing_depth_vision}
\end{figure}

\begin{table}[!t]
\centering
\caption{\textbf{Performance comparison in VFMs} between the probing baseline on the final-layer residual stream ($\texttt{RC2@100}$), the fixed strategy on the final-layer activation ($\vfm$), and the oracle strategy that probes the best representation over layers and modules for each model-dataset pair. Results are reported for the $307$M-parameter variant, common to each model family, with datasets ordered from left to right by increasing mean probing error (easy to hard). The \texttt{Baseline} row reports the probing test accuracy ($\%$), while $\vfm$ and \texttt{Oracle} report accuracy differences in percentage points relative to the baseline, shown in \textcolor{Green}{green} when positive and \textcolor{Red}{red} otherwise.}
\label{tab:vfm_results}
\resizebox{\linewidth}{!}{
\begin{NiceTabular}{lccccccccccccccccccc}
\toprule
Strategy & \rotatebox{90}{Flowers102} & \rotatebox{90}{EuroSAT} & \rotatebox{90}{Cifar10} & \rotatebox{90}{Pet} & \rotatebox{90}{Contrast} & \rotatebox{90}{Food-101} & \rotatebox{90}{Cifar100} & \rotatebox{90}{Motion Blur} & \rotatebox{90}{DTD} & \rotatebox{90}{Speckle Noise} & \rotatebox{90}{Clipart} & \rotatebox{90}{Painting} & \rotatebox{90}{Aircraft} & \rotatebox{90}{Sketch} & \rotatebox{90}{iWildCam} & \rotatebox{90}{Quickdraw} & \rotatebox{90}{Infograph} & \rotatebox{90}{RxRx1} & \rotatebox{90}{Average}
\\[-1pt]
\midrule
\multicolumn{20}{l}{\textit{ViT}} \\
\arrayrulecolor{gray!50}
\midrule
\texttt{Baseline} & 89.9 & 96.5 & 94.9 & 85.4 & 90.2 & 73.2 & 82.8 & 79.5 & 72.6 & 74.5 & 59.9 & 61.0 & 49.6 & 45.1 & 62.8 & 43.5 & 23.9 & 2.1 & 66.0 \\
$\vfm$ & \add{+2.0} & \add{+0.4} & \add{+0.8} & \add{+4.5} & \sub{-0.5} & \add{+7.3} & \add{+2.1} & \add{+1.2} & \add{+1.1} & \add{+0.2} & \add{+0.7} & \add{+4.0} & \add{+4.6} & \add{+3.4} & \add{+3.5} & \add{+0.8} & \add{+4.6} & \add{+0.1} & \add{+2.3} \\
\oraclerow \texttt{Oracle} & \add{+5.4} & \add{+1.9} & \add{+1.2} & \add{+4.8} & \add{+1.4} & \add{+8.8} & \add{+2.9} & \add{+2.7} & \add{+2.6} & \add{+3.5} & \add{+5.0} & \add{+6.7} &\add{+6.2} & \add{+7.3} & \add{+4.4} & \add{+4.1} & \add{+7.2} & \add{+2.7} & \add{+4.4} \\
\arrayrulecolor{black}
\midrule
\multicolumn{20}{l}{\textit{DINOv2}} \\
\arrayrulecolor{gray!50}
\midrule
\texttt{Baseline} & 99.6 & 97.4 & 99.0 & 91.4 & 96.1 & 91.7 & 91.4 & 87.4 & 81.6 & 85.3 & 82.3 & 77.9 & 70.3 & 73.9 & 70.1 & 59.6 & 44.3 & 4.8 & 78.0 \\
$\vfm$ & \sub{-0.1} & \add{+0.3} & \add{+0.1} & \add{+1.4} & 0.0 & \add{+1.6} & \add{+0.5} & \add{+0.7} & \sub{-0.9} & \add{+0.6} & \add{+0.4} & \add{+1.3} & \add{+3.2} & \add{+2.8} & \add{+2.1} & \add{+3.5} & \add{+3.8} & \add{+1.4} & \add{+1.3} \\
\oraclerow \texttt{Oracle} & \add{+0.1} & \add{+1.0} & \add{+0.1} & \add{+1.4} & \add{+0.4} & \add{+1.6} & \add{+0.8} & \add{+1.9} & \add{+1.2} & \add{+2.0} & \add{+0.8} & \add{+1.8} &\add{+6.5} & \add{+2.8} & \add{+3.0} & \add{+8.4} & \add{+4.6} & \add{+5.4} & \add{+2.4} \\
\arrayrulecolor{black}
\midrule
\multicolumn{20}{l}{\textit{DINOv3}} \\
\arrayrulecolor{gray!50}
\midrule
\texttt{Baseline} & 99.7 & 97.5 & 98.8 & 94.8 & 95.6 & 93.8 & 92.0 & 87.5 & 82.7 & 88.4 & 85.1 & 81.0 & 78.9 & 77.4 & 67.1 & 61.0 & 49.9 & 5.1 & 79.8 \\
$\vfm$ & 0.0 & \sub{-0.3} & \add{+0.1} & \sub{-1.1} & \sub{-0.1} & \add{+0.7} & \add{+0.1} & \sub{-0.4} & \sub{-1.8} & \sub{-0.1} & \sub{-0.3} & \sub{-0.1} & \sub{-1.8} & \add{+1.4} & \add{+5.8} & \add{+2.3} & \add{+2.1} & \add{+0.7} & \add{+0.4} \\
\oraclerow \texttt{Oracle} & 0.0 & \add{+0.8} & \add{+0.1} & 0.0 & \add{+0.1} & \add{+0.7} & \add{+0.1} & 0.0 & \add{+1.5} & \add{+0.5} & \add{+0.3} & 0.0 & \add{+4.1} & \add{+1.4} & \add{+5.8} & \add{+7.0} & \add{+2.7} & \add{+4.9} & \add{+1.7} \\
\arrayrulecolor{black}
\midrule
\multicolumn{20}{l}{\textit{AIMv2}} \\
\arrayrulecolor{gray!50}
\midrule
\texttt{Baseline} & 99.6 & 97.7 & 98.1 & 95.1 & 93.5 & 93.8 & 88.9 & 87.5 & 83.9 & 76.2 & 86.3 & 82.1 & 74.1 & 79.4 & 71.6 & 60.7 & 56.4 & 5.1 & 79.4 \\
$\vfm$ & 0.0 & 0.0 & \add{+0.1} & \add{+0.6} & \add{+0.8} & \add{+0.6} & \add{+0.4} & \add{+0.2} & \sub{-0.7} & \sub{-0.2} & \add{+0.1} & \add{+0.6} & \add{+3.8} & \add{+1.0} & \sub{-0.3} & \add{+2.0} & \add{+3.4} & \add{+0.5} & \add{+0.7} \\
\oraclerow \texttt{Oracle} & 0.0 & \add{+1.1} & \add{+0.3} & \add{+0.6} & \add{+0.8} & \add{+0.6} & \add{+0.8} & \add{+0.9} & \add{+0.6} & \add{+4.4} & \add{+0.2} & \add{+0.6} & \add{+3.8} & \add{+1.0} & \add{+1.9} & \add{+6.9} & \add{+3.4} & \add{+4.5} & \add{+1.8} \\
\arrayrulecolor{black}
\bottomrule
\end{NiceTabular}
}
\vspace{-1em}
\end{table}
\paragraph{Towards a fixed probing strategy.} 
\looseness=-1
Ideally, we would like to identify a probing location at a fixed layer-module pair that recovers part, if not all, of the gain. We first examine how often each module appears in the best layer-module pair achieving $\oracle$. We find that $\act$ is optimal for $64.8\%$ of the $270$ model-dataset pairs, compared with $9.3\%$ for $\ffnnorm$, the next most frequent, and $5.9\%$ for the baseline module $\ffnres$ (full results in~\cref{tab:vfm_module_comparison}). This suggests that it is a good candidate for the fixed module choice. Since $\oracle$ is an oracle quantity, the superiority of $\act$ is established a posteriori. \cref{fig:probing_depth_vision} displays the performance of $\act$ and the baseline module $\ffnres$ across layers. We can see that the $\act$ module displays strong performance over layers, highlighting the importance of the module choice. The best $\act$ representation, marked with a star, is generally in the second half of the network, but not necessarily at the final layer. This is particularly clear on Quickdraw, where intermediate representations outperform the final layer. The same phenomenon occurs in DINOv3, despite its classification performance being reported to increase smoothly towards the final layers~\citep[see][Section 6.2.1]{simeoni2025dinov3}. These observations hold across model sizes, as shown in~\cref{fig:probing_depth_vit_part0,fig:probing_depth_vit_part1,fig:probing_depth_vit_part2,fig:probing_depth_dinov2_part0,fig:probing_depth_dinov2_part1,fig:probing_depth_dinov2_part2,fig:probing_depth_dinov3_part0,fig:probing_depth_dinov3_part1,fig:probing_depth_dinov3_part2,fig:probing_depth_aimv2_part0,fig:probing_depth_aimv2_part1,fig:probing_depth_aimv2_part2}. The interaction between layers and modules is illustrated in~\cref{fig:layer_module_interaction} (left), which displays the frequency with which each layer-module pair achieves the oracle gain $\oracle$ across $270$ model-dataset pairs. The $\act$ module indeed has the highest optimality frequency, and despite variations in the depth, optimal representations are most frequently found at the final layer. We therefore consider the simple fixed strategy of probing final-layer $\act$, which we denote by $\vfm$.

\paragraph{Performance comparison.} 
\looseness=-1
\begin{wrapfigure}[20]{r}{0.4\linewidth}
    \vspace{-2em}
    \centering
    \includegraphics[width=\linewidth]{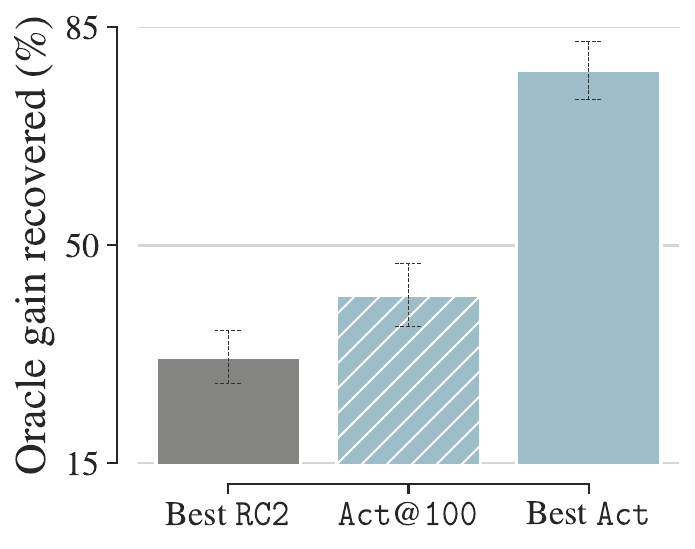}
    \caption{\textbf{$\vfm$ recovers nearly half of the oracle gain.} Recovery of oracle gain $\oracle$, averaged across all model-dataset pairs with $95\%$ confidence intervals. The fixed strategy $\vfm$ outperforms the best layer-wise $\ffnres$, while probing the best layer-wise $\act$ recovers $78.1\%$ of the gain.}
    \label{fig:oracle_recovery_vision}
\end{wrapfigure}
Table~\ref{tab:vfm_results} compares the performance of the probing baseline $\texttt{RC2@100}$ and the fixed strategy $\vfm$ on the $307$M-parameter variant common to each model family (full results deferred to~\cref{app:vfm_full_results} due to space constraints). The oracle strategy, which probes the best layer-module pair, is displayed as an empirical upper bound. $\vfm$ yields gains for each model family, with an average improvement of $2.3$ pp for ViT, $1.3$ pp for DINOv2, $0.4$ pp for DINOv3, and $0.7$ pp for AIMv2. In total, $\vfm$ improves performance over the probing baseline in $63.7\%$ of the $270$ model-dataset pairs. The fixed strategy is not always optimal across model sizes and families. This reflects, in part, the fact that the optimal probing depth varies across datasets (see~\cref{app:probing_depth_vision}). Yet, as shown in~\cref{fig:oracle_recovery_vision}, $\vfm$ recovers without any search $41.9\%$ of the gain from probing the best layer-module pair ($\oracle$), compared to $31.9\%$ for the best layer-wise $\ffnres$. The best layer-wise $\act$ further increases recovery up to $78.1\%$, showing that a simple fixed choice of module can already substantially improve performance. We can see in~\cref{tab:vfm_results} that the gains tend to increase from left to right. They are small on easy datasets such as Flowers102 and Pets, which are typically included in pretraining datasets~\citep[see][Table 15]{oquab2024dinov2}, but more pronounced on challenging benchmarks. On \textsc{Wilds}, where distribution shifts naturally arise~\citep{koh2021wilds}, and DomainNet~\citep{peng2019domainnet}, $\vfm$ improves performance in $89.3\%$ of the $28$ evaluations, with an average gain of $1.8$ pp. A similar trend is observed for the \texttt{Oracle} strategy, suggesting that gains over the probing baseline are larger as transfer difficulty increases. We discuss this in the next paragraph.

\begin{figure}[!h]
\centering
\begin{subfigure}{.25\linewidth}
  \centering
  \includegraphics[width=\linewidth]{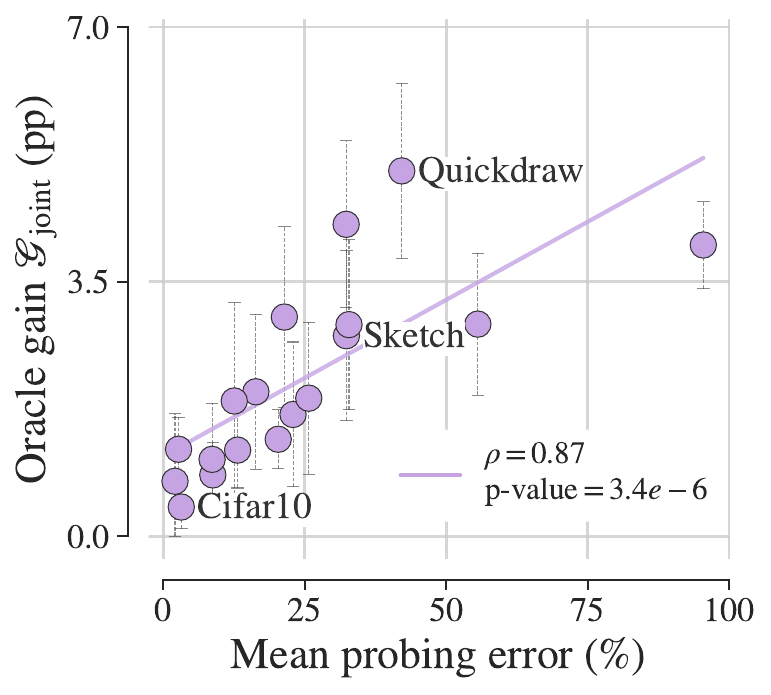}
  \caption{Transfer difficulty.}
  \label{fig:transfer_difficulty}
\end{subfigure}%
\begin{subfigure}{.25\linewidth}
  \centering
  \includegraphics[width=\linewidth]{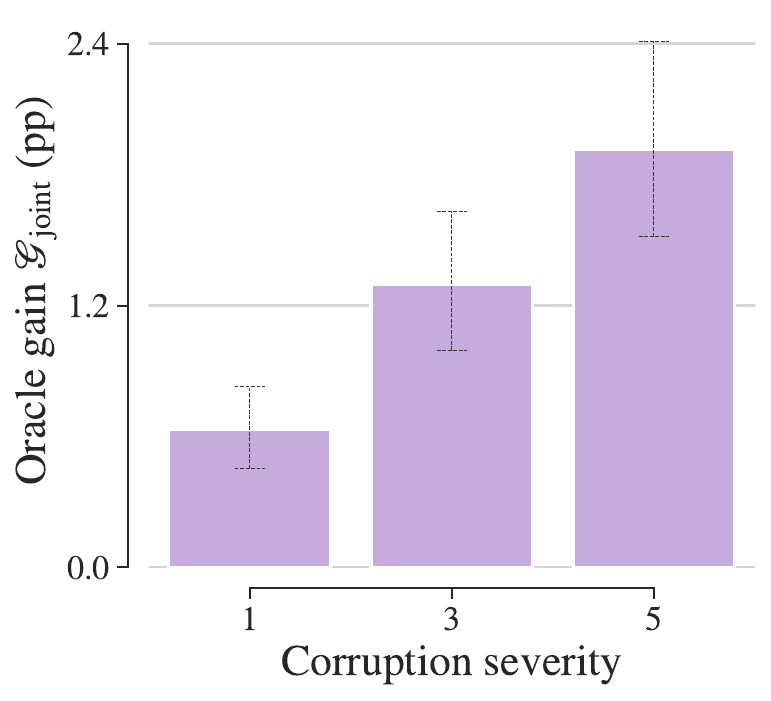}
  \caption{Corruption severity.}
  \label{fig:domain_generalization}
\end{subfigure}%
\begin{subfigure}{.25\linewidth}
  \centering
  \includegraphics[width=\linewidth]{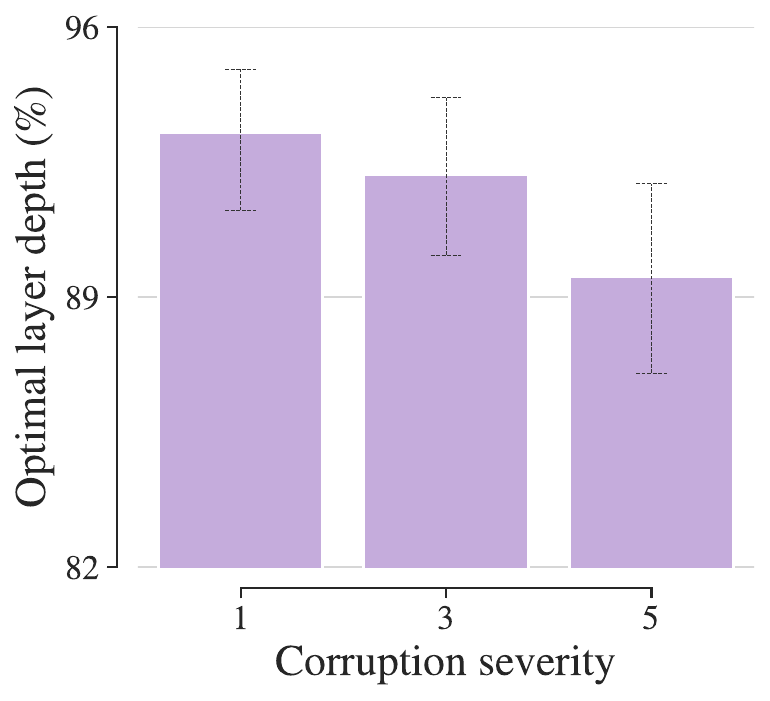}
  \caption{Optimal depth.}
  \label{fig:optimal_depth}
\end{subfigure}%
\begin{subfigure}{.25\linewidth}
  \centering
  \includegraphics[width=\linewidth]{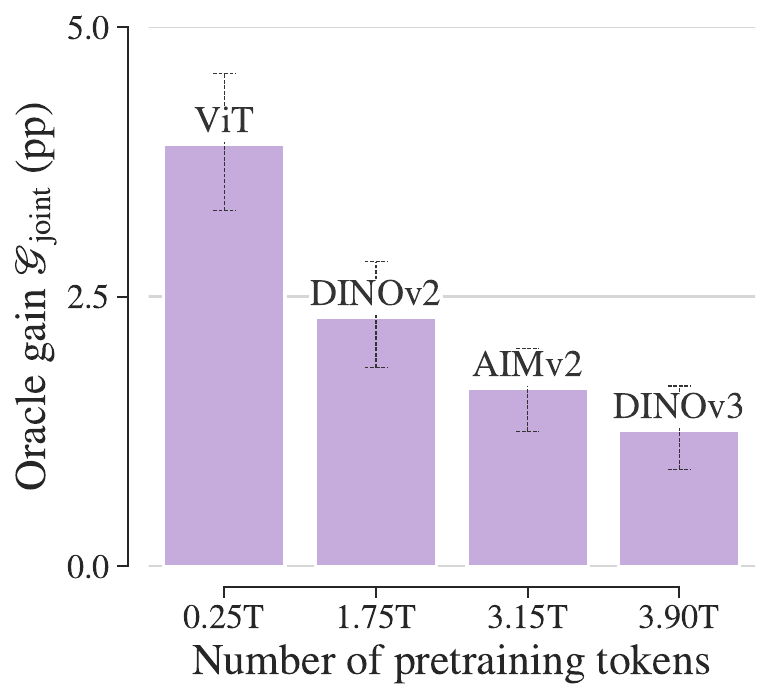}
  \caption{Pretraining scale.}
  \label{fig:pretraining_scale}
\end{subfigure}
\caption{\textbf{Oracle gains increase with transfer difficulty and corruption severity, but decrease with pretraining scale.} \textbf{(a)} Oracle gain as a function of empirical transfer difficulty, defined as the mean probing error over evaluated models. Markers show the gain for each dataset, averaged across $15$ models, with a $95\%$ confidence interval. \textbf{(b)} Oracle gain as a function of corruption severity on $5$ types of corruption from Cifar10-C, averaged across $15$ models, with a $95\%$ confidence interval. \textbf{(c)} The optimal depth associated with $\oracle$ decreases with corruption severity. \textbf{(d)} Oracle gain across model families, averaged across datasets and model sizes, with a $95\%$ confidence interval. Models are ordered by increasing number of tokens seen during pretraining.}
\label{fig:vfm_rq2}
\vspace{-0.5em}
\end{figure}

\paragraph{Gain increases with transfer difficulty.} \looseness=-1
Linear probing evaluates the transferability of frozen representations on downstream data. We therefore define the \emph{empirical transfer difficulty} of a dataset as the mean probing error ($\%$) of the baseline across all evaluated models, with higher values indicating more difficult transfer from pretraining to downstream data. \cref{fig:transfer_difficulty} shows that the oracle gain $\oracle$ is positively correlated with transfer difficulty (Spearman $\rho=0.87$, $\text{p-value}<0.001$). This trend is illustrated by Cifar10, Sketch, and Quickdraw, which progressively depart from natural-image statistics: \cref{fig:model_family_vision} shows that the benefits of intermediate depth and alternative modules increase with transfer difficulty on these representative datasets. Our finding at the level of individual representations is consistent with Head2Toe~\citep{evci2022head2toe}, which shows that aggregating intermediate representations provides larger gains for target domains farther from the pretraining domain. We then consider controlled distribution shifts, following the evaluation protocol of DINOv3~\citep[see][Section 6.2.1]{simeoni2025dinov3}. We train the classifier on features extracted from Cifar10 and evaluate it on a corrupted test set from Cifar10-C over $5$ corruption types (Contrast, Gaussian Noise, Motion Blur, Snow, Speckle Noise) and increasing severities in $\{1,3,5\}$. As shown in~\cref{fig:domain_generalization,fig:optimal_depth}, the oracle gain $\oracle$ increases with corruption severity, while the optimal depth decreases. This suggests that the benefits of intermediate layers increase with distribution shift. 
To further investigate when these representations are beneficial, we conduct a controlled study in~\cref{app:shift}, where we probe a pretrained model across layers before and after finetuning it on downstream data. Before finetuning, accuracy curves are concave when pretraining and downstream data differ substantially, with intermediate representations achieving the best performance. After finetuning, the curves straighten, with performance increasing monotonically towards the final layers. Finally, \cref{fig:pretraining_scale} shows that oracle gains decrease with pretraining scale, computed as the number of tokens seen during pretraining (see~\cref{app:pretraining_scale} for details). The relative error reductions follow the same order, with $15.7\%$ for ViT, $12.1\%$ for DINOv2, $10.8\%$ for AIMv2, and $7.6\%$ for DINOv3. Intuitively, more extensive pretraining increases the chance of seeing samples similar to the downstream images, potentially reducing the discrepancy between pretraining and downstream data distributions and leaving a smaller oracle gain to close. This is consistent with~\cref{fig:transfer_difficulty}. Model size, in contrast, has no significant effect on gains; see~\cref{app:friedman} for details. We note that these findings are observational, not causal, since many confounders exist in our analysis (architectural differences, training objectives, quality and diversity of pretraining data).

\begin{tcolorbox}[colback=\boxcol,
    colframe=black,
    arc=4pt,
    boxsep=0pt,
    boxrule=\boxwidth pt,
]%
\textbf{Takeaway 1.} For VFMs, the best probing locations across layers and modules yield substantial performance gains. The simple fixed strategy $\vfm$ improves performance in a majority of evaluations, recovering nearly half of these gains at no extra cost. Layer and module choices become more important as the discrepancy between pretraining and downstream data increases, while near-in-distribution datasets tend to benefit less from changing the probing location.
\end{tcolorbox}

\subsection{Probing large language models}
\label{sec:llm_exp}
We now extend our analysis to large language models. As in~\citet{skean2025layer}, we consider general-purpose LLMs: Qwen3~\citep{yang2025qwen3}, Llama3~\citep{grattafiori2024llama3}, and Gemma3~\citep{kammath2025gemma3}, for a total of $9$ publicly available models, ranging from $1$B to $12$B parameters. Following prior work on textual embeddings~\citep{lee2025geminiembedding,skean2025layer,li2025mixture,tang2026kv}, we use the Massive Multilingual Text Embedding Benchmark~\citep[\texttt{MMTEB},][]{enevoldsen2025mmteb}. We evaluate using the official \texttt{MMTEB} implementation (available at \url{https://github.com/embeddings-benchmark/mteb/}) via linear probing with logistic regression on $24$ classification tasks from the \texttt{MTEB(Multilingual)} benchmark, spanning $31$ languages and $14$ domains. \cref{tab:mteb_tasks} provides an overview of the tasks. Following \texttt{MTEB} evaluation protocol~\citep{enevoldsen2025mmteb}, the performance of a given representation is its average test accuracy across tasks ($\%$). Overall, the LLM experiments comprise over $54\,000$ probing evaluations.
\paragraph{Optimal representations yield substantial gains.} 
\looseness=-1
\begin{figure}[!h]
    \centering
    \vspace{-1em}
    \includegraphics[width=\linewidth]{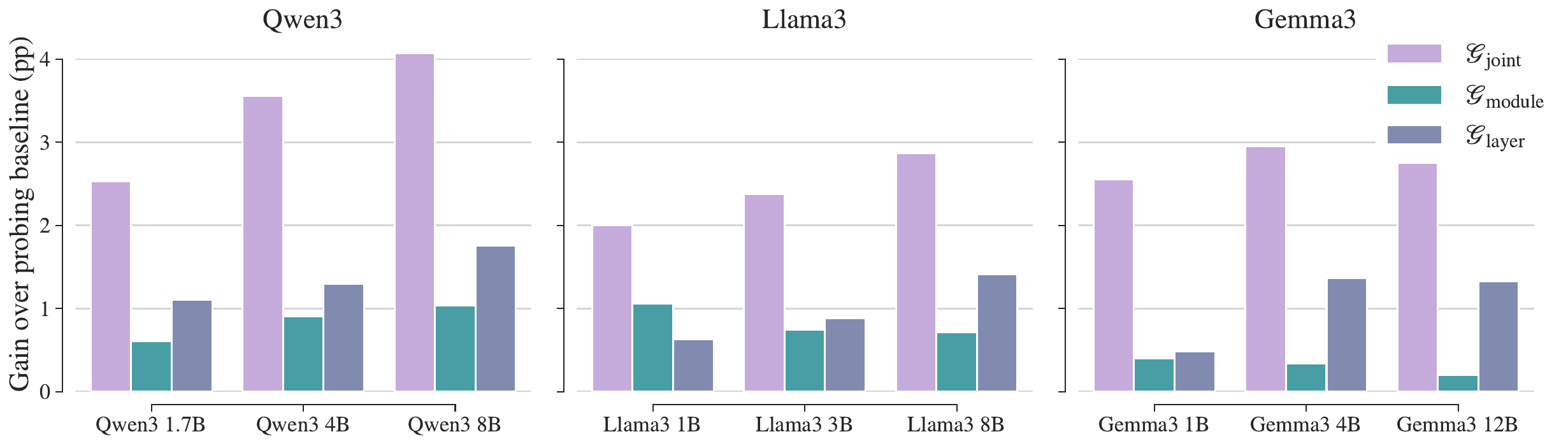}
    \caption{\textbf{Substantial gains from optimal probing locations in LLMs.} Gains over the probing baseline on \texttt{MTEB(Multilingual)} from probing the best representation across layers ($\layer$), modules ($\module$), or both ($\oracle$) in Qwen3, Llama3, and Gemma3 models.}
    \label{fig:model_family_language}
\end{figure}
\cref{fig:model_family_language} displays the gains from probing the best representation over layers alone ($\layer$), modules alone ($\module$), or both jointly ($\oracle$).  Gains are consistent across models and sizes, reaching up to $4$ pp. Qwen3 models display larger gains than Llama3 and Gemma3 models, with the improvement increasing with model size within the Qwen3 and Llama3 families. $\layer \geq \module$ holds for all but one model, while $\oracle$ systematically exceeds both, highlighting the benefits of jointly selecting layers and modules.
\begin{figure}[!b]
    \centering
    \vspace{-1em}
    \includegraphics[width=\linewidth]{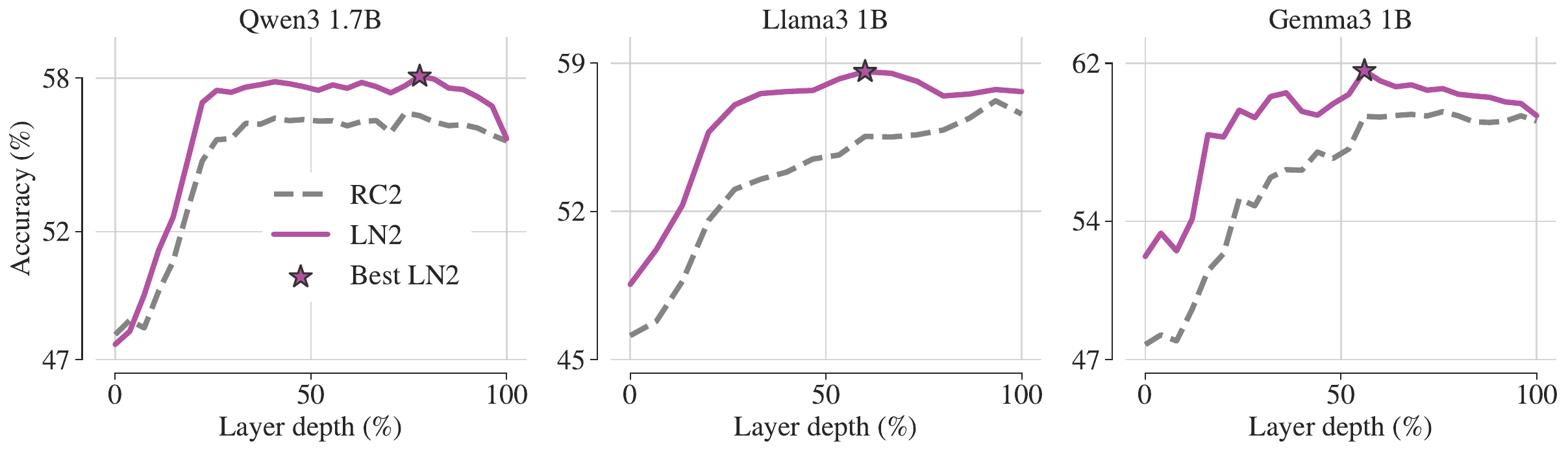}
    \caption{\textbf{Impact of $\ffnnorm$ and intermediate layers on performance.} Probing accuracy across layers for $\ffnnorm$ and $\ffnres$ on \texttt{MTEB(Multilingual)}. The star indicates the best $\ffnnorm$ performance over depth.}
    \label{fig:probing_depth_language}
    \vspace{-1em}
\end{figure}
\paragraph{Towards a fixed probing strategy.}
Similarly to~\cref{sec:vfm_exp}, we aim for a simple fixed probing location that improves performance. The fixed choice is simpler than for VFMs, since $\ffnnorm$ appears in the best layer-module pair achieving $\oracle$ for all but one model (see~\cref{tab:llm_module_comparison}). We note that the baseline module $\ffnres$ is never part of the optimal layer-module pair. Since $\oracle$ is an oracle quantity, the superiority of $\ffnnorm$ is established a posteriori. Akin to~\cref{fig:probing_depth_vision}, we observe in~\cref{fig:probing_depth_language} the performance of $\ffnnorm$ and the baseline module $\ffnres$ over depth.  $\ffnnorm$ consistently surpasses $\ffnres$ across layers, highlighting the importance of the module choice. Intermediate $\ffnnorm$ layers outperform the final layer for all the models considered, which is reminiscent of~\citet{skean2025layer}. Although the optimal depth varies between families, around $70\%$ for Qwen3 models and near $50\%$ for Llama3 and Gemma3 models, the interval $[50\%,70\%]$ is remarkably stable across models and sizes, as shown in~\cref{fig:layer_module_interaction,fig:probing_depth_all_language}. Its midpoint, $60\%$, corresponds to the average optimal depth across models. The benefit of $\ffnnorm$ at $60\%$ depth is illustrated in~\cref{fig:layer_module_interaction} (right), which displays the optimality frequency of each layer-module pair across models. We therefore consider the simple fixed probing strategy of probing $\ffnnorm$ at $60\%$ depth. We denote it by $\llm$.

\begin{table}[!t]
\small
\centering
\caption{\textbf{Performance comparison in LLMs} between the probing baseline on the final-layer residual stream ($\ffnres$), residual-stream probing at $60\%$ depth ($\texttt{RC2@60}$), the fixed strategy $\llm$, and the oracle strategy that probes the best layer-module pair for each model. The accuracy difference in percentage points relative to the probing baseline is shown in \textcolor{Green}{green} if positive, and in \textcolor{Red}{red} otherwise.}
\label{tab:llm_results}
\resizebox{\textwidth}{!}{
\begin{NiceTabular}{@{} l *{9}{w{r}{2.8em}@{\hspace{3pt}}w{r}{1.7em}@{\hspace{1pt}}}@{}}
\toprule
\multirow{2}{*}{Strategy} & \multicolumn{6}{c}{\textit{Qwen3}} & \multicolumn{6}{c}{\textit{Llama3}} & \multicolumn{6}{c}{\textit{Gemma3}} \\
\cmidrule(l){2-7} \cmidrule(l){8-13} \cmidrule(l){14-19}
 & \multicolumn{2}{c}{1.7B} & \multicolumn{2}{c}{4B} & \multicolumn{2}{c}{8B} & \multicolumn{2}{c}{1B} & \multicolumn{2}{c}{3B} & \multicolumn{2}{c}{8B} & \multicolumn{2}{c}{1B} & \multicolumn{2}{c}{4B} & \multicolumn{2}{c}{12B} \\
\midrule
\texttt{Baseline} & 55.5 &  & 56.6 &  & 57.6 &  & 56.6 &  & 58.9 &  & 59.9 &  & 59.1 &  & 60.9 &  & 61.5 &  \\
\texttt{RC2@60} & 56.1 & \addsmall{+0.6} & 56.7 & \addsmall{+0.1} & 58.3 & \addsmall{+0.7} & 55.5 & \subsmall{-1.1} & 59.7 & \addsmall{+0.8} & 61.3 & \addsmall{+1.3} & 59.3 & \addsmall{+0.2} & 61.7 & \addsmall{+0.8} & 62.6 & \addsmall{+1.0} \\
$\llm$ & 57.6 & \addsmall{+2.1} & 59.4 & \addsmall{+2.9} & 61.1 & \addsmall{+3.5} & 58.6 & \addsmall{+2.0} & 61.1 & \addsmall{+2.2} & 62.3 & \addsmall{+2.4} & 61.1 & \addsmall{+2.0} & 63.1 & \addsmall{+2.3} & 64.2 & \addsmall{+2.6} \\
\oraclerow \texttt{Oracle} & 58.1 & \addsmall{+2.5} & 60.1 & \addsmall{+3.6} & 61.6 & \addsmall{+4.1} & 58.6 & \addsmall{+2.0} & 61.3 & \addsmall{+2.4} & 62.8 & \addsmall{+2.9} & 61.6 & \addsmall{+2.6} & 63.8 & \addsmall{+3.0} & 64.3 & \addsmall{+2.8} \\
\bottomrule
\end{NiceTabular}
}
\end{table}
\paragraph{Performance comparison.}
\cref{tab:llm_results} compares the performance of the probing baseline $\texttt{RC2@100}$ and the fixed strategy $\llm$ on \texttt{MTEB(Multilingual)}. The oracle strategy, which probes the best layer-module pair for each model, is displayed as an empirical upper bound.
\begin{wrapfigure}[21]{r}{0.4\linewidth}
    \centering
    \includegraphics[width=\linewidth]{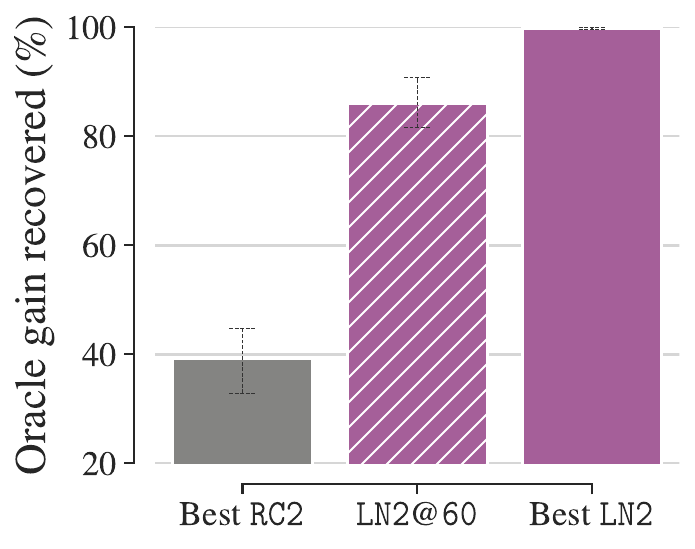}
    \caption{\textbf{$\llm$ recovers substantial gains.} Recovery of oracle gain $\oracle$, averaged across all models with $95\%$ confidence intervals. The fixed $\llm$ strategy outperforms the best layer-wise $\ffnres$ strategy, while probing the best layer-wise $\ffnnorm$ recovers $99.8\%$ of the gain.}
    \label{fig:oracle_recovery_language}
\end{wrapfigure}
To disentangle the effects of layer and module choices, we also report probing on the residual stream at $60\%$ depth ($\texttt{RC2@60}$). At the benchmark level, $\llm$ consistently outperforms the probing baseline, with performance close to the oracle strategy for an average gain of $2.44$ percentage points. $\texttt{RC2@60}$ yields smaller, less consistent gains, highlighting the importance of both layer and module choices. As in~\cref{fig:model_family_language}, the gains of $\llm$ increase with model size within each family. Using $\llm$ allows Gemma3 $4$B to match the performance of GritLM-7B~\citep{muennighoff2025gritlm}, the second-best embedding model on \texttt{MTEB(Multilingual)} at the time of \texttt{MMTEB} publication~\citep[see][Table 2]{enevoldsen2025mmteb}, without finetuning. Similarly, it enables Gemma3 $12$B to surpass GritLM-7B and match EmbeddingGemma~\citep{vera2025embeddinggemma}, the embedding model directly derived from Gemma3 (see~\cref{app:embedding_models} for details). Overall, $\llm$ improves over the probing baseline in $21$ of $24$ tasks ($87.5\%$) and in $77.8\%$ of the $216$ individual model-task evaluations (detailed results are provided in~\cref{app:llm_full_results}). We can see in~\cref{fig:oracle_recovery_language} that $\llm$ recovers without any search $86.0\%$ of the gain from probing the best layer-module pair, compared to $39.3\%$ for the best layer-wise \texttt{RC2}. The best layer-wise $\ffnnorm$ nearly closes the gap, recovering $99.8\%$ of the oracle gain.

\begin{tcolorbox}[colback=\boxcol,
    colframe=black,
    arc=4pt,
    boxsep=0pt,
    boxrule=\boxwidth pt,
]%
\textbf{Takeaway 2.} For LLMs, the best probing locations across layers and modules yield substantial performance gains. The simple fixed strategy $\llm$ recovers $86.0\%$ of these gains at no extra cost, with nearly all of the remaining gap captured by the best layer-wise $\ffnnorm$ representations.
\end{tcolorbox}

\section{Related work}
\label{sec:related_work}

\paragraph{Probing foundation models.} Linear probing~\citep{alain2017probe,belinkov2022probing} has become a standard evaluation protocol to assess representations in foundation models across diverse modalities, including images~\citep{chen2020simclr,oquab2024dinov2}, text~\citep{muennighoff2023mteb,enevoldsen2025mmteb,lee2025geminiembedding}, video~\citep{assran2025vjepa2}, graphs~\citep{velickovic2018deep}, and time series~\citep{roschmann2026tivit,goswami2024moment,feofanov2026mantis}. Beyond linear heads, attentive probing is increasingly adopted to evaluate models with patch-level features~\citep{chen2024cae,elnouby2024aim,fini2025aimv2}. While linear probes are still used, the embedding of sequences of tokens from $(\RR^d)^n$ to $\RR^d$ is learned via an attention mechanism at the level of the sequence. The overall operation thus becomes non-linear and leads to noticeable performance improvements~\citep{elnouby2024aim}. In these works, features are typically extracted from the final layer after a forward pass. We challenge this convention, showing the benefits of probing intermediate representations, including within transformer blocks. Beyond probing at a single location, some approaches aggregate internal representations throughout the model, combining them with learned layer weights~\citep{peters2018elmo}, supervised feature selection~\citep{evci2022head2toe}, attention over all layers~\citep{ciernik2026attentive}, or attention value vectors pooled across layers~\citep{zhang2026llmbased}. Our work is orthogonal to these methods: it could help identify which representations across layers and modules are worth aggregating, and, more broadly, improve our understanding of transformer representations. 

\paragraph{Understanding transformer representations.} Intermediate representations can also be studied to understand what models encode rather than to improve downstream performance. Probing intermediate text representations reveals what linguistic knowledge is captured within the model~\citep{liu2019linguistic} and where: \citet{tenney2019bert} show that BERT~\citep{devlin2019bert} represents the steps of the traditional NLP pipeline in an interpretable and localizable way. Similarly,~\citet{joseph2026interpreting} use linear probes to understand where physical information becomes accessible in encoder-based video transformers, and how it is organized. Linear probes are also central to \emph{mechanistic interpretability}, which aims to reverse-engineer the computations of neural networks, with a strong focus on LLMs. This line of work views the transformer as a set of attention and feedforward modules that read from and write to the shared residual stream~\citep{elhage2021mathematical}. Probes and interventions have thus been used to analyze attention heads~\citep{li2023inferencetime}, feedforward modules~\citep{geva2021ffn}, and the residual stream~\citep{gurnee2024language,marks2024geometry}, where the latter can be steered to control the model behavior~\citep{turner2024steering,rimsky2024steering}. Note that, akin to feature evaluation, the residual stream is the typical representation studied in the mechanistic interpretability literature, often referred to as \emph{residual stream activation}. Beyond interpretability, \citet{odonnat2026vision} analyze the smoothness of individual transformer components to determine which ones to prioritize during finetuning. Our work follows this direction of looking inside transformer blocks. Yet, the goal differs: we aim to identify which internal representations are most useful as frozen features for downstream tasks, not to reverse-engineer computations, elicit a specific behavior in the model, or finetune it. We discuss additional related work in~\cref{app:related_work}.

\section{Discussion}
This paper investigates how useful different transformer representations are for linear probing in vision foundation models and large language models. Our extensive \emph{layer by layer, module by module} analysis shows that layer and module choices should be considered jointly: the best layer-module pair yields substantially larger performance gains than selecting the best layer with the residual stream fixed, while module performance can vary substantially across depth. Building on these findings, we identify simple fixed probing strategies, $\vfm$ and $\llm$ for VFMs and LLMs, respectively, that improve performance in a majority of evaluations and recover a large fraction of the gains from the best layer-module pair. For vision models, the benefit of $\vfm$ increases with transfer difficulty between pretraining and downstream data. These findings highlight the importance of looking beyond depth alone when probing transformer models. We hope our work can contribute to a better understanding of their internal representations.

\paragraph{Limitations and future work.} Our analysis focuses on linear probing for classification and could naturally be extended to other downstream tasks and modalities, including tabular and time series foundation models. Moreover, our insights into the usefulness of individual transformer representations could help guide the design of aggregation mechanisms that combine intermediate representations rather than treating them uniformly.

\section*{AI use statement}
Generative AI tools were used to aid and polish the writing. They were also used to discuss research ideas and experimental design, including the experimental setups of prior work, and to assist with rapid prototyping of visualizations. The authors carefully checked all LLM-generated outputs and independently implemented and verified the final experiments, analyses, and visualizations. The authors take full responsibility for the contents of this paper.

\section*{Ethics statement}
The goal of this work is to advance the field of Machine Learning. There are many potential societal consequences of our work, none of which we feel must be specifically highlighted here.

\section*{Reproducibility statement}
The code is open-sourced at \url{https://github.com/ambroiseodt/waldo}. All the implementation details to reproduce our results are provided in~\cref{app:implem_details}, including datasets, benchmarks, models, and pretrained weights used. Experimental setups are provided in the main paper, along with additional details and results in~\cref{app:add_exp}. 

\bibliography{references}
\bibliographystyle{template/iclr2027_conference}

\clearpage
\appendix
\textbf{\LARGE Appendix}
\paragraph{Roadmap.} We discuss additional related work in~\cref{app:related_work}. The full implementation details are given in~\cref{app:implem_details} and, finally, extensive additional experiments are provided in~\cref{app:add_exp}. We display the corresponding table of contents below.

\addtocontents{toc}{\protect\setcounter{tocdepth}{3}}

\renewcommand*\contentsname{\Large Table of Contents}
\tableofcontents
\clearpage

\section{Extended related work}
\label[secinapp]{app:related_work}
In this section, we extend the discussion of prior works related to our paper in~\cref{sec:related_work}.

\paragraph{Better transfer of intermediate layers.}
Several works propose explanations as to why intermediate representations can generalize better than the final layer, especially under distribution shift. The intrinsic dimension of representations first increases then decreases with depth~\citep{ansuini2019intrinsic,recanatesi2019dimensionality}, and final layers exhibit neural collapse, compressing features onto class means~\citep{papyan2020neuralcollapse}. This phenomenon also appears, more weakly, in intermediate layers~\citep{rangamani2023neuralcollapse}. In addition, the tunnel effect describes later layers compressing representations that are already linearly separable through intermediate layers, thus degrading out-of-distribution performance~\citep{masarczyk2023tunnel}. Although these works mostly study task-specific deep neural networks rather than large-scale pretrained foundation models, their observations are consistent with our finding that the optimal probing depth moves to earlier layers as the shift between pretraining and downstream data grows.

\paragraph{Intermediate representations for adaptation and embeddings.} Intermediate representations can also be used to adapt pretrained models efficiently. Visual Query Tuning~\citep{tu2023vqt} learns queries that summarize intermediate features of vision transformers, while~\citet{wu2024structured} implement an efficient adaptation model that selects features using the structure of the model and adapts to task difficulty. In language, LLMEmbed~\citep{liu2024llmembed} combines representations from several layers of lightweight LLMs for text classification. When LLMs are used as frozen embedding models, training-free methods can improve the quality of extracted features by modifying either the input, with prompts~\citep{jiang2024scaling}, or repetitions~\citep{springer2025repetition}, or the forward pass, by re-routing internal key-value states~\citep{tang2026kv}. The adaptation methods aggregate residual stream representations across depth, while the training-free methods read the final-layer residual stream. Both could benefit from our insights by investigating other transformer modules and, for the latter, other depths as well.

\section{Implementation details}
\label[secinapp]{app:implem_details}
In this section, we provide the full implementation details necessary to reproduce our results.

\subsection{Transformer representations in gated FFN}
\label[secinapp]{app:gated_ffn}
In large-scale transformers~\citep{oquab2024dinov2,touvron2023llama}, an increasingly common alternative to the two-layer MLP feedforward network is the use of gated activations~\citep{dauphin2017gated, shazeer2020glu}. Let $d$ be the embedding dimension, $(x_1, \ldots, x_n) \in (\RR^d)^n$ be an input sequence of tokens, viewed as a matrix in $\RR^{d \times n }$, $h \in \NN$ be the FFN dimension, and $\sigma$ be an activation function. Let $W_1 \in \mathbb{R}^{h \times d},\ W_2 \in \mathbb{R}^{d \times h}, W_3 \in \mathbb{R}^{h \times d}$ be weight matrices. Denoting the element-wise multiplication by $\odot$ and omitting the biases for readability, the standard FFN and the gated FFN are written as follows
\begin{equation}
\label{eq:ffn}
    \mathrm{FFN}(x) = W_2\sigma(W_1 x), \quad \mathrm{FFN}_\mathrm{gated}(x) = W_2 (\sigma(W_1 x) \odot W_3x).
\end{equation}
The main difference is that the gated FFN combines a non-linear projection with an additional linear projection through element-wise multiplication. To ensure consistency between the standard and the gated FFN, we recover the FFN representations as follows: $\linact$ is the linear layer fed as input to the activation function ($W_1x$ in both cases), $\act$ is the non-linear transformation fed as input to the second linear function ($\sigma(W_1x)$ for the standard FFN and $\sigma(W_1 x) \odot W_3x$ for the gated FFN), and, finally, $\actlin$ is the output of the feedforward network ($\mathrm{FFN}(x)$ and $\mathrm{FFN}_\mathrm{gated}(x)$ in~\eqref{eq:ffn}). This is illustrated in~\cref{fig:transformer_block}.

\subsection{Vision foundation models}
\label[secinapp]{app:vfm_implem}
This section provides details on VFMs' transformer representations and the linear probing setup.

\subsubsection{Transformer representations}
\label[secinapp]{app:vfm_transformer_representation}
We recover embeddings across layers and modules with a single forward pass using PyTorch forward hooks and pre-hooks. In our code, all VFMs inherit from the abstract base \texttt{BaseVisionModel} displayed in~\cref{fig:base_vfm}. The feature extraction is done with the \texttt{get\_embeddings} method. The most important abstract method to implement is \texttt{block\_adapter}, which defines the $8$ representations to extract within each transformer block. The corresponding abstract base class \texttt{TransformerBlockAdapter} is displayed in~\cref{fig:base_block}. We provide below the implementation details on the VFMs used in our experiments.

\begin{figure}[!t]
\begin{center}
\begin{minipage}{\linewidth}
\begin{lstlisting}[language=Python, frame=single]
class BaseVisionModel(nn.Module, ABC):
    r"""Abstract base class for VFMs."""

    def __init__(self):
        super().__init__()
        self.device: str
        self.model: PreTrainedModel

    @abstractmethod
    def _load_model(self, model_name: str, device: str) -> PreTrainedModel:
        r"""Load a pretrained HuggingFace model."""
        ...

    @abstractmethod
    def pooling(self, hidden_states: torch.Tensor) -> torch.Tensor:
        r"""Global average pooling method over content tokens."""
        ...

    @property
    @abstractmethod
    def block_adapter(self) -> type[TransformerBlockAdapter]:
        r"""Return the block adapter class for the model's transformer blocks."""
        ...

    @property
    @abstractmethod
    def blocks(self) -> ModuleList:
        r"""Return the list of transformer blocks."""
        ...

    def _make_pre_hook(self, embeddings: dict[str, torch.Tensor], key: str) -> Callable:
        r"""Create a forward pre-hook that captures the INPUT of a module."""
        def pre_hook_fn(module: nn.Module, input: tuple[torch.Tensor, ...]) -> None:
            embeddings[key] = self.pooling(hidden_states=input[0])
        return pre_hook_fn

    def _make_hook(self, embeddings: dict[str, torch.Tensor], key: str) -> Callable:
        r"""Create a forward hook that captures the OUTPUT of a module."""
        def hook_fn(module: nn.Module, input: tuple[torch.Tensor, ...], output: torch.Tensor | tuple[torch.Tensor, ...]) -> None:
            out = output[0] if isinstance(output, tuple) else output
            embeddings[key] = self.pooling(hidden_states=out)
        return hook_fn

    @torch.inference_mode()
    def get_embeddings(self, pixel_values: torch.Tensor, layers: list[int], modules: list[str]) -> dict[str, torch.Tensor]:
        r"""Single forward pass recovering transformer  representations over layers and modules."""
        # Recover layers and modules
        layers = layers if layers is not None else list(range(len(self.blocks)))
        modules = modules if modules is not None else TRANSFORMER_MODULES

        # Loop over transformer blocks to register hooks across specified layers and modules
        embeddings: dict[str, torch.Tensor] = {}
        handles: list[torch.utils.hooks.RemovableHandle] = []
        layer_set = set(layers)
        for layer_idx, layer in enumerate(self.blocks):
            if layer_idx not in layer_set:
                continue
            adapter = self.block_adapter(layer)
            for module_name, module in adapter.module_map().items():
                if module_name not in modules:
                    continue
                key = f"block_{layer_idx}_{module_name}"

                # Modules for which we hook the inputs
                if module_name in adapter.pre_hooks:
                    handles.append(
                        module.register_forward_pre_hook(self._make_pre_hook(embeddings=embeddings, key=key))
                    )
                # Modules for which we hook the output
                else:
                    handles.append(module.register_forward_hook(self._make_hook(embeddings=embeddings, key=key)))

        # Forward pass
        try:
            self.model(pixel_values=pixel_values)
        finally:
            for handle in handles:
                handle.remove()

        return embeddings
\end{lstlisting}
\end{minipage}
\end{center}
\caption{\texttt{BaseVisionModel} implementation.}
\label{fig:base_vfm}
\end{figure}

\begin{figure}[!t]
\begin{center}
\begin{minipage}{\linewidth}
\begin{lstlisting}[language=Python, frame=single]
class TransformerBlockAdapter(ABC):
    r"""Adapter that maps standard module names to transformer block's submodules.

    Subclasses need to implement the 8 methods that outputs the following modules:
        ln1: first layer norm
        mha: multi-head attention
        rc1: first residual connection
        ln2: second layer norm
        fc1: first feedforward linear layer
        act: feedforward activation function
        fc2: second feedforward linear layer
        rc2: second residual connection
    """

    def __init__(self, layer: nn.Module):
        self.ln1 = self._get_ln1(layer)
        self.mha = self._get_mha(layer)
        self.rc1 = self._get_rc1(layer)
        self.ln2 = self._get_ln2(layer)
        self.fc1 = self._get_fc1(layer)
        self.act = self._get_act(layer)
        self.fc2 = self._get_fc2(layer)
        self.rc2 = self._get_rc2(layer)
        self.pre_hooks = self._get_pre_hooks_modules(layer)

    @abstractmethod
    def _get_ln1(self, layer: nn.Module) -> nn.Module: ...

    @abstractmethod
    def _get_mha(self, layer: nn.Module) -> nn.Module: ...

    @abstractmethod
    def _get_rc1(self, layer: nn.Module) -> nn.Module: ...

    @abstractmethod
    def _get_ln2(self, layer: nn.Module) -> nn.Module: ...

    @abstractmethod
    def _get_fc1(self, layer: nn.Module) -> nn.Module: ...

    @abstractmethod
    def _get_act(self, layer: nn.Module) -> nn.Module: ...

    @abstractmethod
    def _get_fc2(self, layer: nn.Module) -> nn.Module: ...

    @abstractmethod
    def _get_rc2(self, layer: nn.Module) -> nn.Module: ...

    @abstractmethod
    def _get_pre_hooks_modules(self, layer: nn.Module) -> set[str]:
        r"""Returns the modules for which pre-hooks needs to be used in place of forward hooks.
        """
        ...

    def module_map(self) -> dict[str, nn.Module]:
        return {
            "ln1": self.ln1,
            "mha": self.mha,
            "rc1": self.rc1,
            "ln2": self.ln2,
            "fc1": self.fc1,
            "act": self.act,
            "fc2": self.fc2,
            "rc2": self.rc2,
        }
\end{lstlisting}
\end{minipage}
\end{center}
\caption{\texttt{TransformerBlockAdapter} implementation.}
\label{fig:base_block}
\end{figure}

\paragraph{ViT.} The Vision Transformer architecture~\citep[ViT,][]{dosovitskiy2021vit} consists of a succession of transformer encoders with LayerNorms~\citep{ba2016layernormalization} and a standard feedforward network with GELU activation~\citep{hendrycks2016gelu}. Inputs are $2$D images that are split into square patches of size $P$, which are flattened and linearly embedded in dimension $d$. A classification token \texttt{CLS} is prepended to the sequence of tokens before adding positional embeddings. The checkpoints used in our experiments follow the HuggingFace transformers~\citep{wolf2020hftransformers} implementation available at \href{https://github.com/huggingface/transformers/blob/main/src/transformers/models/vit/modeling_vit.py}{\code{transformers/models/vit/modeling\_vit.py}}.

\paragraph{DINOv2.} The DINOv2 backbone~\citep{oquab2024dinov2} follows the ViT architecture described above, adding a LayerScale~\citep{touvron2021layerscale} to stabilize training. All variants use a standard FFN, except the $1.1$B one, which uses a gated FFN with a SwiGLU activation~\citep{shazeer2020glu}. The checkpoints used in our experiments follow the HuggingFace transformers' implementation available at \href{https://github.com/huggingface/transformers/blob/main/src/transformers/models/dinov2/modeling_dinov2.py}{\code{transformers/models/dinov2/modeling\_dinov2.py}}.

\paragraph{DINOv3.} The DINOv3 backbone~\citep{simeoni2025dinov3} follows the DINOv2 one and implements Rotary Positional Embeddings~\citep[RoPE,][]{su2024rope}. All variants use a standard FFN, except the $7$B one, which uses a gated FFN with a SwiGLU activation~\citep{shazeer2020glu}. The checkpoints used in our experiments follow the HuggingFace transformers' implementation available at \href{https://github.com/huggingface/transformers/blob/main/src/transformers/models/dinov3_vit/modeling_dinov3_vit.py}{\code{transformers/models/dinov3\_vit/modeling\_dinov3\_vit.py}}.

\paragraph{AIMv2.} The AIMv2 model~\citep{fini2025aimv2} pairs a ViT backbone with a multimodal decoder to autoregressively generate raw image patches and text tokens. In our work, we only focus on the vision part, which consists of a succession of transformer encoders with RMS norms~\citep{zhang2019rms} and a gated feedforward network with SwiGLU
activation~\citep{shazeer2020glu}. Inputs are $2$D images that are split into square patches of size $P$, which are flattened and linearly embedded in dimension $d$. Unlike the original ViT, no \texttt{CLS} token is prepended to the sequence: attention is instead constrained with a prefix causal mask during pretraining, enabling bidirectional attention over image patches at inference without further tuning. The checkpoints used in our experiments follow the HuggingFace transformers' implementation available at \href{https://github.com/huggingface/transformers/blob/main/src/transformers/models/aimv2/modeling_aimv2.py}{\code{transformers/models/aimv2/modeling\_aimv2.py}}.

\paragraph{Feature extraction.} Applying linear probing with a logistic regression requires feeding the regressor a matrix of embeddings over the entire train split (resp. test split).  However, even over a single batch of $512$ samples, a forward pass with a \say{small} model like the ViT-B with $86$M parameters amounts to saving $49$K tensors in $\RR^{768}$, taking around $144$MiB in memory. Such an extraction over an entire dataset is thus intractable, even on high-performance GPUs like H200. To tackle this issue, we perform the probing in two steps: 1) feature extraction with embeddings saved on disk, and 2) logistic regression over frozen features. For the first step, we save each batch of pooled representations on the CPU and stream it directly to disk to maintain low GPU memory consumption and idle time, independent of the number of layers. This step is done for both the training and test samples. Then, for each (layer, module) pair, we rebuild the dataset-level embeddings over the train split (resp. test split) to fit (resp. evaluate) the logistic regression. Our implementation allows us to use logistic regression even for large-scale VFMs with up to $7$B parameters.

\subsubsection{Probing setup}
\label[secinapp]{app:vfm_probing_setup}
We conduct the probing experiments across various pretraining paradigms, model sizes, and datasets.
\paragraph{ViT.} We conduct experiments on supervised ViT models~\citep{dosovitskiy2021vit} pretrained on ImageNet-21K~\citep{deng2009imagenet}, the superset of the ILSVRC-2012 ImageNet dataset with $21$k classes and $14$ million images. We consider $3$ variants with $86$M, $307$M, and $632$M parameters, and a patch size of $16$, $16$, and $14$, respectively. The input images have a resolution of $224$ for all models. Following the original paper nomenclature~\citep[see][Section 4.1]{dosovitskiy2021vit}, we refer to those models as ViT-B, ViT-L, and ViT-H (we omit the patch size in the nomenclature since we consider one value for each variant in our work). All variants were pretrained from scratch. Their characteristics are given in~\cref{tab:vit_variants}, along with the link to the checkpoints used in our experiments.
\begin{table}[!h]
    \centering
        \caption{\textbf{Characteristics of ViT variants~\citep{dosovitskiy2021vit}.} We provide the number of layers, the embedding dimension, the FFN hidden dimension, the number of attention heads, the patch size, the number of parameters, and the link to the HuggingFace checkpoint.}
        \resizebox{\linewidth}{!}{
            \begin{NiceTabular}{lccccccc}
            \toprule
            Model & Layers & Emb. dim. & FFN dim. & Attn. heads & Patch size & Parameters & Checkpoint \\
            \midrule
            ViT-B  & 12 & 768  & 3072 & 12 & 16 & 86M  & \href{https://huggingface.co/google/vit-base-patch16-224-in21k}{\texttt{vit-base}}\\
            ViT-L & 24 & 1024 & 4096 & 16 & 16 & 307M & \href{https://huggingface.co/google/vit-large-patch16-224-in21k}{\texttt{vit-large}} \\
            ViT-H  & 32 & 1280 & 5120 & 16 & 14 & 632M & \href{https://huggingface.co/google/vit-huge-patch14-224-in21k}{\texttt{vit-huge}} \\
            \bottomrule
            \end{NiceTabular}
        }
    \label{tab:vit_variants}
\end{table}

\paragraph{DINOv2.} We also consider DINOv2 models~\citep{oquab2024dinov2} pretrained in discriminative self-supervised fashion. For DINOv2, we consider $4$ variants with $22$M, $86$M, $307$M, and $1.1$B parameters. The patch size is fixed at $14$, and the input images have a resolution $224$ for all models. The corresponding backbones are referred to as ViT-S/14, ViT-B/14, ViT-L/14, and ViT-g/14, respectively. Since they all share the same backbone and patch size, we denote them by DINOv2-S, DINOv2-B, DINOv2-L, and DINOv2-g, by abuse of notation. The largest variant, DINOv2-g, is pretrained from scratch on $142$M images from the LVD-142M dataset, while the others are distilled from DINOv2-g using knowledge distillation~\citep{hinton2015knowledgedistillation,duval2023lowcompute}. Their characteristics are given in~\cref{tab:dinov2_variants}, along with the link to the checkpoints used in our experiments.
\begin{table}[!h]
    \centering
        \caption{\textbf{Characteristics of DINOv2 variants~\citep{oquab2024dinov2}.} We provide the number of layers, the embedding dimension, the FFN hidden dimension, the number of attention heads, the patch size, the number of parameters, and the link to the HuggingFace checkpoint.}
        \resizebox{\linewidth}{!}{
            \begin{NiceTabular}{lcccccccc}
            \toprule
            Model & Backbone & Layers & Emb. dim. & FFN dim. & Attn. heads & Patch size & Parameters & Checkpoint \\
            \midrule
            DINOv2-S & ViT-S/14  & 12 & 384  & 1536  & 6 & 14 & 22M  & \href{https://huggingface.co/facebook/dinov2-small}{\texttt{dinov2-small}}\\
            DINOv2-B & ViT-B/14 & 12 & 768  & 3072 & 12 & 14 & 86M  & \href{https://huggingface.co/facebook/dinov2-base}{\texttt{dinov2-base}}\\
            DINOv2-L & ViT-L/14 & 24 & 1024 & 4096 & 16 & 14 & 307M & \href{https://huggingface.co/facebook/dinov2-large}{\texttt{dinov2-large}}\\
            DINOv2-g & ViT-g/16  & 40 & 1536 & 4096 & 24 & 14 & 1.1B & \href{https://huggingface.co/facebook/dinov2-giant}{\texttt{dinov2-giant}}\\
            \bottomrule
            \end{NiceTabular}
        }
    \label{tab:dinov2_variants}
\end{table}
\paragraph{DINOv3.} Similarly, we consider $4$ variants of DINOv3 models~\citep{simeoni2025dinov3}, also pretrained in discriminative self-supervised fashion, with $22$M, $86$M, $307$M, and $7$B parameters. The patch size is fixed at $16$, and the input images have a resolution $224$ for all models. The corresponding backbones are referred to as ViT-S/16, ViT-B/16, ViT-L/16, and ViT-7B/16, respectively. Akin to DINOv2, we denote them by DINOv3-S, DINOv3-B, DINOv3-L, and DINOv3-7B, by abuse of notation. The largest variant, DINOv3-7B, is pretrained from scratch on $1.7$B images from the LVD-1689M dataset, while the others are distilled from DINOv3-7B using knowledge distillation~\citep{hinton2015knowledgedistillation,duval2023lowcompute}. Their characteristics are given in~\cref{tab:dinov3_variants}, along with the link to the checkpoints used in our experiments.
\begin{table}[!h]
    \centering
        \caption{\textbf{Characteristics of DINOv3 variants~\citep{simeoni2025dinov3}.} We provide the number of layers, the embedding dimension, the FFN hidden dimension, the number of attention heads, the patch size, the number of parameters, and the link to the HuggingFace checkpoint.}
        \resizebox{\linewidth}{!}{
            \begin{NiceTabular}{lcccccccc}
            \toprule
            Model & Backbone & Layers & Emb. dim. & FFN dim. & Attn. heads & Patch size & Parameters & Checkpoint \\
            \midrule
            DINOv3-S & ViT-S/16  & 12 & 384  & 1536  & 6 & 16 & 22M  & \href{https://huggingface.co/facebook/dinov3-vits16-pretrain-lvd1689m}{\texttt{dinov3-small}}\\
            DINOv3-B & ViT-B/16 & 12 & 768  & 3072 & 12 & 16 & 86M  & \href{https://huggingface.co/facebook/dinov3-vitb16-pretrain-lvd1689m}{\texttt{dinov3-base}}\\
            DINOv3-L & ViT-L/16 & 24 & 1024 & 4096 & 16 & 16 & 307M & \href{https://huggingface.co/facebook/dinov3-vitl16-pretrain-lvd1689m}{\texttt{dinov3-large}}\\
            DINOv3-7B & ViT-7B/16  & 40 & 4096 & 8192 & 32 & 16 & 7B & \href{https://huggingface.co/facebook/dinov3-vit7b16-pretrain-lvd1689m}{\texttt{dinov3-7B}}\\
            \bottomrule
            \end{NiceTabular}
        }
    \label{tab:dinov3_variants}
\end{table}
\paragraph{AIMv2.} Finally, we consider $4$ variants of autoregressive AIMv2 models~\citep{fini2025aimv2} with $307$M, $632$M, $1.2$B, and $2.7$B parameters, denoted by AIMv2-L, AIMv2-H, AIMv2-1B, and AIMv2-3B, respectively. Note that contrary to the iteration of autoregressive image models~\citep[AIM]{elnouby2024aim}, the AIMv2 family is multimodal and consists of a vision backbone coupled with a multimodal decoder. Our experiments focus only on the vision part. The patch size is fixed at $14$, and the input images have a resolution of $224$ for all models. All variants were pretrained from scratch on a total of $12$B image-text pairs from a combination of public and private datasets. Their characteristics are given in~\cref{tab:aimv2_variants}, along with the link to the checkpoints used in our experiments.
\begin{table}[!h]
    \centering
        \caption{\textbf{Characteristics of AIMv2 variants~\citep{fini2025aimv2}.} We provide the number of layers, the embedding dimension, the FFN hidden dimension, the number of attention heads, the patch size, the number of parameters, and the link to the HuggingFace checkpoint.}
        \resizebox{\linewidth}{!}{
            \begin{NiceTabular}{lccccccc}
            \toprule
            Model & Layers & Emb. dim. & FFN dim. & Attn. heads & Patch size & Parameters & Checkpoint \\
            \midrule
            AIMv2-L   & 24 & 1024  & 1536  & 6 & 14 & 307M  & \href{https://huggingface.co/apple/aimv2-large-patch14-224}{\texttt{aimv2-large}}\\
            AIMv2-H & 24 & 1536  & 3072 & 12 & 14 & 632M  & \href{https://huggingface.co/apple/aimv2-huge-patch14-224}{\texttt{aimv2-huge}}\\
            AIMv2-1B & 24 & 2048 & 4096 & 16 & 14 & 1.2B & \href{https://huggingface.co/apple/aimv2-1B-patch14-224}{\texttt{aimv2-1B}}\\
            AIMv2-3B & 24 & 3072 & 8192 & 32 & 14 & 2.7B & \href{https://huggingface.co/apple/aimv2-3B-patch14-224}{\texttt{aimv2-3B}}\\
            \bottomrule
            \end{NiceTabular}
        }
    \label{tab:aimv2_variants}
\end{table}

\paragraph{Datasets.} The vision experiments are conducted on a diverse set of $18$ image classification benchmarks: Aircraft~\citep{maji2013aircraft}; Cifar10 and Cifar100~\citep{krizhevsky2009cifar}; DTD~\citep{cimpoi2014dtd}; EuroSAT~\citep{helber2019eurosat}; Flowers102~\citep{nilsback2008flowers102}; Food101~\citep{bossard2014food101}; $3$ variants from Cifar10-C with severity $5$~\citep{hendrycks2018imagenetc}, namely Contrast, Motion Blur, Speckle Noise; Pets~\citep{parkhi2012pets}; the $5$ domains of DomainNet~\citep{peng2019domainnet}, namely Clipart, Infograph, Quickdraw, Painting, Sketch; and $2$ datasets from \textsc{Wilds}~\citep{koh2021wilds}, namely iWildCam and RxRx1. $15$ variants of the Cifar10-C benchmark~\citep{hendrycks2018imagenetc} are also used for experiments with controlled distribution shifts. Each variant consists of a corrupted version of the Cifar10 test set, with a type of corruption among Contrast, Gaussian Noise, Motion Blur, Snow, and Speckle Noise, and a level of corruption severity among $1$, $3$, and $5$, from the lowest to the strongest. The datasets' characteristics are given in~\cref{tab:image_classification}.
\begin{table}[!h]
    \centering
    \caption{\textbf{Characteristics of our $18$ image classification benchmarks.} We provide the number of training and test images, as well as the number of classes. Contrast, Motion Blur, and Speckle Noise are corrupted variants of the Cifar10 test set with a corruption severity of $5$, from the Cifar10-C benchmark~\citep{hendrycks2018imagenetc}, Clipart, Infograph, Painting, Quickdraw, and Sketch are domains from the DomainNet benchmark~\citep{peng2019domainnet}. Cifar10 $\rightarrow$ Cifar10-C indicates that the linear probe is trained on the training set of Cifar10 and evaluated on Cifar10-C, i.e., on a corrupted version of the Cifar10 test set.}
    \label{tab:image_classification}
        \begin{NiceTabular}{llrrr}
        \toprule
        Dataset & Reference & Train & Test & Classes \\
        \midrule
        Aircraft & \citet{maji2013aircraft}         & 6\,667  & 3\,333  & 100 \\
        Cifar10 & \citet{krizhevsky2009cifar}        & 50\,000 & 10\,000 & 10  \\
        Cifar100 &\citet{krizhevsky2009cifar}       & 50\,000 & 10\,000 & 100 \\
        Clipart & \citet{peng2019domainnet} & 39\,066  & 9\,767 & 345 \\
        Contrast & \citet{hendrycks2018imagenetc} & 8\,000  & 2\,000 & 10  \\
        DTD & \citet{cimpoi2014dtd}                  & 1\,880  & 1\,880  & 47  \\
        EuroSAT & \citet{helber2019eurosat}          & 21\,600 & 5\,400  & 10  \\
        Flowers102 & \citet{nilsback2008flowers102}  & 1\,020  & 6\,149  & 102 \\
        Food101 & \citet{bossard2014food101}         & 75\,750 & 25\,250 & 101 \\
        Infograph & \citet{peng2019domainnet} & 42\,560  & 10\,641 & 345 \\
        iWildCam & \citet{koh2021wilds} & 129\,809  & 42\,791 & 182 \\
        Motion Blur & \citet{hendrycks2018imagenetc} & 8\,000  & 2\,000 & 10  \\
        Painting & \citet{peng2019domainnet} & 60\,607  & 15\,152  & 345 \\
        Pets & \citet{parkhi2012pets} & 3\,680  & 3\,669  & 37  \\
        Quickdraw & \citet{peng2019domainnet} & 138\,000  & 34\,500 & 345 \\
        RxRx1 & \citet{koh2021wilds} & 40\,612  & 34\,432  & 1\,139 \\
        Sketch & \citet{peng2019domainnet} & 56\,308 & 14\,078 & 345 \\
        Speckle Noise & \citet{hendrycks2018imagenetc} & 8\,000  & 2\,000 & 10  \\
        Cifar10 $\rightarrow$ Cifar10-C & \citet{hendrycks2018imagenetc} & 50\,000 & 10\,000 & 10 \\
        \bottomrule
        \end{NiceTabular}
\end{table}

\paragraph{Data preprocessing.} We follow the preprocessing protocol of~\citet[Section 7.2]{oquab2024dinov2}: we apply a $224 \times 224$ resizing, matching our models' native resolution, before center cropping images. All images are then normalized using the ImageNet1k~\citep{deng2009imagenet} statistics. It ensures images with mean $[0.485, 0.456, 0.406]$ and standard deviation $[0.229, 0.224, 0.225]$. Note that no data augmentation is used.  They are indeed not necessary since training is not stochastic: the feature extraction is deterministic, and the classifier is fitted once over all the frozen training features and evaluated once on the corresponding test features (data augmentation would, however, be required with an SGD-based linear probing~\citep[see][]{marouani2026revisiting}). For datasets that do not have predefined training and test sets, e.g., EuroSAT, or datasets from Cifar10-C and DomainNet, we manually create \emph{deterministic} training and test sets following an $80\%$-$20\%$ split. The deterministic part is crucial to ensure no data leakage. 

\paragraph{Evaluation.} We follow the linear probing protocol of DINOv2\footnote{\label{fn:dinov2}Code available at \href{https://github.com/facebookresearch/dinov2/blob/main/dinov2/eval/log_regression.py}{\texttt{facebookresearch/dinov2/blob/main/dinov2/eval/log\_regression.py}}} by training a logistic regression on frozen features for at most $1000$ iterations, with L-BFGS solver, an $\ell_2$ penalty, and a tolerance of $1\mathrm{e}{-4}$. In our experiments, we use a regularization strength $C=1.0$, and preprocess representations by removing the mean and scaling to unit variance with a standard scaler. We ensure efficient evaluation by making use of the GPU-friendly implementation of the \code{StandardScaler} and \code{LogisticRegression} functions from the CuML library~\citep{raschka2020cuml}. For standard evaluation, training is done on the training set of each dataset, and evaluation is done on the corresponding test set. The reported performance is the top-1 test accuracy. For the robustness analysis on Cifar10-C, akin to DINOv3 domain generalization experiments~\citep[see][Section 6.2.1]{simeoni2025dinov3}, training is done on the training set of Cifar10, and the learned classifier is then evaluated independently on the $15$ corrupted versions of the Cifar10 test set described in the \say{Datasets} paragraph above.

\subsection{Large language models}
\label[secinapp]{app:llm_implem}
This section provides details on LLMs' transformer representations and the linear probing setup.

\subsubsection{Transformer representations}
\label[secinapp]{app:llm_transformer_representation}
The \texttt{MMTEB} benchmark~\citep {enevoldsen2025mmteb}, which expands the \texttt{MTEB} benchmark~\citep{muennighoff2023mteb} with additional domains, tasks, and languages, requires defining an \texttt{encode} method for a given LLM, which is then used to evaluate representations over a predefined set of tasks. As such, we set the desired (layer, module) pair via a \texttt{set\_probing} method, use PyTorch hooks and pre-hooks in the \texttt{encode} to extract representations during the forward pass, and recover the results over the corresponding set of tasks. In our code, all LLMs inherit from the abstract base \texttt{BaseLanguageModel} displayed in~\cref{fig:base_llm}. Abstract and hook methods are similar to \texttt{BaseVisionModel} (not shown due to space limitations), with an additional \texttt{\_load\_tokenizer} to load the tokenizer. The feature extraction is done with the \texttt{encode} method. Similarly to VFMs (see~\cref{app:vfm_transformer_representation}), the most important abstract method to implement is \texttt{block\_adapter}, which defines the $8$ representations to extract within each transformer block and inherits from the base class \texttt{TransformerBlockAdapter} displayed in~\cref{fig:base_block}. We provide below the implementation details on the LLMs used in our experiments.

\paragraph{Qwen3.} The Qwen3 architecture~\citep{yang2025qwen3} consists of a succession of transformer decoders with RMSnorms~\citep{zhang2019rms}, Rotary Positional Embeddings~\citep[RoPE,][]{su2024rope} and a gated feedforward network with SwiGLU activation~\citep{shazeer2020glu}. The checkpoints used in our experiments follow the HuggingFace transformers~\citep{wolf2020hftransformers} implementation available at \href{https://github.com/huggingface/transformers/blob/main/src/transformers/models/qwen3/modeling_qwen3.py}{\code{transformers/models/qwen3/modeling\_qwen3.py}}.

\paragraph{Llama3.} The Llama3 architecture~\citep{grattafiori2024llama3} largely follows the Llama original one~\citep{touvron2023llama}. Similarly to Qwen3, it consists of a succession of transformer decoders with RMSnorms~\citep{zhang2019rms}, Rotary Positional Embeddings~\citep[RoPE,][]{su2024rope} and a gated feedforward network with SwiGLU activation~\citep{shazeer2020glu}. The checkpoints used in our experiments follow the HuggingFace transformers~\citep{wolf2020hftransformers} implementation available at \href{https://github.com/huggingface/transformers/blob/main/src/transformers/models/llama/modeling_llama.py}{\code{transformers/models/llama/modeling\_llama.py}}.

\begin{figure}[H]
\begin{center}
\begin{minipage}{\linewidth}
\begin{lstlisting}[language=Python, frame=single]
class BaseLanguageModel(nn.Module, ABC):
    r"""Abstract base class for LLMs. Abstract and hook methods (not shown) are akin to BaseVisionModel."""

    def __init__(self):
        super().__init__()
        self.device: str
        self.model: PreTrainedModel
        self.tokenizer: PreTrainedTokenizerBase

        # Default probing done on the last residual stream
        self.layer: int = -1
        self.module: str = "rc2"
        
    @torch.inference_mode()
    def _get_hidden_states(self, input_ids: torch.Tensor, attention_mask: torch.Tensor) -> torch.Tensor:
        r"""Single forward pass recovering the sequence of tokens of shape (N, L, E) after the specified layer and module."""
        embeddings: dict[str, torch.Tensor] = {}
        handles: list[torch.utils.hooks.RemovableHandle] = []

        # Recover embedding from the specified layer and module
        key = f"block_{self.layer}_{self.module}"
        adapter = self.block_adapter(self.blocks[self.layer])
        module_obj = adapter.module_map()[self.module]

        if self.module in adapter.pre_hooks:
            handles.append(module_obj.register_forward_pre_hook(self._make_pre_hook(embeddings=embeddings, key=key)))
        else:
            handles.append(module_obj.register_forward_hook(self._make_hook(embeddings=embeddings, key=key)))

        try:
            self.model(input_ids=input_ids, attention_mask=attention_mask)
        finally:
            for handle in handles:
                handle.remove()

        return embeddings[key]

    def _encode_sentences(self, sentences: list[str], batch_size: int = 256) -> torch.Tensor:
        r"""Encode a flat list of sentences into mean-pooled embeddings."""
        all_embeddings = []
        for i in range(0, len(sentences), batch_size):
            text = sentences[i : i + batch_size]

            # Recover encoding, input_ids and padding tokens (attention_mask)
            encoding = self.tokenizer(text=text, padding=True, truncation=True, return_tensors="pt").to(self.device)
            input_ids = encoding.input_ids
            attention_mask = encoding.attention_mask

            # Recover hidden states
            hidden_states = self._get_hidden_states(input_ids=input_ids, attention_mask=attention_mask)

            # Recover pooled embeddings
            embeddings = self.pooling(
                hidden_states=hidden_states,
                attention_mask=attention_mask,
                input_ids=input_ids,
                special_token_ids=self.special_token_ids,
            )
            all_embeddings.append(embeddings)

        return torch.cat(all_embeddings, dim=0)

    def encode(self, inputs: DataLoader[BatchedInput] | list[str], *, task_metadata: TaskMetadata, hf_split: str, hf_subset: str, prompt_type: PromptType | None = None, **kwargs: Unpack[EncodeKwargs]) -> Array:
        r"""Encode inputs compatible with MTEB v2 EncoderProtocol."""
        if isinstance(inputs, DataLoader):
            all_embeddings = []
            for batch in inputs:
                batch_embeddings = self._encode_sentences(sentences=batch["text"])
                all_embeddings.append(batch_embeddings)
            return torch.cat(all_embeddings, dim=0)
        elif isinstance(inputs, list):
            return self._encode_sentences(inputs)
        else:
            raise ValueError("Unsupported input type. Expected DataLoader or list of strings.")
            
    def set_probing(self, layer: int | None, module: str | None) -> None:
        r"""Set the layer and module to probe for embedding extraction."""
        if layer is not None:
            if layer >= self.n_layers:
                raise ValueError(f"Layer index {layer} is out of range for a model with {self.n_layers} layers.")
            self.layer = layer
        if module is not None:
            if module not in TRANSFORMER_MODULES:
                raise ValueError(f"Invalid module '{module}'. Options are {TRANSFORMER_MODULES}.")
            self.module = module

\end{lstlisting}
\end{minipage}
\end{center}
\caption{\texttt{BaseLanguageModel} implementation.}
\label{fig:base_llm}
\end{figure}

\paragraph{Gemma3.} The Gemma3 architecture~\citep{kammath2025gemma3} is akin to the Qwen3 and Llama3 ones, except that the gated FFN uses an approximated GeGLU activation~\citep{shazeer2020glu} and that both the input and output of the attention and feedforward blocks are normalized. For a consistent evaluation, the LN1 and LN2 representations correspond to the pre-norms.

\paragraph{Feature extraction.} The feature extraction follows the official implementation available at \url{https://github.com/embeddings-benchmark/mteb/}. Documents are embedded with the \texttt{encode} method, and the resulting representations are fed to a logistic regression.

\subsubsection{Probing setup}
\label[secinapp]{app:llm_probing_setup}
\paragraph{Qwen3.} We conduct experiments on dense general-purpose Qwen3 models~\citep{yang2025qwen3} with a vocabulary size of $150$K. We consider $3$ variants with $1.7$B, $4$B and $8$B parameters, referred to as Qwen3 1.7B, Qwen3 4B, and Qwen3 8B, respectively. All variants were pretrained from scratch on $36$T tokens from a multilingual dataset with 119 languages and dialects, then post-trained with a hybrid thinking/non-thinking mode. Their characteristics are given in~\cref{tab:qwen3_variants}, along with the link to the checkpoints used in our experiments.
\begin{table}[!h]
    \centering
        \caption{\textbf{Characteristics of Qwen3 variants~\citep{yang2025qwen3}.} We provide the number of layers, the embedding dimension, the FFN hidden dimension, the number of attention heads, the context length, the number of parameters, and the link to the HuggingFace checkpoint.}
        \resizebox{\linewidth}{!}{
            \begin{NiceTabular}{lccccccc}
            \toprule
            Model & Layers & Emb. dim. & FFN dim. & Attn. heads & Context length & Parameters & Checkpoint \\
            \midrule
            Qwen3 1.7B  & 28 & 2048  & 6144 & 16 & 32K & 1.72B  & \href{https://huggingface.co/Qwen/Qwen3-1.7B}{\texttt{Qwen3-1.7B}}\\
            Qwen3 4B & 36 & 2560 & 9728 & 32  & 128K & 4.02B & \href{https://huggingface.co/Qwen/Qwen3-4B}{\texttt{Qwen3-4B}} \\
            Qwen3 8B  & 36 & 4096 & 12288 & 32 & 128K & 8.19B & \href{https://huggingface.co/Qwen/Qwen3-8B}{\texttt{Qwen3-8B}} \\
            \bottomrule
            \end{NiceTabular}
        }
    \label{tab:qwen3_variants}
\end{table}

\paragraph{Llama3.} We conduct experiments on dense general-purpose Llama3 models~\citep{grattafiori2024llama3} with a vocabulary size of $128$K. We consider $3$ variants with $1$B, $3$B and $8$B parameters, referred to as Llama3 1B, Llama3 3B, and Llama3 8B, respectively\footnote{The official model variant is Llama3.2 for the $1$B and $3$B sizes, and Llama3.1 for the $8$B one}. The 8B variant was pretrained from scratch on $\sim 15$T tokens. The 1B and 3B variants are pruned from the 8B model and trained on up to 9T tokens with knowledge distillation from the 8B and 70B models. Their characteristics are given in~\cref{tab:llama3_variants}, along with the link to the checkpoints used in our experiments.
\begin{table}[!h]
    \centering
        \caption{\textbf{Characteristics of Llama3 variants~\citep{grattafiori2024llama3}.} We provide the number of layers, the embedding dimension, the FFN hidden dimension, the number of attention heads, the context length, the number of parameters, and the link to the HuggingFace checkpoint.}
        \resizebox{\linewidth}{!}{
            \begin{NiceTabular}{lccccccc}
            \toprule
            Model & Layers & Emb. dim. & FFN dim. & Attn. heads & Context length & Parameters & Checkpoint \\
            \midrule
            Llama3 1B  & 16 & 2048  & 8192 & 32 & 128K & 1.24B  & \href{https://huggingface.co/meta-llama/Llama-3.2-1B}{\texttt{Llama3.2 1B}}\\
            Llama3 3B & 28 & 3072 & 9728 & 24  & 128K & 3.21B & \href{https://huggingface.co/meta-llama/Llama-3.2-3B}{\texttt{Llama3.2 3B}} \\
            Llama3 8B  & 32 & 4096 & 14336 & 32 & 128K & 8.03B & \href{https://huggingface.co/meta-llama/Llama-3.1-8B}{\texttt{Llama3.1 8B}} \\
            \bottomrule
            \end{NiceTabular}
        }
    \label{tab:llama3_variants}
\end{table}

\paragraph{Gemma3.} We conduct experiments on dense general-purpose Gemma3 models~\citep{kammath2025gemma3} with a vocabulary size of $262$K. We consider $3$ variants with $1$B, $4$B and $12$B parameters, referred to as Gemma3 1B, Gemma3 4B, and Gemma3 12B, respectively. All variants were pretrained from scratch on $2$T, $4$T, and $12$T multilingual tokens. Their characteristics are given in~\cref{tab:gemma3_variants}, along with the link to the checkpoints used in our experiments.
\begin{table}[!h]
    \centering
        \caption{\textbf{Characteristics of Gemma3 variants~\citep{kammath2025gemma3}.} We provide the number of layers, the embedding dimension, the FFN hidden dimension, the number of attention heads, the number of parameters, and the link to the HuggingFace checkpoint.}
        \resizebox{\linewidth}{!}{
            \begin{NiceTabular}{lccccccc}
            \toprule
            Model & Layers & Emb. dim. & FFN dim. & Attn. heads & Context length & Parameters & Checkpoint \\
            \midrule
            Gemma3 1B  & 26 & 1152  & 6912 & 4 & 128K & 1.00B  & \href{https://huggingface.co/google/gemma-3-1b-pt}{\texttt{gemma-3-1b-pt}}\\
            Gemma3 4B & 34 & 2560 & 10240 & 8  & 128K & 4.30B & \href{https://huggingface.co/google/gemma-3-4b-pt}{\texttt{gemma-3-4b-pt}} \\
            Gemma3 12B  & 48 & 3840 & 15360 & 16 & 128K & 12.19B & \href{https://huggingface.co/google/gemma-3-12b-pt}{\texttt{gemma-3-12b-pt}} \\
            \bottomrule
            \end{NiceTabular}
        }
    \label{tab:gemma3_variants}
\end{table}

\paragraph{Tasks.} We consider classification tasks from the \texttt{MTEB(Multilingual)} benchmark with at most $8\,192$ documents. This leads to a total of $24$ tasks spanning $14$ domains and $31$ languages. The characteristics of the tasks are given in~\cref{tab:mteb_tasks}.
\begin{table}[t]
\small
\centering
\caption{\textbf{Characteristics of \texttt{MTEB(Multilingual)} classification tasks.} Experiments are conducted on the $24$ tasks below, spanning a diverse set of languages and domains. The column names follow~\citet{enevoldsen2025mmteb}, and the list of ISO 639 codes can be found at \url{https://en.wikipedia.org/wiki/List_of_ISO_639_language_codes}.}
\label{tab:mteb_tasks}
\resizebox{\linewidth}{!}{
\begin{NiceTabular}{lllllr}
\toprule
Task & Reference & Languages & Domains & Sample creation & Samples \\
\midrule
BulgarianStoreReviewSentimentClassification & \citet{trifonova2018bulgarian} & bul & Reviews, Written & found, human-annotated & 182 \\
DalajClassification & \citet{volodina2021dalaj} & swe & Non-fiction, Written & created, expert-annotated & 888 \\
DBpediaClassification & \citet{zhang2015dbpedia} & eng & Encyclopaedic, Written & found, derived & 2\,048 \\
EstonianValenceClassification & \citet{pajupuu2023estonian}& est & News, Written & found, human-annotated & 818 \\
FinancialPhrasebankClassification & \citet{malo2014financial}& eng & News, Written, Financial & found, expert-annotated & 2\,264 \\
GujaratiNewsClassification & - & guj & News, Written & found, derived & 1\,318 \\
IndonesianIdClickbaitClassification & \citet{willaim2020indonesian} & ind & News, Written & found, expert-annotated & 2\,048 \\
IsiZuluNewsClassification & \citet{madodonga2023izi}& zul & News, Written & found, human-annotated & 752 \\
ItaCaseholdClassification & \citet{licari2023itacasehold} & ita & Legal, Government, Written & found, expert-annotated & 221 \\
KorSarcasmClassification & \citet{kim2019korsarcasm} & kor & Social, Written & found, expert-annotated & 2\,048 \\
KurdishSentimentClassification & \citet{badawi2024kurdish} & kur & Web, Written & found, derived & 1\,987 \\
NepaliNewsClassification & \citet{arora2020nepali} & nep & News, Written & found, derived & 2\,048 \\
NusaX-senti & \citet{winata2023nusax} & ace, ban, bbc, \dots & Reviews, Web, Social, \dots & found, expert-annotated & 4\,800 \\
PAC &\citet{augustyniak2022pac} & pol & Legal, Written & - & 3\,453 \\
PoemSentimentClassification &\citet{sheng2020poem} & eng & Reviews, Written & found, human-annotated & 209 \\
PolEmo2.0-OUT & - & pol & Written, Social & - & 494 \\
PunjabiNewsClassification & \citet{kunchukuttan2020punjabi} & pan & News, Written & found, derived & 157 \\
ScalaClassification & \citet{nielsen2023scala}& dan, nno, nob, \dots & Fiction, News, Non-fiction, \dots & created, human-annotated & 8\,192 \\
SentimentAnalysisHindi & \citet{parida2023sentiment} & hin & Reviews, Written & found, derived & 2\,048 \\
SinhalaNewsClassification & \citet{desilva2015sinhala} & sin & News, Written & found, derived & 2\,048 \\
SiswatiNewsClassification & \citet{madodonga2023izi}& ssw & News, Written & found, human-annotated & 80 \\
SwahiliNewsClassification & \citet{davis2020swahili} & swa & News, Written & found, derived & 2\,048 \\
TswanaNewsClassification & \citet{marivate2023tswana} & tsn & News, Written & found, derived & 487 \\
TweetTopicSingleClassification & \citet{antypas2022tweet} & eng & Social, News, Written & found, expert-annotated & 1\,693 \\
\bottomrule
\end{NiceTabular}
}
\end{table}

\paragraph{Evaluation.} The \texttt{MMTEB} evaluation protocol applies linear probing via a logistic regression on frozen features for at most $100$ iterations. It uses the \texttt{LogisticRegression} method from Sklearn~\citep{pedregosa2011sklearn}, implemented with the L-BFGS solver, an $\ell_2$ penalty, a regularization strength $C=1.0$, and a tolerance of $1\mathrm{e}{-4}$. Quoting~\citet{enevoldsen2025mmteb}, \say{a train set is constructed by sampling $n$ (8-16) samples for each label. If only a test set is available, a section is split off as a training set. Both sets are then embedded and used to train a logistic regression using a maximum of 100 iterations. Afterwards, performance metrics are calculated. For robustness, this process is repeated 10 times}.

\section{Additional experiments}
\label[secinapp]{app:add_exp}
In this section, we provide details on the experiments in the main paper and on additional experiments. 

\subsection{Vision foundation models}
\label[secinapp]{app:vfm_add_exp}
The results presented in this section relate to the experiments conducted on VFMs in~\cref{sec:vfm_exp}.

\subsubsection{Intermediate layers under distribution shift}
\label[secinapp]{app:shift}
\citet{skean2025layer} observed the benefits of intermediate layers in LLMs but not in VFMs, where performance increased towards the final layers, except for autoregressive models such as AIM~\citep{elnouby2024aim}. Recent work, however, suggests that intermediate representations can also be beneficial in non-autoregressive VFMs~\citep{bolya2025perception}, particularly under distribution shifts~\citep{uselis2025intermediate}. In practice, discrepancies between pretraining and downstream data can degrade performance by making the extracted features less informative~\citep{quionero2009shift}. To illustrate this effect, we consider a ViT-B~\citep{dosovitskiy2021vit} pretrained on ImageNet-21K and probe it before and after finetuning on downstream data. The experimental setup is described below.

\paragraph{Datasets.} We conduct experiments on Cifar10~\citep{krizhevsky2009cifar}; $2$ variants from Cifar10-C~\citep{hendrycks2018imagenetc} with severity $5$ (Contrast and Speckle Noise), and Flowers102~\citep{nilsback2008flowers102}. The preprocessing follows~\citet{dosovitskiy2021vit} and~\citet{kolesnikov2020bit}: for training data, we apply random cropping, a 224$\times$224 image resizing, and random horizontal flip. For validation and test data, the 224$\times$224 image resizing is applied before center cropping images. All images are normalized using the ImageNet~\citep{deng2009imagenet} statistics. It ensures images with mean $[0.485, 0.456, 0.406]$ and standard deviation $[0.229, 0.224, 0.225]$. For Cifar10-C datasets that do not have predefined training and test sets, we manually create \emph{deterministic} training and test sets following an $80\%$-$20\%$ split. The deterministic part is crucial to ensure no data contamination. 

\paragraph{Finetuning setup.} Our finetuning experiments follow the protocol from~\citet{dosovitskiy2021vit} with a resolution of $224 \times 224$. We optimize models with Stochastic Gradient Descent (SGD), a momentum of 0.9, no weight decay, a cosine learning rate decay, a batch size of $512$, and gradient clipping at norm $1$. For each pair of dataset-configuration, we perform a sweep over $4$ learning rates, as summarized in~\cref{tab:training_details}, and conduct $3$ runs with different seeds relative to network initialization and dataloaders. For each run, we monitor the training using a validation set ($20\%$ of the training set). The final performance is the test accuracy of the checkpoint that achieves the best validation accuracy. 

\begin{figure}[H]
    \centering
    \includegraphics[width=\linewidth]{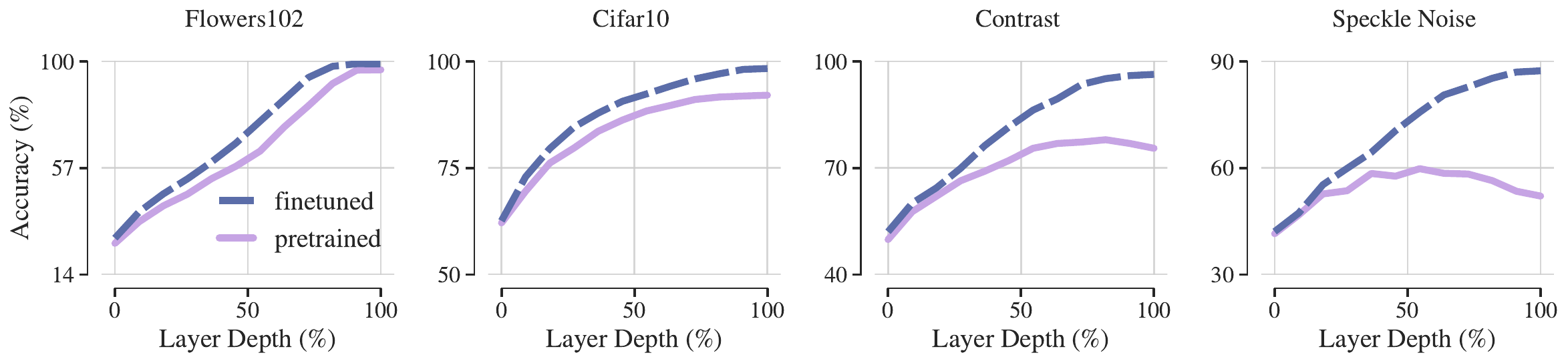}
    \caption{\textbf{Intermediate layers under distribution shift.} Evolution of the linear probing performance across layers of a ViT-B ($86$M parameters) pretrained on ImageNet-21K. The solid line is the accuracy on the pretrained model, and the dashed line is the accuracy on the same model after finetuning on the dataset at hand. Datasets are ordered by decreasing finetuning performance.}
    \label{fig:shift}
\end{figure}

\paragraph{Results.} In~\cref{fig:shift}, we order datasets by decreasing finetuning accuracy, using it as a rough proxy for increasing distribution shift:  $\text{99.15}_{\tiny \pm \text{0.05}}$ on Flowers102, $\text{99.02}_{\tiny \pm \text{0.02}}$ on Cifar10, $\text{97.23}_{\tiny \pm \text{0.18}}$ on Contrast, and $\text{89.58}_{\tiny \pm \text{0.43}}$ on Speckle Noise. We can see that the final-layer residual stream representations remain strong on near in-distribution datasets, while intermediate representations become increasingly beneficial under distribution shift. After finetuning, in contrast, performance increases monotonically towards the final layers. This shows that, akin to LLMs, intermediate representations can be effective in VFMs, and that their usefulness depends on the discrepancy between pretraining and downstream data. 

\begin{table}[!h]
    \centering
       \caption{\textbf{Finetuning setup}. We report the optimizer, batch size, training steps, and learning rates.}
        \begin{NiceTabular}{lcccc}
        \toprule
        Dataset & Optimizer & Batch size & Training steps & Learning rates $\eta$\\
        \midrule
        Cifar10 & SGD & 512 & $10000$ & \{$1\mathrm{e}{-3}$, $3\mathrm{e}{-3}$, $1\mathrm{e}{-2}$, $3\mathrm{e}{-2}$\} \\
        Contrast & SGD & 512&  $10000$ &\{$1\mathrm{e}{-3}$, $3\mathrm{e}{-3}$, $1\mathrm{e}{-2}$, $3\mathrm{e}{-2}$\} \\
        Speckle Noise & SGD & 512&  $10000$ &\{$1\mathrm{e}{-3}$, $3\mathrm{e}{-3}$, $1\mathrm{e}{-2}$, $3\mathrm{e}{-2}$\} \\
        Flowers102 & SGD & 512&  $5000$ & \{$1\mathrm{e}{-3}$, $3\mathrm{e}{-3}$, $1\mathrm{e}{-2}$, $3\mathrm{e}{-2}$\} \\
        \bottomrule
        \end{NiceTabular}
    \label{tab:training_details}
\end{table}

\subsubsection{Gains across pretraining scales}
\label[secinapp]{app:pretraining_scale}
Following~\citet{hoffmann2022chinchilla}, we measure the pretraining scale of a model as the total number of (visual) tokens seen during pretraining. \cref{tab:vfm_pretraining_images} summarizes the number of images seen during pretraining, using the figures reported in the official paper and documentation of each model. For ViT models, the paper only reports the number of epochs ($90$). Since the batch size is $4096$ and pretraining is done on ImageNet-21K, which contains $14$M images, this leads to $\sim 307\,617$ training iterations.

\begin{table}[!h]
    \centering
    \caption{\textbf{Number of images seen during pretraining.} We report the batch size, number of training iterations, and total number of images seen during pretraining for each model, along with the official reference for the figures.}
    \begin{NiceTabular}{lcccc}
     \toprule
     Model & Reference  & Batch size & Iterations & Pretraining images \\
     \midrule
     ViT & \citet{dosovitskiy2021vit} & 4\,096 & 307\,617 & 1\,259\,999\,232 \\
     DINOv2 & \citet{oquab2024dinov2} & 3\,072 & 625\,000 & 1\,920\,000\,000 \\
     DINOv3 & \citet{simeoni2025dinov3} & 4\,096 & 1\,000\,000 & 4\,096\,000\,000 \\
     AIMv2 & \citet{fini2025aimv2} & 8\,192 & 1\,500\,000 & 12\,288\,000\,000 \\
    \bottomrule
    \end{NiceTabular}
    \label{tab:vfm_pretraining_images}
\end{table}
Then, we need to recover the number of visual tokens per image. For ViT, this only depends on the image resolution and patch size. Taking into account the \texttt{CLS} token, each image contains $\frac{HW}{P^2} + 1$ tokens, where $H$, $W$, and $P$ denote the height, width, and patch size, respectively. For AIMv2, the computation is similar, without the $+1$, since autoregressive models~\citep{fini2025aimv2,elnouby2024aim} do not use a \texttt{CLS} token. For DINOv2 and DINOv3, we need to take into account the multi-crop strategy of~\citet{caron2020unsupervised}: for each image, the student model sees $2$ global crops and $8$ local crops, each contributing $\frac{HW}{P^2}$ tokens, with the height and width of the crop. On top of these, each crop includes a \texttt{CLS} token for DINOv2, and a \texttt{CLS} token and $4$ registers for DINOv3, leading to an overhead of $10$ and $50$ tokens per image, respectively. Recalling that images have resolution $224 \times 224$, the number of tokens per image can thus be summarized as follows:
\begin{table}[!h]
    \centering
    \caption{\textbf{Number of visual tokens per image.} We report for each model the patch size, image resolution, global and local crops resolution (when supported) and the corresponding amount of visual tokens per image.}
    \resizebox{\linewidth}{!}{
    \begin{NiceTabular}{lcccccc}
     \toprule
     Model & Size & Patch size & Image resolution & Global crops (nb. - resolution) & Local crops & Nb. of tokens per image \\
     \midrule
     \multirow{3}{*}{ViT} & B & 16 & $224\times 224$ &-&-&197\\
     & L &16 & $224\times 224$ &-&-&197\\
     & H & 14 & $224\times 224$ &-&-&257\\
     \midrule
     \multirow{4}{*}{DINOv2} & S & 14 & $224\times 224$ &2 - $224\times224$&8-$98\times98$&914\\
     & B &14 & $224\times 224$ &2 - $224\times224$&8-$98\times98$&914\\
     & L & 14 & $224\times 224$ &2 - $224\times224$&8-$98\times98$&914\\
     & g & 14 & $224\times 224$ &2 - $224\times224$&8-$98\times98$&914\\
     \midrule
     \multirow{4}{*}{DINOv3} & S & 16 & $224\times 224$ &2 - $256\times256$&8-$112\times112$&954\\
     & B &16 & $224\times 224$ &2 - $256\times256$&8-$112\times112$&954\\
     & L & 16 & $224\times 224$ &2 - $256\times256$&8-$112\times112$&954\\
     & $7$B & 16 & $224\times 224$ &2 - $256\times256$&8-$112\times112$&954\\
     \midrule
     \multirow{4}{*}{AIMv2} & L & 14 & $224\times 224$ &-&-&256\\
     & H &14 & $224\times 224$ &-&-&256\\
     & $1$B & 14 & $224\times 224$ &-&-&256\\
     & $3$B & 14 & $224\times 224$ &-&-&256\\
    \bottomrule
    \end{NiceTabular}
    }
    \label{tab:vfm_visual_tokens}
\end{table}
Putting~\cref{tab:vfm_pretraining_images,tab:vfm_visual_tokens} together provides the number of visual tokens seen during pretraining: $0.25$T for ViT, $1.75$T for DINOv2, $3.90$T for DINOv3, and $3.15$T for AIMv2. While the exact number varies within the ViT family due to the larger sequence length of the Huge variant, the order of magnitude is the same ($0.25$T compared to $0.32$T).

\subsubsection{Gain across model families}
\begin{figure}[H]
    \centering
    \includegraphics[width=\linewidth]{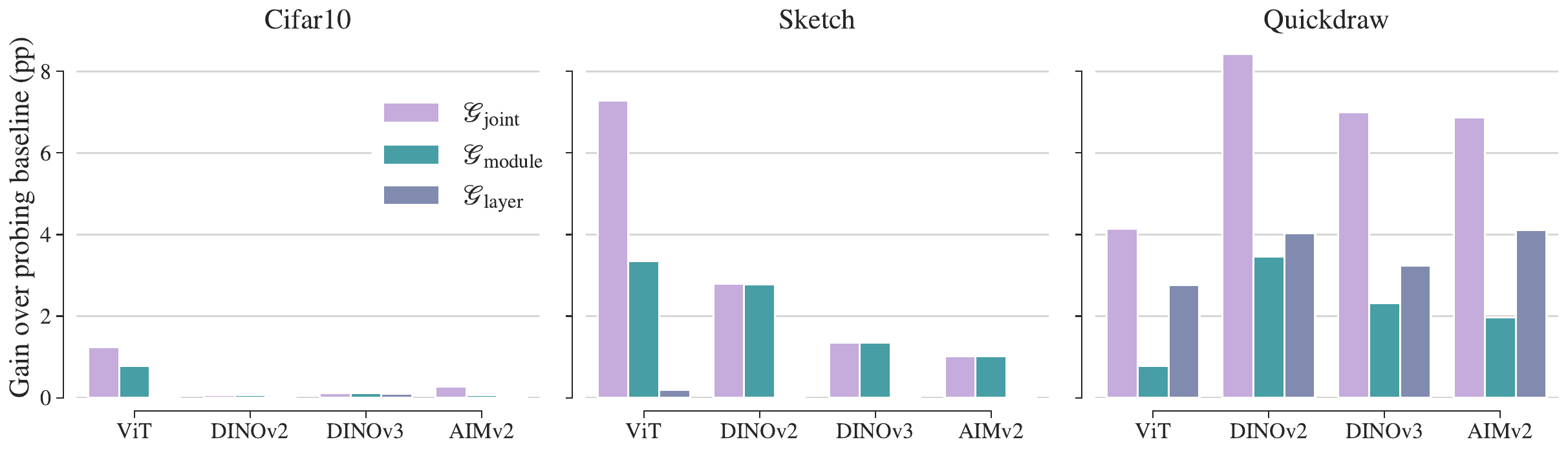}
    \caption{\textbf{Gain over model families.} The three types of gains ($\layer$, $\module$, and $\oracle$) are displayed on the representative datasets spanning low, medium, and high transfer difficulty (Cifar10, Sketch, Quickdraw). Results are reported on the Large variant ($307$M parameters) of each family. The magnitude of gains increases with transfer difficulty, consistent with~\cref{fig:transfer_difficulty}.}
    \label{fig:model_family_vision}
\end{figure}

In this section, we compare the improvement over the probing baseline across model families. \cref{fig:model_family_vision} displays the gains of searching over layers alone ($\layer$), modules alone ($\module$), and both jointly ($\oracle$) on the Large variant ($307$M parameters) of each family. Results are presented on Cifar10, Sketch, and Quickdraw, in increasing order of transfer difficulty (see~\cref{fig:transfer_difficulty}). We can see that on Cifar10, gains are small for all models. By contrast, as transfer difficulty increases, the benefits of layers and modules become more salient. Notably, the choice of modules matters on Sketch, especially for ViT, which has the smallest pretraining scale. Moreover, intermediate layers substantially outperform the final layer for all model families on Quickdraw.

\subsubsection{Module comparison}
\label[secinapp]{app:vfm_module_comparison}
In~\cref {tab:vfm_module_comparison}, we show the percentage of VFM evaluations where a given module belongs to the layer-module pair that achieves the oracle gain $\oracle$. It is referred to as the winning frequency. We obtain: 
\begin{table}[!h]
\centering
    \caption{\textbf{Module comparison.} Percentage of evaluations, out of a total of $270$, for which a given module belongs to the layer-module pair achieving the oracle gain $\oracle$.}
    \label{tab:vfm_module_comparison}
    \begin{NiceTabular}{lcccccccc}
    \CodeBefore
    \columncolor{tablegray}{7} 
    \Body
    \toprule
    Module & LN1 & MHA & RC1 & LN2 & FC1 & Act & FC2 & RC2 \\
    \midrule
    \emph{winning frequency} ($\%$) & 6.3 & 7.4 & 3.7 & 9.3 & 0.4 & \textbf{64.8} & 2.2 & 5.9 \\
    \bottomrule 
    \end{NiceTabular}
\end{table}

\subsubsection{Performance along the depth}
\begin{figure}[H]
    \centering
    \includegraphics[width=0.9\linewidth]{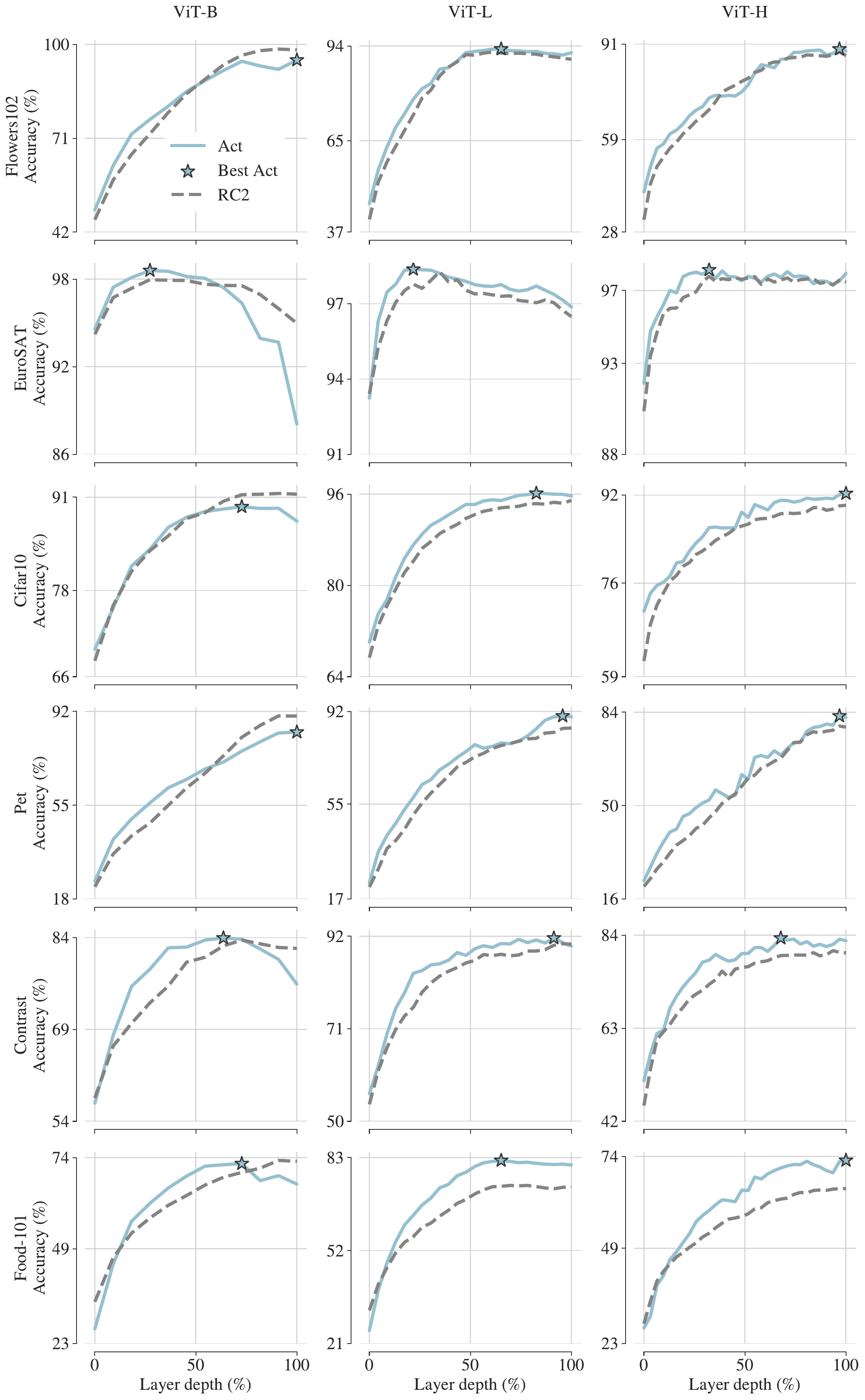}
    \caption{\textbf{Impact of $\act$ and intermediate layers in ViT (first part).} Probing accuracy across layers of ViT models for $\act$ and $\ffnres$. The star indicates the best $\act$ performance over depth. Datasets are ordered from top to bottom by increasing transfer difficulty.}
    \label{fig:probing_depth_vit_part0}
\end{figure}

\begin{figure}[H]
    \centering
    \includegraphics[width=0.9\linewidth]{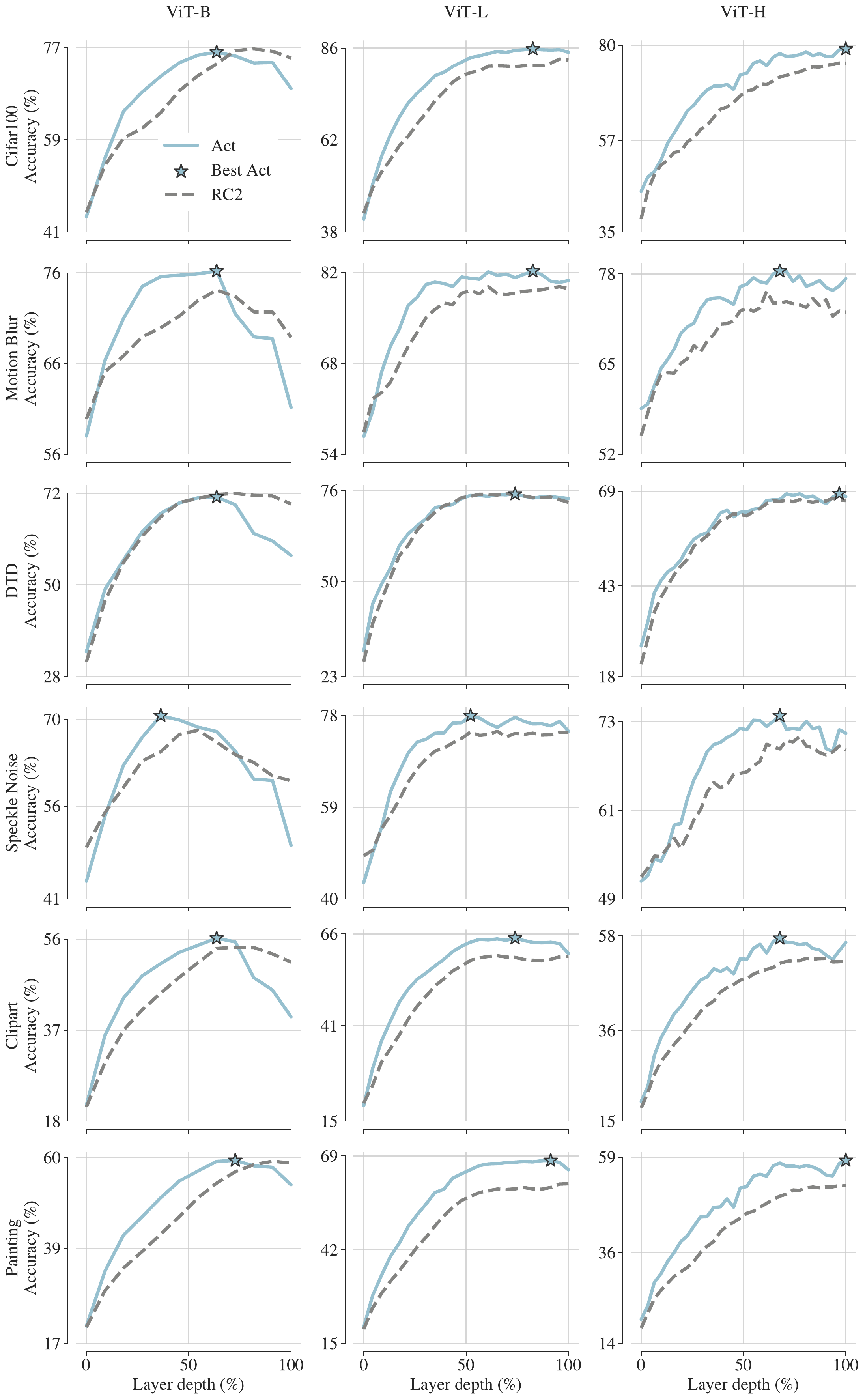}
    \caption{\textbf{Impact of $\act$ and intermediate layers in ViT (second part).} Probing accuracy across layers of ViT models for $\act$ and $\ffnres$. The star indicates the best $\act$ performance over depth. Datasets are ordered from top to bottom by increasing transfer difficulty.}
    \label{fig:probing_depth_vit_part1}
\end{figure}

\begin{figure}[H]
    \centering
    \includegraphics[width=0.9\linewidth]{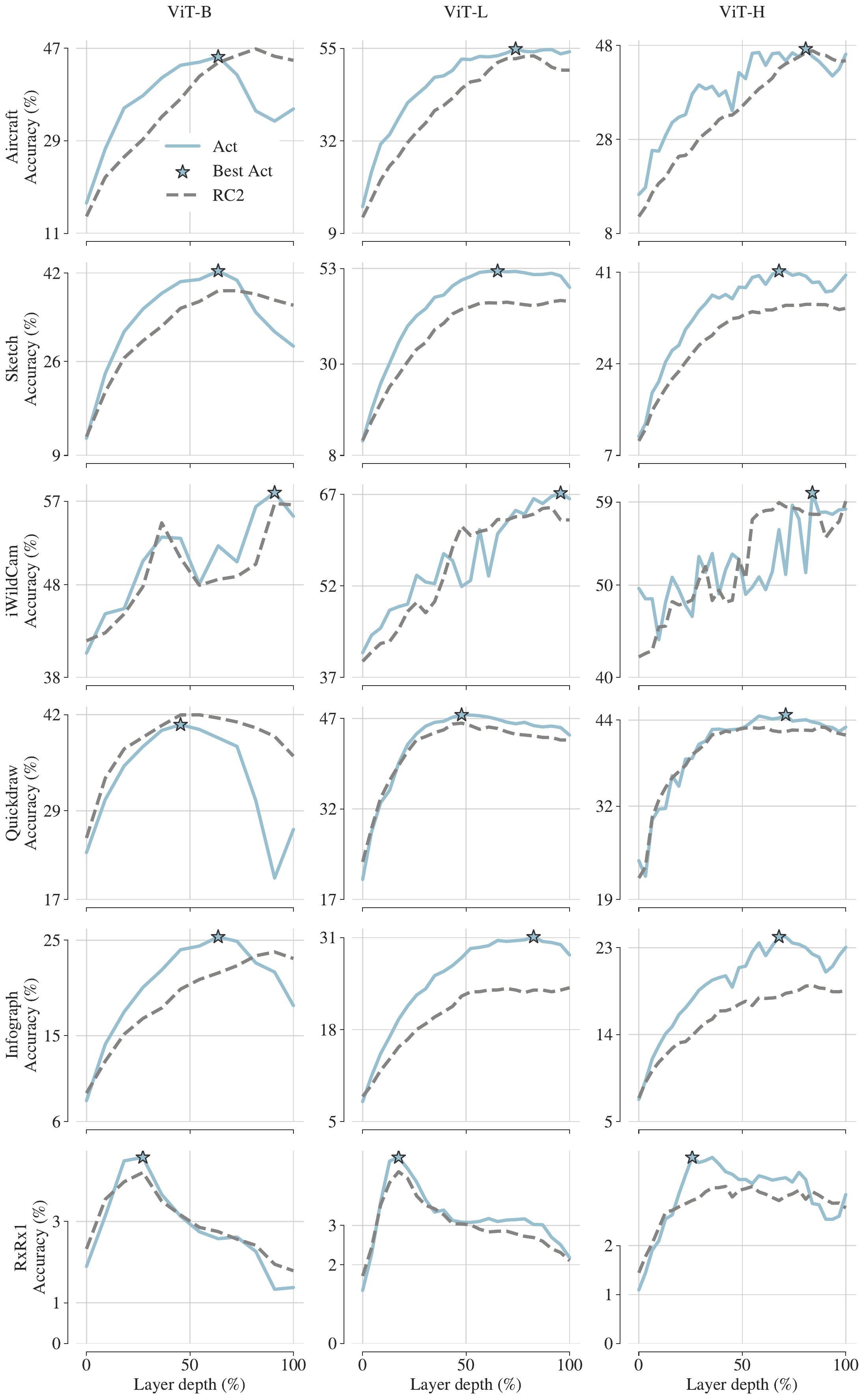}
    \caption{\textbf{Impact of $\act$ and intermediate layers in ViT (last part).} Probing accuracy across layers of ViT models for $\act$ and $\ffnres$. The star indicates the best $\act$ performance over depth. Datasets are ordered from top to bottom by increasing transfer difficulty.}
    \label{fig:probing_depth_vit_part2}
\end{figure}

\label[secinapp]{app:probing_depth_vision}
In this section, we display the performance of $\act$ and $\ffnres$ over depth. Results are consistent with~\cref{fig:probing_depth_vision}: the $\act$ module shows strong performance, often surpassing the baseline $\ffnres$. The optimal depth is generally near the end of the network, but not systematically at the final layer. This is particularly salient for ViT-B, for which intermediate $\act$ layers consistently outperform the final layer, whose performance drops, explaining the low performance of $\vfm$ observed in~\cref{tab:vfm_full_results}. Nevertheless, as shown in~\cref{fig:probing_depth_vit_part0,fig:probing_depth_vit_part1,fig:probing_depth_vit_part2}, intermediate $\act$ representations in ViT models surpass the probing baseline $\texttt{RC2@100}$ for most model-dataset pairs. This is consistent with~\cref{fig:oracle_recovery_vision}: while the fixed strategy $\vfm$ closes nearly half of the gap to the $\oracle$ gain, allowing the $\act$ depth to vary increases recovery to $78.1\%$.
\clearpage

\begin{figure}[H]
    \centering
    \includegraphics[width=\linewidth]{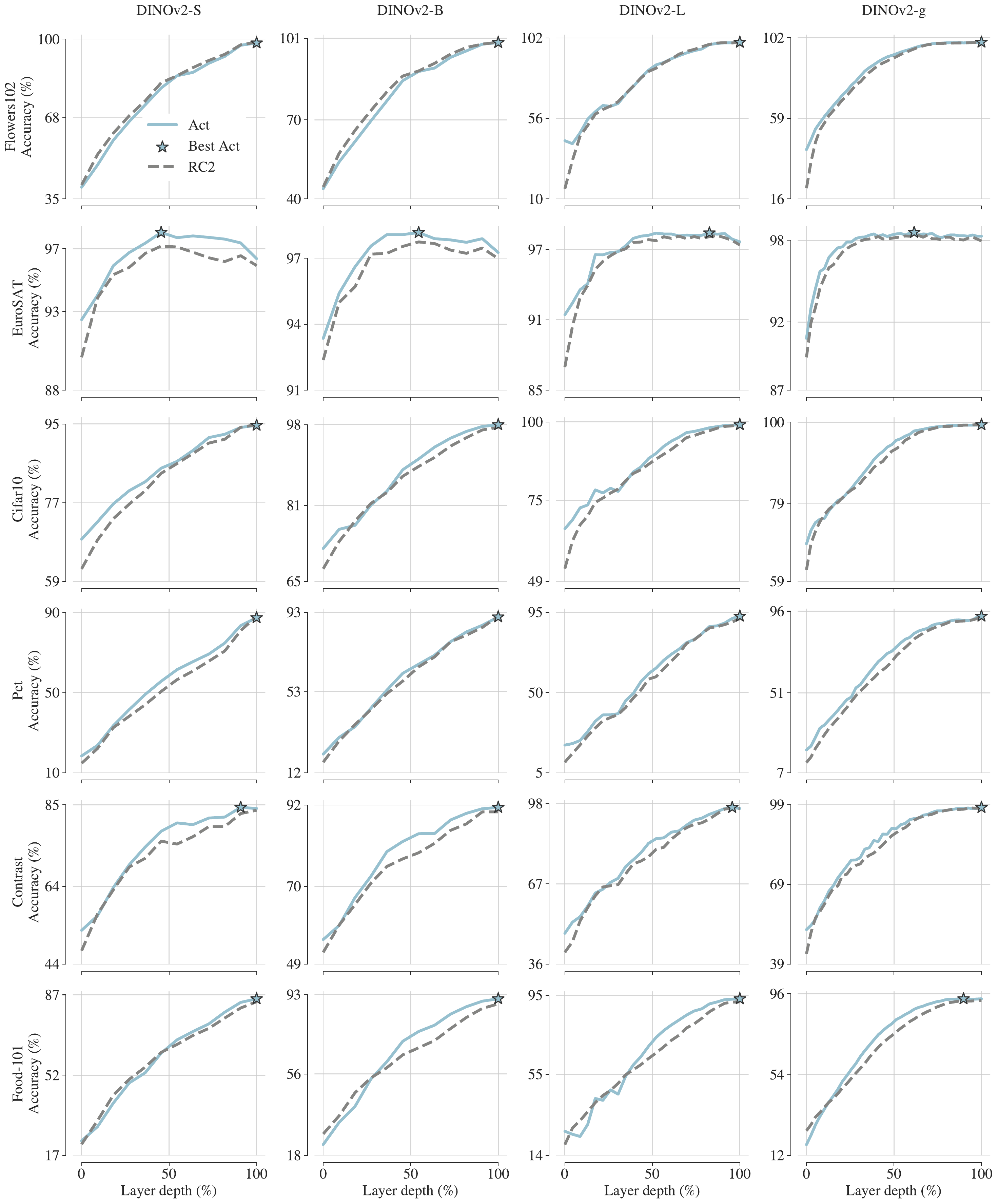}
    \caption{\textbf{Impact of $\act$ and intermediate layers in DINOv2 (first part).} Probing accuracy across layers of DINOv2 models for $\act$ and $\ffnres$. The star indicates the best $\act$ performance over depth. Datasets are ordered from top to bottom by increasing transfer difficulty.}
    \label{fig:probing_depth_dinov2_part0}
\end{figure}

\begin{figure}[H]
    \centering
    \includegraphics[width=\linewidth]{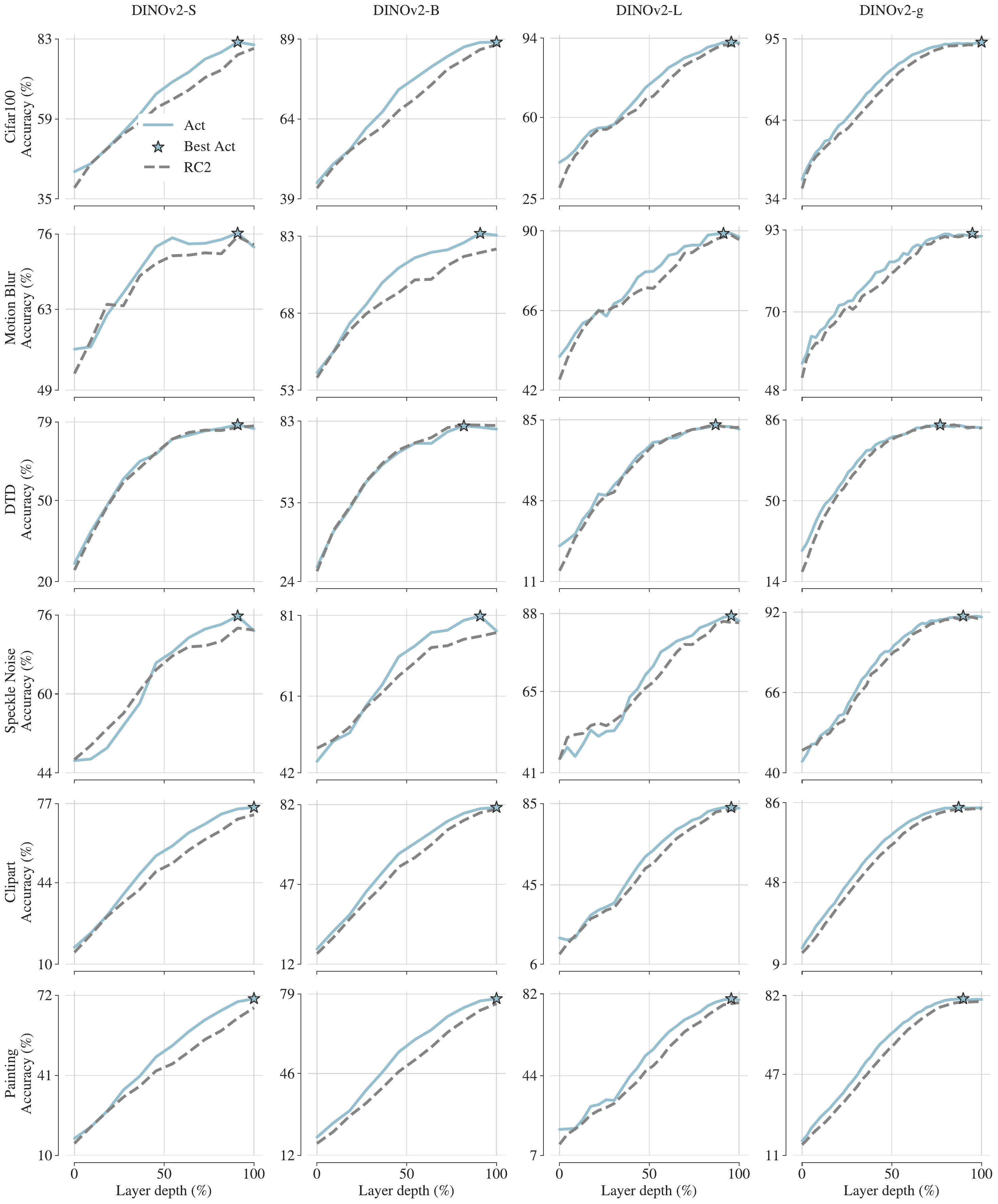}
    \caption{\textbf{Impact of $\act$ and intermediate layers in DINOv2 (second part).} Probing accuracy across layers of DINOv2 models for $\act$ and $\ffnres$. The star indicates the best $\act$ performance over depth. Datasets are ordered from top to bottom by increasing transfer difficulty.}
    \label{fig:probing_depth_dinov2_part1}
\end{figure}

\begin{figure}[H]
    \centering
    \includegraphics[width=\linewidth]{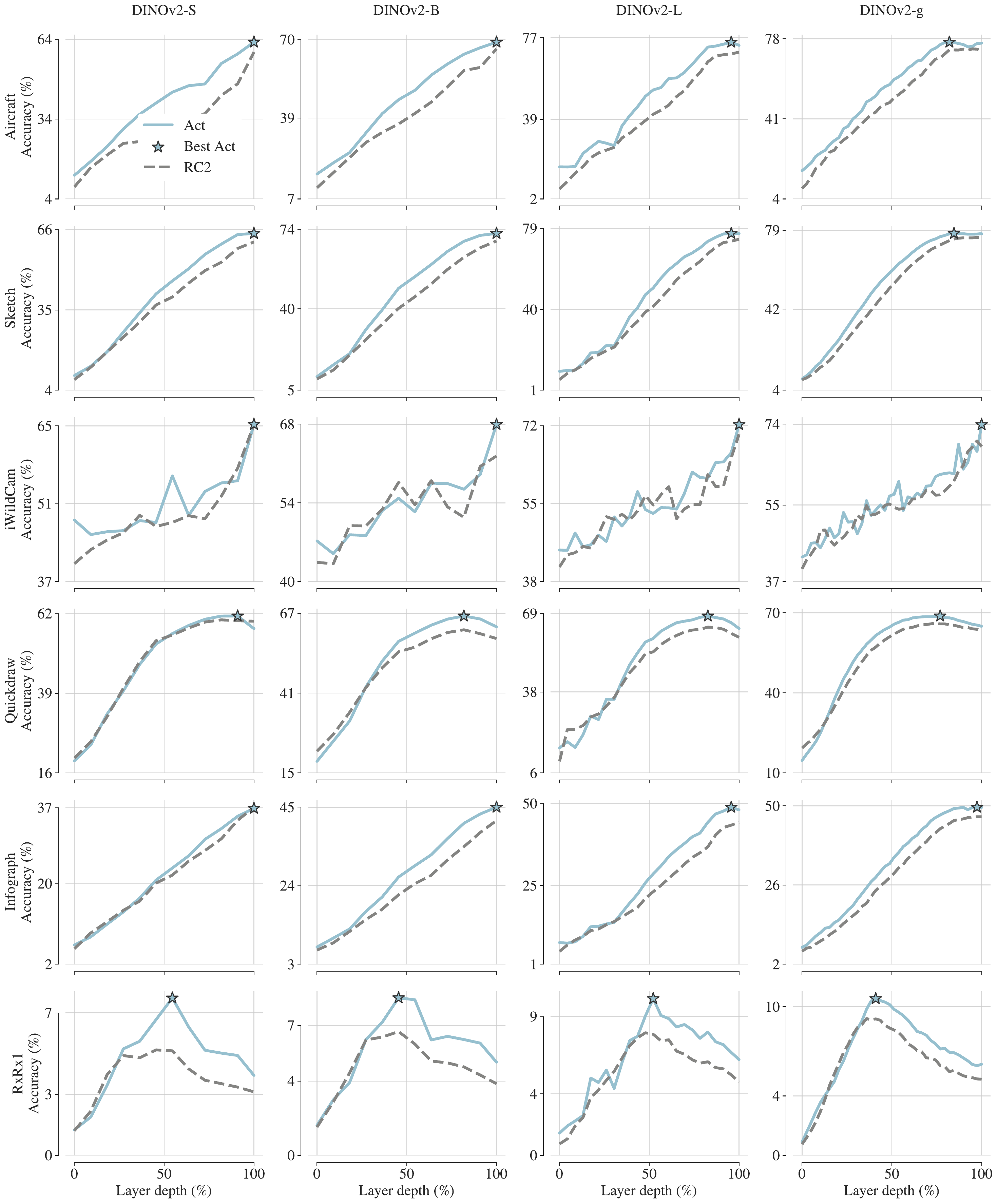}
    \caption{\textbf{Impact of $\act$ and intermediate layers in DINOv2 (last part).} Probing accuracy across layers of DINOv2 models for $\act$ and $\ffnres$. The star indicates the best $\act$ performance over depth. Datasets are ordered from top to bottom by increasing transfer difficulty.}
    \label{fig:probing_depth_dinov2_part2}
\end{figure}

\begin{figure}[H]
    \centering
    \includegraphics[width=\linewidth]{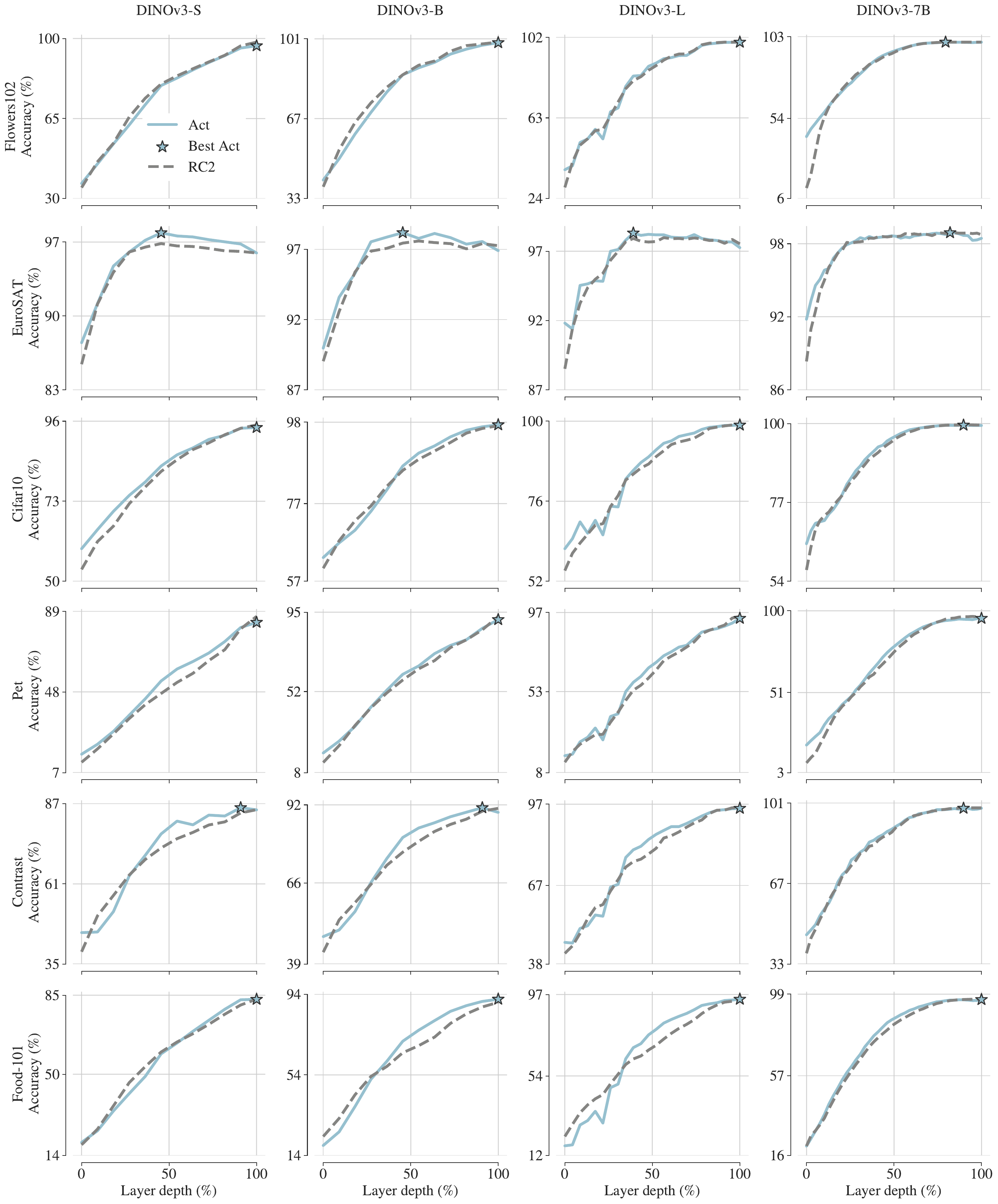}
    \caption{\textbf{Impact of $\act$ and intermediate layers in DINOv3 (first part).} Probing accuracy across layers of DINOv3 models for $\act$ and $\ffnres$. The star indicates the best $\act$ performance over depth. Datasets are ordered from top to bottom by increasing transfer difficulty.}
    \label{fig:probing_depth_dinov3_part0}
\end{figure}

\begin{figure}[H]
    \centering
    \includegraphics[width=\linewidth]{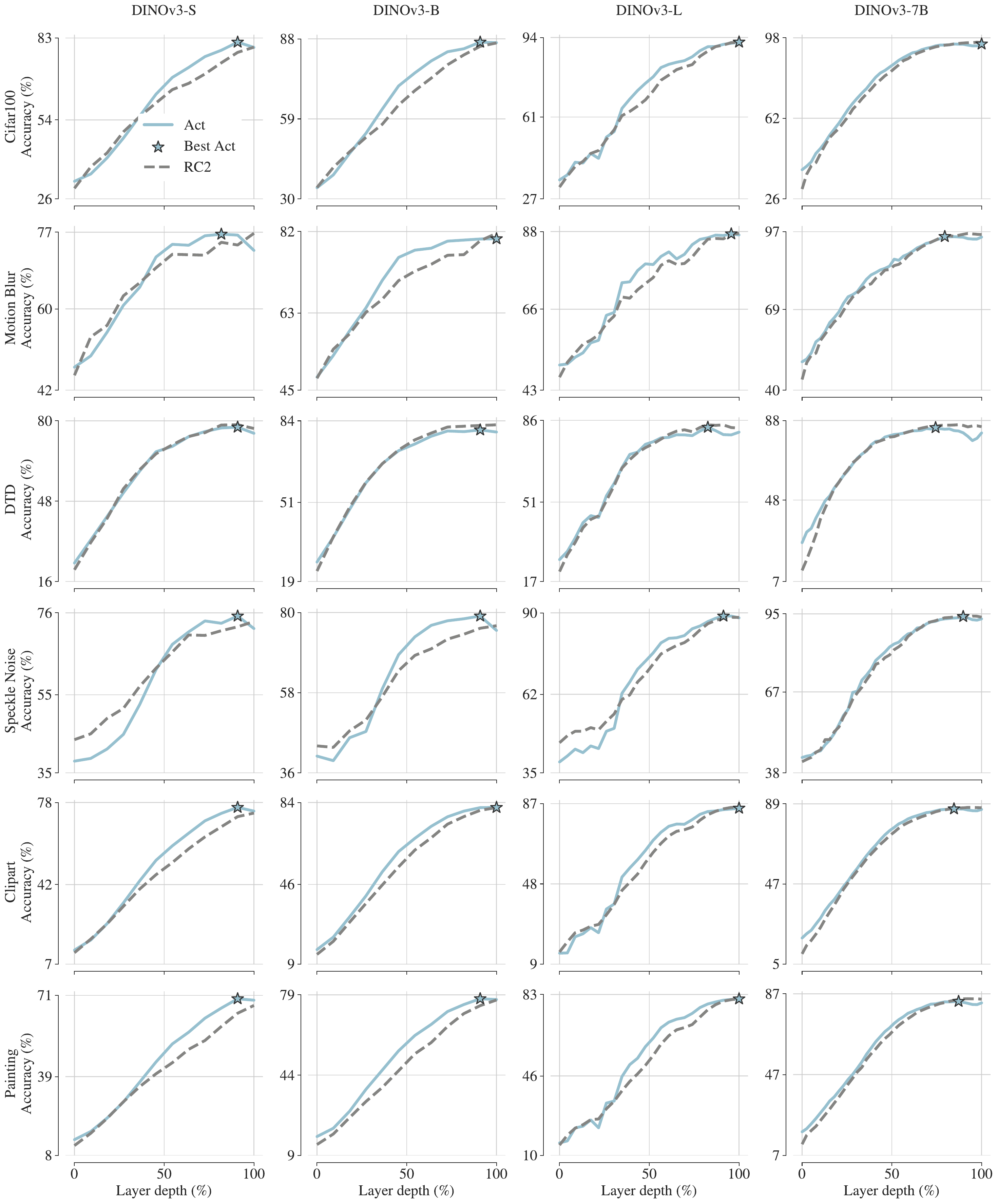}
    \caption{\textbf{Impact of $\act$ and intermediate layers in DINOv3 (second part).} Probing accuracy across layers of DINOv3 models for $\act$ and $\ffnres$. The star indicates the best $\act$ performance over depth. Datasets are ordered from top to bottom by increasing transfer difficulty.}
    \label{fig:probing_depth_dinov3_part1}
\end{figure}

\begin{figure}[H]
    \centering
    \includegraphics[width=\linewidth]{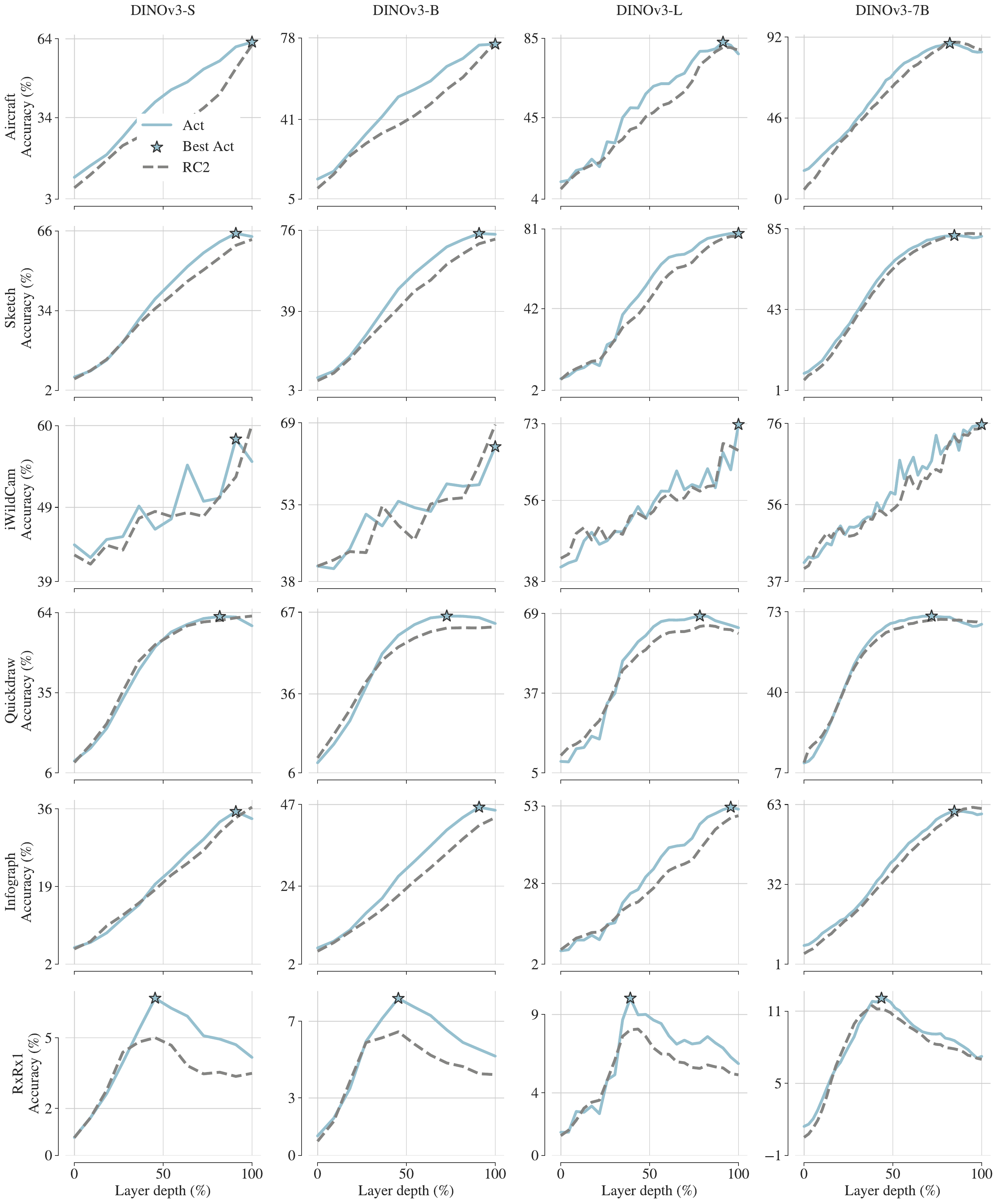}
    \caption{\textbf{Impact of $\act$ and intermediate layers in DINOv3 (last part).} Probing accuracy across layers of DINOv3 models for $\act$ and $\ffnres$. The star indicates the best $\act$ performance over depth. Datasets are ordered from top to bottom by increasing transfer difficulty.}
    \label{fig:probing_depth_dinov3_part2}
\end{figure}

\begin{figure}[H]
    \centering
    \includegraphics[width=\linewidth]{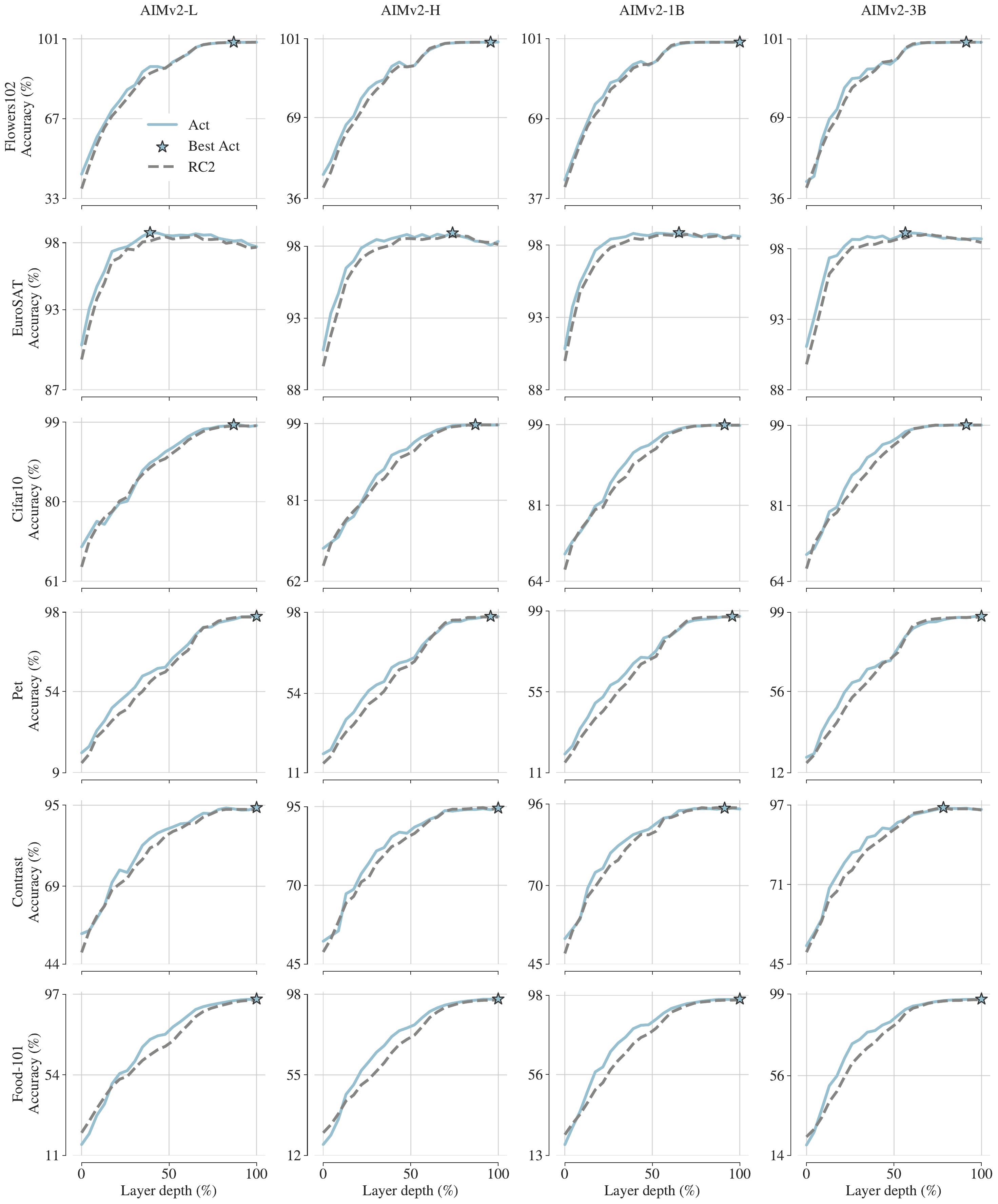}
    \caption{\textbf{Impact of $\act$ and intermediate layers in AIMv2 (first part).} Probing accuracy across layers of AIMv2 models for $\act$ and $\ffnres$. The star indicates the best $\act$ performance over depth. Datasets are ordered from top to bottom by increasing transfer difficulty.}
    \label{fig:probing_depth_aimv2_part0}
\end{figure}

\begin{figure}[H]
    \centering
    \includegraphics[width=\linewidth]{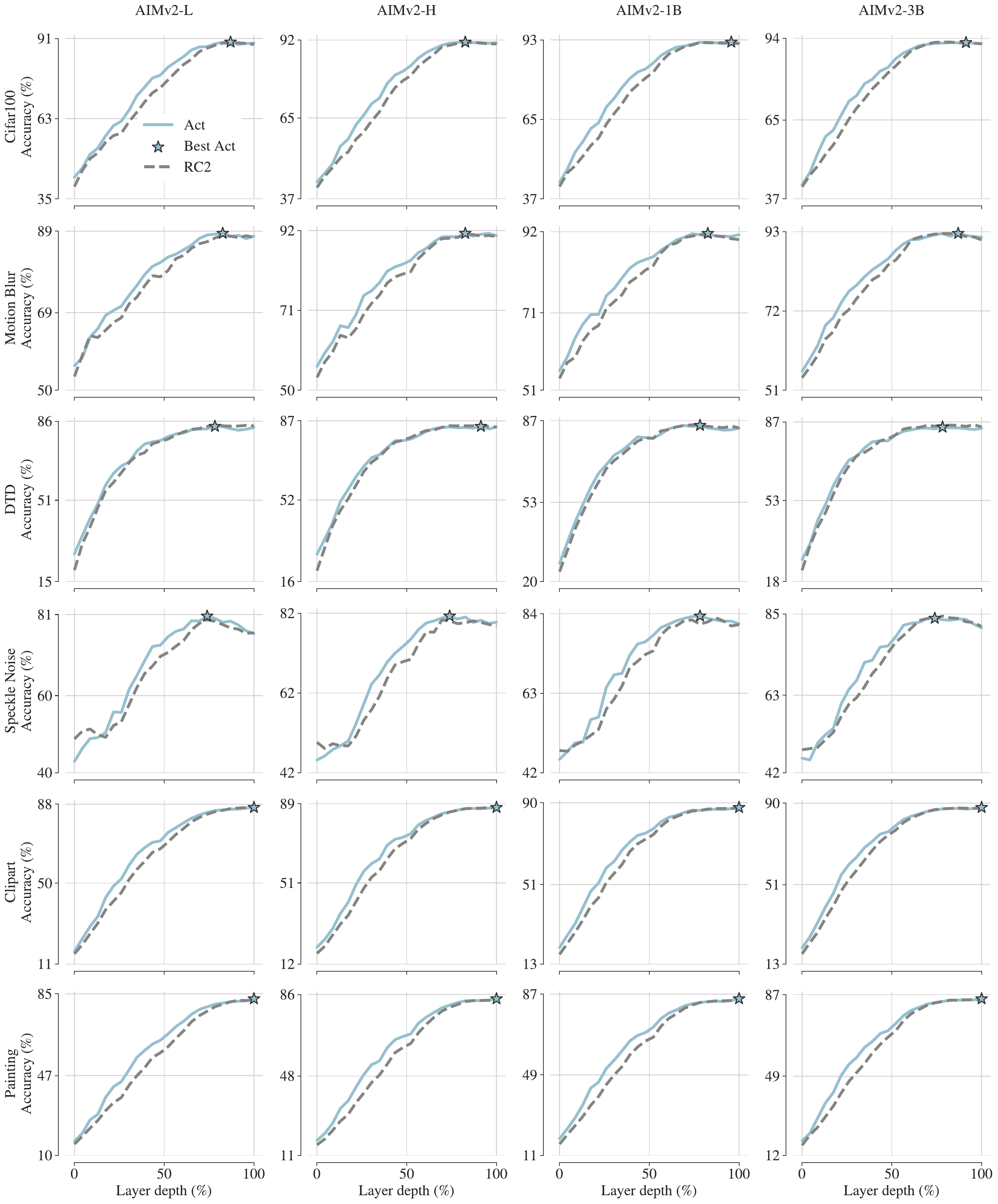}
    \caption{\textbf{Impact of $\act$ and intermediate layers in AIMv2 (second part).} Probing accuracy across layers of AIMv2 models for $\act$ and $\ffnres$. The star indicates the best $\act$ performance over depth. Datasets are ordered from top to bottom by increasing transfer difficulty.}
    \label{fig:probing_depth_aimv2_part1}
\end{figure}

\begin{figure}[H]
    \centering
    \includegraphics[width=\linewidth]{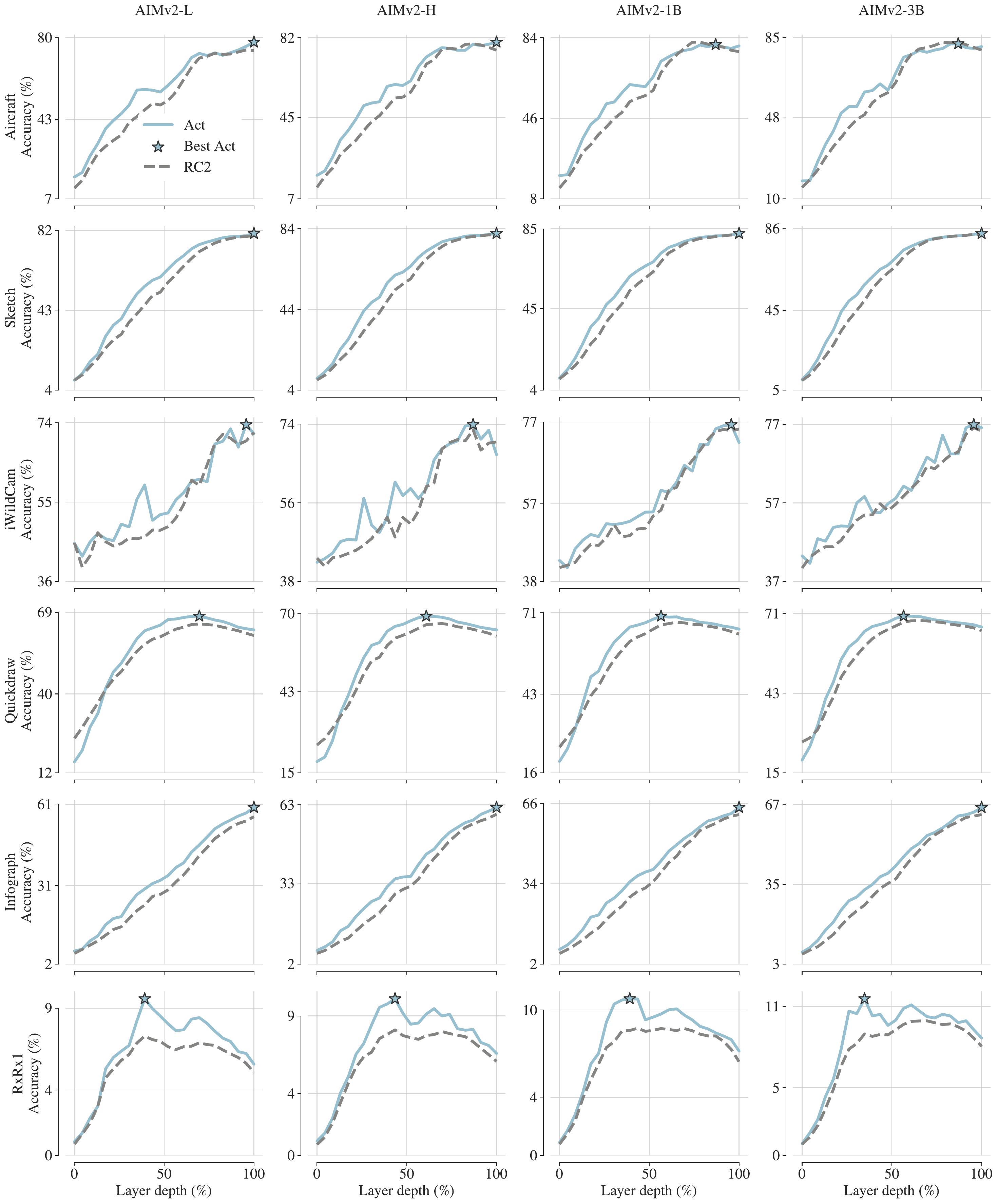}
    \caption{\textbf{Impact of $\act$ and intermediate layers in AIMv2 (last part).} Probing accuracy across layers of AIMv2 models for $\act$ and $\ffnres$. The star indicates the best $\act$ performance over depth. Datasets are ordered from top to bottom by increasing transfer difficulty.}
    \label{fig:probing_depth_aimv2_part2}
\end{figure}

\subsubsection{Gains across model sizes}
\label[secinapp]{app:friedman}
We further investigate the influence of model size on the oracle gain $\oracle$. Since larger models within a family generally achieve higher downstream accuracy~\citep{dosovitskiy2021vit,oquab2024dinov2,simeoni2025dinov3,fini2025aimv2}, they may leave less room for improvement through probing. Displayed in~\cref{tab:friedman}, a Friedman test across model sizes reveals no statistically significant differences for any family (all \(p > 0.05\)), although the limited number of model sizes restricts the power of this analysis. The corresponding Kendall's $W$ values are small ($W \leq 0.088$), indicating weak effects of model size on probing gains.
\begin{table}[!h]
\centering
\caption{\textbf{Friedman test on oracle gain $\oracle$.} Model sizes play the role of treatments, and the $18$ datasets are blocks. No family shows a statistically significant effect at $\alpha=0.05$, and Kendall's $W$ indicates small effects throughout.}
\label{tab:friedman}
\begin{NiceTabular}{lccc}
\toprule
Family  & $\chi^2$ & p-value & Kendall's $W$ \\
\midrule
ViT & 0.778 & 0.678 & 0.022 \\
DINOv2 & 4.773 & 0.189 & 0.088 \\
DINOv3 & 0.627 & 0.890 & 0.012 \\
AIMv2 & 4.600 & 0.204 & 0.085 \\
\bottomrule
\end{NiceTabular}
\end{table}

\subsubsection{Full results}
\label[secinapp]{app:vfm_full_results}
We now present the full performance comparison across all model sizes and datasets between the probing baseline $\texttt{RC2@100}$, the fixed strategy $\vfm$, and the oracle strategy, which probes the best layer-module pair for each model-dataset evaluation. The results in~\cref{tab:vfm_full_results} show that $\vfm$ gains are broadly consistent across model sizes within most families, although some exceptions arise, most notably for ViT-B and several DINOv3 variants. The depth curves in~\cref{app:probing_depth_vision} provide further insight into this variation: while $\act$ often outperforms the probing baseline $\texttt{RC2@100}$, its best-performing depth varies across model-dataset pairs and is not systematically the final layer. This effect is particularly salient for ViT-B, where intermediate $\act$ representations often surpass $\texttt{RC2@100}$, while the final-layer $\act$ used by $\vfm$ is subpar on many datasets.
\begin{table}[!h]
\centering
\caption{\textbf{Full performance comparison} between the probing baseline on the final-layer residual stream ($\texttt{RC2@100}$), the fixed strategy on the final-layer activation ($\vfm$), and the oracle strategy that probes the best representation over layers and modules for each model-dataset pair. Probing accuracies are reported for all models, with datasets ordered from left to right by increasing mean probing error (easy to hard). We report for each strategy the probing test accuracy ($\%$).}
\label{tab:vfm_full_results}
\resizebox{\linewidth}{!}{
\begin{NiceTabular}{lcccccccccccccccccccc}
\toprule
Model & Strategy & \rotatebox{90}{Flowers102} & \rotatebox{90}{EuroSAT} & \rotatebox{90}{Cifar10} & \rotatebox{90}{Pet} & \rotatebox{90}{Contrast} & \rotatebox{90}{Food-101} & \rotatebox{90}{Cifar100} & \rotatebox{90}{Motion Blur} & \rotatebox{90}{DTD} & \rotatebox{90}{Speckle Noise} & \rotatebox{90}{Clipart} & \rotatebox{90}{Painting} & \rotatebox{90}{Aircraft} & \rotatebox{90}{Sketch} & \rotatebox{90}{iWildCam} & \rotatebox{90}{Quickdraw} & \rotatebox{90}{Infograph} & \rotatebox{90}{RxRx1} & \rotatebox{90}{Average}
\\[-1pt]
\midrule
\multicolumn{21}{l}{\textit{ViT}} \\
\arrayrulecolor{gray!50}
\midrule
\multirow{3}{*}{B} & \texttt{Baseline} & 98.3 & 95.0 & 91.4 & 90.2 & 82.2 & 73.0 & 75.0 & 68.9 & 69.4 & 60.1 & 51.2 & 58.8 & 44.6 & 36.2 & 56.6 & 36.4 & 23.1 & 1.8 & 61.8 \\
 & $\vfm$ & 95.2 & 88.1 & 87.7 & 83.8 & 76.4 & 66.7 & 69.0 & 61.2 & 57.1 & 49.6 & 39.8 & 53.7 & 35.2 & 28.8 & 55.4 & 26.5 & 18.1 & 1.4 & 55.2 \\
 & \oraclerow \texttt{Oracle} & 98.6 & 98.6 & 92.2 & 90.6 & 84.2 & 73.5 & 77.3 & 76.2 & 72.4 & 70.6 & 56.2 & 59.6 & 47.8 & 42.4 & 61.7 & 42.6 & 25.3 & 4.6 & 65.2 \\
\arrayrulecolor{gray!50}
\cline{2-21}
\multirow{3}{*}{L} & \texttt{Baseline} & 89.9 & 96.5 & 94.9 & 85.4 & 90.2 & 73.2 & 82.8 & 79.5 & 72.6 & 74.5 & 59.9 & 61.0 & 49.6 & 45.1 & 62.8 & 43.5 & 23.9 & 2.1 & 66.0 \\
 & $\vfm$ & 91.9 & 96.9 & 95.7 & 89.9 & 89.8 & 80.5 & 84.9 & 80.8 & 73.7 & 74.8 & 60.6 & 65.0 & 54.2 & 48.4 & 66.3 & 44.2 & 28.5 & 2.2 & 68.2 \\
 & \oraclerow \texttt{Oracle} & 95.3 & 98.4 & 96.1 & 90.2 & 91.6 & 82.0 & 85.7 & 82.2 & 75.2 & 78.0 & 64.8 & 67.7 & 55.8 & 52.4 & 67.2 & 47.6 & 31.1 & 4.8 & 70.3 \\
\arrayrulecolor{gray!50}
\cline{2-21}
\multirow{3}{*}{H} & \texttt{Baseline} & 87.2 & 97.5 & 90.2 & 78.5 & 80.0 & 65.3 & 75.7 & 72.5 & 66.5 & 69.2 & 52.1 & 52.1 & 44.8 & 34.3 & 59.1 & 41.9 & 18.6 & 2.8 & 60.4 \\
 & $\vfm$ & 88.7 & 97.9 & 92.3 & 82.1 & 82.8 & 73.0 & 79.1 & 77.3 & 67.6 & 71.5 & 56.4 & 58.2 & 46.1 & 40.5 & 58.2 & 43.0 & 23.0 & 3.0 & 63.4 \\
 & \oraclerow \texttt{Oracle} & 92.3 & 98.1 & 92.3 & 82.6 & 83.4 & 73.0 & 79.1 & 78.4 & 68.4 & 73.8 & 57.4 & 58.2 & 47.3 & 41.2 & 60.0 & 44.7 & 24.1 & 3.8 & 64.3 \\
\arrayrulecolor{black}
\midrule
\multicolumn{21}{l}{\textit{DINOv2}} \\
\arrayrulecolor{gray!50}
\midrule
\multirow{3}{*}{S} & \texttt{Baseline} & 98.7 & 95.9 & 94.8 & 88.2 & 83.6 & 84.0 & 80.2 & 74.2 & 77.6 & 73.0 & 72.3 & 67.2 & 59.2 & 61.3 & 65.2 & 59.8 & 37.1 & 3.1 & 70.9 \\
 & $\vfm$ & 98.4 & 96.4 & 94.7 & 87.4 & 84.0 & 85.3 & 81.3 & 73.8 & 76.5 & 72.8 & 75.4 & 70.7 & 62.9 & 64.5 & 65.3 & 57.7 & 36.9 & 3.9 & 71.6 \\
 & \oraclerow \texttt{Oracle} & 98.8 & 98.0 & 95.2 & 88.2 & 84.4 & 85.3 & 82.1 & 76.2 & 78.0 & 75.8 & 75.4 & 70.7 & 62.9 & 64.5 & 66.9 & 61.3 & 37.1 & 7.7 & 72.7 \\
\arrayrulecolor{gray!50}
\cline{2-21}
\multirow{3}{*}{B} & \texttt{Baseline} & 99.5 & 97.0 & 97.4 & 91.1 & 90.2 & 88.7 & 87.2 & 80.5 & 81.4 & 76.8 & 80.0 & 75.0 & 66.4 & 69.1 & 62.3 & 58.8 & 41.4 & 3.9 & 74.8 \\
 & $\vfm$ & 99.3 & 97.3 & 98.0 & 90.7 & 91.3 & 91.0 & 88.0 & 83.2 & 80.1 & 77.1 & 80.8 & 77.1 & 69.0 & 72.3 & 68.0 & 62.6 & 45.0 & 5.0 & 76.4 \\
 & \oraclerow \texttt{Oracle} & 99.5 & 98.2 & 98.0 & 91.5 & 91.3 & 91.0 & 88.0 & 83.5 & 82.0 & 80.8 & 80.8 & 77.1 & 71.0 & 72.3 & 69.8 & 66.1 & 45.0 & 8.5 & 77.5 \\
\arrayrulecolor{gray!50}
\cline{2-21}
\multirow{3}{*}{L} & \texttt{Baseline} & 99.6 & 97.4 & 99.0 & 91.4 & 96.1 & 91.7 & 91.4 & 87.4 & 81.6 & 85.3 & 82.3 & 77.9 & 70.3 & 73.9 & 70.1 & 59.6 & 44.3 & 4.8 & 78.0 \\
 & $\vfm$ & 99.5 & 97.6 & 99.1 & 92.8 & 96.1 & 93.2 & 91.9 & 88.1 & 80.7 & 85.9 & 82.7 & 79.2 & 73.4 & 76.7 & 72.2 & 63.0 & 48.1 & 6.2 & 79.3 \\
 & \oraclerow \texttt{Oracle} & 99.7 & 98.4 & 99.1 & 92.8 & 96.5 & 93.2 & 92.3 & 89.2 & 82.8 & 87.3 & 83.1 & 79.7 & 76.7 & 76.7 & 73.2 & 68.0 & 48.9 & 10.2 & 80.4 \\
\arrayrulecolor{gray!50}
\cline{2-21}
\multirow{3}{*}{g} & \texttt{Baseline} & 99.6 & 97.9 & 99.3 & 92.8 & 97.9 & 92.5 & 93.1 & 91.1 & 82.3 & 90.0 & 83.1 & 79.5 & 72.2 & 75.5 & 68.7 & 63.2 & 46.7 & 5.1 & 79.5 \\
 & $\vfm$ & 99.6 & 98.3 & 99.3 & 93.2 & 98.0 & 93.4 & 93.7 & 91.3 & 82.6 & 90.5 & 83.6 & 80.2 & 76.0 & 77.3 & 73.9 & 64.9 & 49.5 & 6.1 & 80.6 \\
 & \oraclerow \texttt{Oracle} & 99.7 & 98.6 & 99.3 & 95.2 & 98.0 & 93.4 & 93.7 & 92.0 & 83.8 & 90.8 & 83.8 & 80.6 & 82.2 & 77.5 & 74.2 & 68.8 & 49.6 & 10.5 & 81.8 \\
\arrayrulecolor{black}
\midrule
\multicolumn{21}{l}{\textit{DINOv3}} \\
\arrayrulecolor{gray!50}
\midrule
\multirow{3}{*}{S} & \texttt{Baseline} & 98.5 & 96.0 & 94.9 & 86.5 & 85.0 & 83.3 & 79.8 & 76.8 & 76.9 & 73.8 & 73.4 & 67.0 & 61.7 & 62.5 & 60.2 & 62.7 & 36.4 & 3.5 & 71.0 \\
 & $\vfm$ & 96.8 & 96.0 & 94.2 & 83.4 & 85.0 & 83.2 & 79.7 & 73.0 & 75.0 & 72.0 & 74.2 & 69.1 & 62.7 & 63.7 & 55.2 & 59.2 & 33.8 & 4.2 & 70.0 \\
 & \oraclerow \texttt{Oracle} & 98.5 & 97.9 & 94.9 & 86.5 & 85.7 & 83.5 & 81.5 & 77.1 & 78.5 & 75.1 & 75.9 & 69.6 & 62.7 & 65.0 & 60.2 & 62.8 & 36.4 & 6.7 & 72.1 \\
\arrayrulecolor{gray!50}
\cline{2-21}
\multirow{3}{*}{B} & \texttt{Baseline} & 99.6 & 97.3 & 97.1 & 92.8 & 90.8 & 90.0 & 86.5 & 81.5 & 82.4 & 76.4 & 81.7 & 76.7 & 76.1 & 72.0 & 68.6 & 61.4 & 43.3 & 4.2 & 76.6 \\
 & $\vfm$ & 99.3 & 96.9 & 97.3 & 91.0 & 89.5 & 91.6 & 86.6 & 80.3 & 79.5 & 75.1 & 81.7 & 76.9 & 75.2 & 74.2 & 64.3 & 62.7 & 45.4 & 5.2 & 76.3 \\
 & \oraclerow \texttt{Oracle} & 99.6 & 98.2 & 97.3 & 92.8 & 91.0 & 91.6 & 86.8 & 81.5 & 82.5 & 79.0 & 81.7 & 77.3 & 76.7 & 74.5 & 69.1 & 65.5 & 46.2 & 8.2 & 77.8 \\
\arrayrulecolor{gray!50}
\cline{2-21}
\multirow{3}{*}{L} & \texttt{Baseline} & 99.7 & 97.5 & 98.8 & 94.8 & 95.6 & 93.8 & 92.0 & 87.5 & 82.7 & 88.4 & 85.1 & 81.0 & 78.9 & 77.4 & 67.1 & 61.0 & 49.9 & 5.1 & 79.8 \\
 & $\vfm$ & 99.6 & 97.3 & 98.8 & 93.7 & 95.5 & 94.5 & 92.1 & 87.1 & 81.0 & 88.3 & 84.8 & 81.0 & 77.1 & 78.8 & 72.8 & 63.3 & 52.0 & 5.9 & 80.2 \\
 & \oraclerow \texttt{Oracle} & 99.7 & 98.3 & 98.9 & 94.8 & 95.7 & 94.5 & 92.1 & 87.5 & 84.3 & 88.9 & 85.4 & 81.0 & 83.0 & 78.8 & 72.8 & 68.0 & 52.6 & 10.0 & 81.5 \\
\arrayrulecolor{gray!50}
\cline{2-21}
\multirow{3}{*}{7B} & \texttt{Baseline} & 99.7 & 98.7 & 99.6 & 96.2 & 99.0 & 96.2 & 95.8 & 96.0 & 84.9 & 93.8 & 87.0 & 84.4 & 84.7 & 82.3 & 74.9 & 68.2 & 61.5 & 7.0 & 83.9 \\
 & $\vfm$ & 99.7 & 98.4 & 99.5 & 95.4 & 98.8 & 96.2 & 95.2 & 95.0 & 81.8 & 93.2 & 86.1 & 82.5 & 83.5 & 81.1 & 75.7 & 67.9 & 59.4 & 7.2 & 83.1 \\
 & \oraclerow \texttt{Oracle} & 99.7 & 99.0 & 99.7 & 96.9 & 99.2 & 96.4 & 96.0 & 96.3 & 86.1 & 94.3 & 87.3 & 84.6 & 90.5 & 82.5 & 75.7 & 71.2 & 61.9 & 12.1 & 85.0 \\
\arrayrulecolor{black}
\midrule
\multicolumn{21}{l}{\textit{AIMv2}} \\
\arrayrulecolor{gray!50}
\midrule
\multirow{3}{*}{L} & \texttt{Baseline} & 99.6 & 97.7 & 98.1 & 95.1 & 93.5 & 93.8 & 88.9 & 87.5 & 83.9 & 76.2 & 86.3 & 82.1 & 74.1 & 79.4 & 71.6 & 60.7 & 56.4 & 5.1 & 79.4 \\
 & $\vfm$ & 99.5 & 97.7 & 98.2 & 95.7 & 94.2 & 94.4 & 89.3 & 87.8 & 83.2 & 76.1 & 86.4 & 82.7 & 77.9 & 80.4 & 71.4 & 62.7 & 59.8 & 5.6 & 80.2 \\
 & \oraclerow \texttt{Oracle} & 99.6 & 98.8 & 98.4 & 95.7 & 94.2 & 94.4 & 89.8 & 88.4 & 84.5 & 80.6 & 86.5 & 82.7 & 77.9 & 80.4 & 73.5 & 67.6 & 59.8 & 9.6 & 81.2 \\
\arrayrulecolor{gray!50}
\cline{2-21}
\multirow{3}{*}{H} & \texttt{Baseline} & 99.6 & 98.1 & 98.7 & 95.4 & 94.4 & 94.9 & 90.6 & 90.5 & 84.3 & 78.8 & 87.0 & 83.2 & 76.2 & 80.9 & 69.9 & 62.2 & 59.4 & 6.1 & 80.6 \\
 & $\vfm$ & 99.6 & 98.3 & 98.7 & 95.7 & 94.5 & 95.4 & 91.0 & 90.7 & 84.3 & 79.8 & 87.2 & 84.0 & 80.0 & 81.6 & 67.0 & 64.4 & 62.0 & 6.6 & 81.2 \\
 & \oraclerow \texttt{Oracle} & 99.7 & 98.9 & 98.7 & 95.7 & 94.9 & 95.4 & 91.3 & 91.2 & 85.2 & 81.3 & 87.2 & 84.0 & 80.0 & 81.6 & 73.9 & 69.1 & 62.0 & 10.1 & 82.2 \\
\arrayrulecolor{gray!50}
\cline{2-21}
\multirow{3}{*}{1B} & \texttt{Baseline} & 99.6 & 98.4 & 98.9 & 96.0 & 94.8 & 95.6 & 91.7 & 90.0 & 84.0 & 81.2 & 87.3 & 84.0 & 77.5 & 81.9 & 75.3 & 63.7 & 61.5 & 6.4 & 81.5 \\
 & $\vfm$ & 99.6 & 98.6 & 98.8 & 96.0 & 94.4 & 96.0 & 91.9 & 91.2 & 83.9 & 81.3 & 87.7 & 84.8 & 80.2 & 82.7 & 72.0 & 65.4 & 64.4 & 7.2 & 82.0 \\
 & \oraclerow \texttt{Oracle} & 99.7 & 98.9 & 99.0 & 96.0 & 95.0 & 96.0 & 92.3 & 91.5 & 86.0 & 83.4 & 87.7 & 84.8 & 82.0 & 82.7 & 76.3 & 69.9 & 64.4 & 10.8 & 83.1 \\
\arrayrulecolor{gray!50}
\cline{2-21}
\multirow{3}{*}{3B} & \texttt{Baseline} & 99.6 & 98.4 & 99.1 & 96.2 & 95.3 & 95.9 & 92.1 & 90.8 & 84.8 & 81.6 & 87.4 & 84.5 & 79.1 & 82.8 & 75.3 & 65.0 & 63.0 & 8.0 & 82.2 \\
 & $\vfm$ & 99.6 & 98.7 & 99.0 & 96.8 & 95.5 & 96.3 & 92.2 & 91.5 & 84.2 & 81.2 & 87.9 & 85.1 & 80.8 & 83.4 & 76.1 & 66.2 & 65.8 & 8.7 & 82.7 \\
 & \oraclerow \texttt{Oracle} & 99.7 & 99.1 & 99.1 & 96.8 & 96.2 & 96.3 & 92.7 & 92.6 & 85.8 & 84.5 & 87.9 & 85.1 & 83.2 & 83.4 & 76.9 & 70.1 & 65.8 & 11.6 & 83.7 \\
\arrayrulecolor{black}
\bottomrule
\end{NiceTabular}
}
\end{table}

\paragraph{Further discussion on $\act$.}
Linear probing measures the linear separability of representations on downstream data. Transformer architectures preserve the representation dimensionality across layers, as the output of each block is passed as input to the subsequent block. However, different modules within a transformer block can operate at different dimensionalities. In particular, residual-stream representations have the embedding dimension $d$, whereas FFN intermediate representations have the hidden dimension $h$, typically larger than $d$ (e.g., $h=4d$ in standard FFNs). Prior work has shown that gains from feature augmentation cannot be attributed solely to increased feature dimensionality~\citep[see][Section 3.2]{evci2022head2toe}. Their setting differs from ours, however, as they artificially increase the feature dimension by aggregating embeddings across the network, while we preserve the native dimensionality of each representation. Therefore, $\act$ has a larger dimensionality than the residual stream. To assess whether its larger dimensionality alone explains the gains of $\act$, we compare it with $\mathrm{FC1}$, which has the same FFN hidden dimension $h>d$. This controlled comparison is on the $307$M-parameter variant common to all four model families. As shown in~\cref{tab:vfm_results_act_fc1}, overall, $\vfm$ outperforms $\texttt{FC1@100}$, with better performance in a majority of evaluations. $\texttt{FC1@100}$ is also subpar compared to the probing baseline $\texttt{RC2@100}$, indicating that the gains of $\vfm$ cannot be explained by its larger dimensionality alone.

\begin{table}[!h]
    \centering
    \caption{\textbf{Additional performance comparison} between the probing baseline on the final-layer residual stream ($\texttt{RC2@100}$), the final-layer FFN first linear layer ($\texttt{FC1@100}$), and the fixed strategy on the final-layer activation ($\vfm$). Probing accuracies are reported on the $307$M-parameter variant common to each model family, providing a controlled comparison across families. Datasets are ordered by increasing mean probing error, from easy to hard. We report for each strategy the probing test accuracy ($\%$).}
    \label{tab:vfm_results_act_fc1}
    \resizebox{\linewidth}{!}{
   \begin{NiceTabular}{lccccccccccccccccccc}
    \toprule
    Strategy & \rotatebox{90}{Flowers102} & \rotatebox{90}{EuroSAT} & \rotatebox{90}{Cifar10} & \rotatebox{90}{Pet} & \rotatebox{90}{Contrast} & \rotatebox{90}{Food-101} & \rotatebox{90}{Cifar100} & \rotatebox{90}{Motion Blur} & \rotatebox{90}{DTD} & \rotatebox{90}{Speckle Noise} & \rotatebox{90}{Clipart} & \rotatebox{90}{Painting} & \rotatebox{90}{Aircraft} & \rotatebox{90}{Sketch} & \rotatebox{90}{iWildCam} & \rotatebox{90}{Quickdraw} & \rotatebox{90}{Infograph} & \rotatebox{90}{RxRx1} & \rotatebox{90}{Average}
    \\[-1pt]
    \midrule
    \multicolumn{20}{l}{\textit{ViT}} \\
    \arrayrulecolor{gray!50}
    \midrule
    \texttt{Baseline} & 89.9 & 96.5 & 94.9 & 85.4 & 90.2 & 73.2 & 82.8 & 79.5 & 72.6 & 74.5 & 59.9 & 61.0 & 49.6 & 45.1 & 62.8 & 43.5 & 23.9 & 2.1 & 66.0 \\
    $\texttt{FC1@100}$ & 91.1 & 96.0 & 94.0 & 86.5 & 88.7 & 72.8 & 80.9 & 77.8 & 71.7 & 69.9 & 54.8 & 58.0 & 50.0 & 40.3 & 61.2 & 40.2 & 21.1 & 1.9 & 64.3 \\
    $\texttt{Act@100}$ & 91.9 & 96.9 & 95.7 & 89.9 & 89.8 & 80.5 & 84.9 & 80.8 & 73.7 & 74.8 & 60.6 & 65.0 & 54.2 & 48.4 & 66.3 & 44.2 & 28.5 & 2.2 & 68.2 \\
    \arrayrulecolor{black}
    \midrule
    \multicolumn{20}{l}{\textit{DINOv2}} \\
    \arrayrulecolor{gray!50}
    \midrule
    \texttt{Baseline} & 99.6 & 97.4 & 99.0 & 91.4 & 96.1 & 91.7 & 91.4 & 87.4 & 81.6 & 85.3 & 82.3 & 77.9 & 70.3 & 73.9 & 70.1 & 59.6 & 44.3 & 4.8 & 78.0 \\
    $\texttt{FC1@100}$ & 99.5 & 96.9 & 98.9 & 90.8 & 95.8 & 91.3 & 90.7 & 87.8 & 80.0 & 85.3 & 81.3 & 77.0 & 72.7 & 72.3 & 71.7 & 58.1 & 43.5 & 4.4 & 77.7 \\
    $\texttt{Act@100}$ & 99.5 & 97.6 & 99.1 & 92.8 & 96.1 & 93.2 & 91.9 & 88.1 & 80.7 & 85.9 & 82.7 & 79.2 & 73.4 & 76.7 & 72.2 & 63.0 & 48.1 & 6.2 & 79.3 \\
    \arrayrulecolor{black}
    \midrule
    \multicolumn{20}{l}{\textit{DINOv3}} \\
    \arrayrulecolor{gray!50}
    \midrule
    \texttt{Baseline} & 99.7 & 97.5 & 98.8 & 94.8 & 95.6 & 93.8 & 92.0 & 87.5 & 82.7 & 88.4 & 85.1 & 81.0 & 78.9 & 77.4 & 67.1 & 61.0 & 49.9 & 5.1 & 79.8 \\
    $\texttt{FC1@100}$ & 99.7 & 97.3 & 98.8 & 93.6 & 95.3 & 93.3 & 91.4 & 87.2 & 81.2 & 87.2 & 83.9 & 79.8 & 77.1 & 76.4 & 68.2 & 59.6 & 48.9 & 4.7 & 79.1 \\
    $\texttt{Act@100}$ & 99.6 & 97.3 & 98.8 & 93.7 & 95.5 & 94.5 & 92.1 & 87.1 & 81.0 & 88.3 & 84.8 & 81.0 & 77.1 & 78.8 & 72.8 & 63.3 & 52.0 & 5.9 & 80.2 \\
    \arrayrulecolor{black}
    \midrule
    \multicolumn{20}{l}{\textit{AIMv2}} \\
    \arrayrulecolor{gray!50}
    \midrule
    \texttt{Baseline} & 99.6 & 97.7 & 98.1 & 95.1 & 93.5 & 93.8 & 88.9 & 87.5 & 83.9 & 76.2 & 86.3 & 82.1 & 74.1 & 79.4 & 71.6 & 60.7 & 56.4 & 5.1 & 79.4 \\
    $\texttt{FC1@100}$ & 99.6 & 97.6 & 98.0 & 95.1 & 93.4 & 93.4 & 88.7 & 86.7 & 83.5 & 74.7 & 85.9 & 81.6 & 73.1 & 78.3 & 64.9 & 60.0 & 54.5 & 4.9 & 78.5 \\
    $\texttt{Act@100}$ & 99.5 & 97.7 & 98.2 & 95.7 & 94.2 & 94.4 & 89.3 & 87.8 & 83.2 & 76.1 & 86.4 & 82.7 & 77.9 & 80.4 & 71.4 & 62.7 & 59.8 & 5.6 & 80.2 \\
    \arrayrulecolor{black}
    \bottomrule
    \end{NiceTabular}
    }
\end{table}
\clearpage

\subsection{Large language models}
\label[secinapp]{app:llm_add_exp}
The results presented in this section relate to the experiments conducted on LLMs in~\cref{sec:llm_exp}.

\subsubsection{Module comparison}
\label[secinapp]{app:llm_module_comparison}
In~\cref {tab:llm_module_comparison}, we display across models the module that belongs to the layer-module pair achieving the oracle gain $\oracle$. We can see that $\ffnnorm$ is optimal for all models but one. 
\begin{table}[!h]
\centering
    \caption{\textbf{Module comparison.} Module in the layer-module pair that achieves the oracle gain $\oracle$.}
    \label{tab:llm_module_comparison}
    \begin{NiceTabular}{lccccccccc}
    \toprule
    \multirow{2}{*}{Model} & \multicolumn{3}{c}{\textit{Qwen3}} & \multicolumn{3}{c}{\textit{Llama3}} & \multicolumn{3}{c}{\textit{Gemma3}} \\
    \cmidrule(l){2-4} \cmidrule(l){5-7} \cmidrule(l){8-10}
     & 1.7B & 4B & 8B & 1B & 3B & 8B & 1B & 4B & 12B\\
    \midrule
    \emph{best module} & $\ffnnorm$ & $\ffnnorm$  & $\ffnnorm$  & $\ffnnorm$  & $\ffnnorm$  & $\ffnnorm$ &   $\ffnnorm$ & $\linact$  & $\ffnnorm$ \\
    \bottomrule
    \end{NiceTabular}
\end{table}

\subsubsection{Performance along the depth}
In this section, we display the performance of $\ffnnorm$ and $\ffnres$ over depth for all models. Results are consistent with~\cref{fig:probing_depth_language}. $\ffnnorm$ outperforms $\ffnres$, with the highest performance achieved in the interval between $50\%$ and $70\%$ depth.
\begin{figure}[!h]
    \centering
    \includegraphics[width=\linewidth]{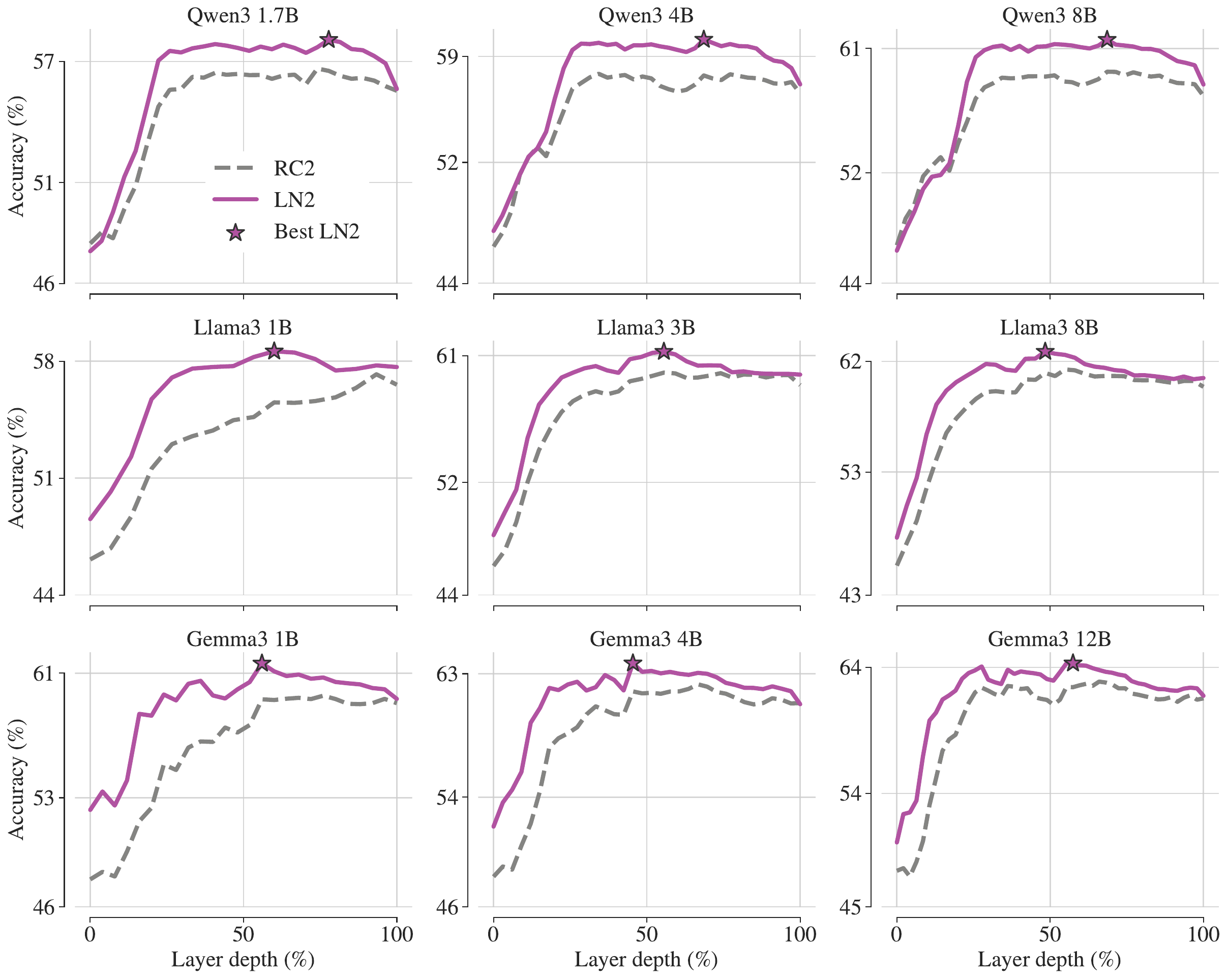}
    \caption{\textbf{Impact of $\ffnnorm$ and intermediate layers on performance across model sizes.} Probing accuracy across layers for $\ffnnorm$ and $\ffnres$ on \texttt{MTEB(Multilingual)}. The star indicates the best performance obtained on $\ffnnorm$.}
    \label{fig:probing_depth_all_language}
\end{figure}

\subsubsection{Full results}
\label[secinapp]{app:llm_full_results}
We now present the detailed performance comparison on \texttt{MTEB(Multilingual)} between the probing baseline $\texttt{RC2@100}$, the depth-matched $\texttt{RC2@60}$, and the fixed strategy $\llm$. \cref{tab:llm_full_results_task} provides an overview with performance averaged across models. \cref{tab:llm_full_results_qwen3,tab:llm_full_results_llama3,tab:llm_full_results_gemma3} provide the detailed results across model families and tasks. We can see that the gains of $\llm$ are consistent across models and tasks.
\begin{table}[!h]
\centering
\caption{\textbf{Full performance comparison per task} between the probing baseline on the final-layer residual stream ($\ffnres$), residual-stream probing at $60\%$ depth ($\texttt{RC2@60}$), and the fixed strategy $\llm$. The accuracy is averaged over models for each task, and the difference in percentage points relative to the probing baseline is shown in \textcolor{Green}{green} if improved, and in \textcolor{Red}{red} otherwise.}
\label{tab:llm_full_results_task}
\resizebox{\linewidth}{!}{
\begin{NiceTabular}{@{} l *{3}{w{r}{2.8em}@{\hspace{6pt}}w{r}{2.2em}@{\hspace{4pt}}}@{}}
\toprule
Task & \multicolumn{2}{c}{\texttt{Baseline}} & \multicolumn{2}{c}{\texttt{RC2@60}} & \multicolumn{2}{c}{$\llm$} \\
\midrule
BulgarianStoreReviewSentimentClassification & 50.2 & & 56.3 & \addsmall{+6.1} & 59.0 & \addsmall{+8.9} \\
DalajClassification & 50.2 & & 50.5 & \addsmall{+0.3} & 50.5 & \addsmall{+0.3} \\
DBpediaClassification & 91.3 & & 89.2 & \subsmall{-2.2} & 91.8 & \addsmall{+0.5} \\
EstonianValenceClassification & 37.2 & & 39.6 & \addsmall{+2.4} & 40.5 & \addsmall{+3.2} \\
FinancialPhrasebankClassification & 64.3 & & 66.6 & \addsmall{+2.2} & 69.1 & \addsmall{+4.8} \\
GujaratiNewsClassification & 77.2 & & 77.4 & \addsmall{+0.2} & 80.5 & \addsmall{+3.3} \\
IndonesianIdClickbaitClassification & 63.1 & & 64.3 & \addsmall{+1.3} & 64.7 & \addsmall{+1.6} \\
IsiZuluNewsClassification & 24.5 & & 24.6 & \addsmall{+0.1} & 25.6 & \addsmall{+1.1} \\
ItaCaseholdClassification & 54.8 & & 49.4 & \subsmall{-5.4} & 56.0 & \addsmall{+1.2} \\
KorSarcasmClassification & 61.3 & & 57.9 & \subsmall{-3.4} & 58.7 & \subsmall{-2.7} \\
KurdishSentimentClassification & 60.8 & & 62.9 & \addsmall{+2.1} & 63.4 & \addsmall{+2.6} \\
NepaliNewsClassification & 89.8 & & 89.5 & \subsmall{-0.2} & 91.4 & \addsmall{+1.6} \\
NusaX-senti & 58.0 & & 60.6 & \addsmall{+2.6} & 61.5 & \addsmall{+3.5} \\
PAC & 68.8 & & 68.6 & \subsmall{-0.2} & 70.4 & \addsmall{+1.6} \\
PoemSentimentClassification & 39.8 & & 41.8 & \addsmall{+2.0} & 43.1 & \addsmall{+3.3} \\
PolEmo2.0-OUT & 44.3 & & 53.2 & \addsmall{+8.9} & 55.2 & \addsmall{+10.9} \\
PunjabiNewsClassification & 72.9 & & 73.3 & \addsmall{+0.4} & 75.5 & \addsmall{+2.6} \\
ScalaClassification & 54.2 & & 55.2 & \addsmall{+1.0} & 55.1 & \addsmall{+0.9} \\
SentimentAnalysisHindi & 51.2 & & 55.3 & \addsmall{+4.2} & 56.3 & \addsmall{+5.2} \\
SinhalaNewsClassification & 59.4 & & 60.0 & \addsmall{+0.6} & 63.1 & \addsmall{+3.7} \\
SiswatiNewsClassification & 72.6 & & 69.6 & \subsmall{-2.9} & 73.5 & \addsmall{+1.0} \\
SwahiliNewsClassification & 61.3 & & 61.1 & \subsmall{-0.1} & 63.0 & \addsmall{+1.8} \\
TswanaNewsClassification & 30.3 & & 27.4 & \subsmall{-2.9} & 30.0 & \subsmall{-0.3} \\
TweetTopicSingleClassification & 66.9 & & 62.2 & \subsmall{-4.7} & 64.7 & \subsmall{-2.2} \\
\bottomrule
\end{NiceTabular}
}
\end{table}

\begin{table}[!h]
\centering
\caption{\textbf{Full performance comparison on Qwen3} between the probing baseline on the final-layer residual stream ($\ffnres$), residual-stream probing at $60\%$ depth ($\texttt{RC2@60}$), and the fixed strategy $\llm$. We report for each strategy the probing test accuracy ($\%$).}
\label{tab:llm_full_results_qwen3}
\resizebox{\linewidth}{!}{
\begin{NiceTabular}{@{} l *{3}{w{c}{3.6em}@{\hspace{12pt}}}*{3}{w{c}{3.6em}@{\hspace{12pt}}}*{3}{w{c}{3.6em}@{\hspace{12pt}}}@{}}
\toprule
\multirow{2}{*}{Task} & \multicolumn{3}{c}{\textit{Qwen3 1.7B}} & \multicolumn{3}{c}{\textit{Qwen3 4B}} & \multicolumn{3}{c}{\textit{Qwen3 8B}} \\
\cmidrule(l){2-4} \cmidrule(l){5-7} \cmidrule(l){8-10}
 & \texttt{Baseline} & \texttt{RC2@60} & $\llm$ & \texttt{Baseline} & \texttt{RC2@60} & $\llm$ & \texttt{Baseline} & \texttt{RC2@60} & $\llm$ \\
\midrule
BulgarianStoreReviewSentimentClassification & 45.0 & 53.6 & 57.0 & 46.4 & 53.5 & 57.3 & 46.8 & 56.4 & 61.8 \\
DalajClassification & 49.7 & 50.5 & 50.2 & 49.7 & 50.4 & 50.4 & 50.0 & 50.3 & 50.7 \\
DBpediaClassification & 91.2 & 85.8 & 88.4 & 91.3 & 82.4 & 88.8 & 91.1 & 86.1 & 90.2 \\
EstonianValenceClassification & 33.7 & 36.4 & 37.3 & 36.4 & 39.7 & 40.3 & 36.9 & 40.3 & 41.3 \\
FinancialPhrasebankClassification & 64.3 & 68.6 & 70.0 & 64.2 & 65.5 & 70.9 & 64.3 & 67.3 & 72.3 \\
GujaratiNewsClassification & 72.1 & 71.0 & 77.8 & 72.8 & 73.4 & 80.2 & 76.8 & 76.3 & 82.3 \\
IndonesianIdClickbaitClassification & 60.1 & 62.5 & 62.5 & 62.4 & 62.5 & 64.0 & 63.7 & 63.8 & 65.4 \\
IsiZuluNewsClassification & 23.1 & 23.0 & 23.0 & 23.4 & 23.2 & 24.2 & 23.3 & 22.7 & 24.3 \\
ItaCaseholdClassification & 48.1 & 47.0 & 48.1 & 52.1 & 45.8 & 49.4 & 53.7 & 46.5 & 52.7 \\
KorSarcasmClassification & 61.8 & 55.6 & 56.8 & 56.5 & 57.0 & 58.3 & 56.4 & 57.4 & 58.6 \\
KurdishSentimentClassification & 56.6 & 57.9 & 57.9 & 58.7 & 59.6 & 61.2 & 61.2 & 62.8 & 64.4 \\
NepaliNewsClassification & 84.6 & 84.5 & 87.8 & 85.9 & 85.4 & 89.5 & 87.8 & 87.4 & 90.8 \\
NusaX-senti & 55.2 & 58.1 & 58.5 & 58.2 & 58.2 & 60.4 & 57.9 & 60.8 & 61.2 \\
PAC & 69.3 & 67.0 & 71.2 & 70.9 & 68.5 & 72.2 & 70.3 & 68.3 & 72.0 \\
PoemSentimentClassification & 39.5 & 43.5 & 45.7 & 37.4 & 42.0 & 45.6 & 39.1 & 43.3 & 47.6 \\
PolEmo2.0-OUT & 43.0 & 52.3 & 55.0 & 42.4 & 52.8 & 57.6 & 43.0 & 53.0 & 58.5 \\
PunjabiNewsClassification & 68.3 & 73.9 & 73.6 & 73.2 & 71.2 & 74.3 & 73.9 & 72.4 & 75.7 \\
ScalaClassification & 52.9 & 53.5 & 53.2 & 54.1 & 55.3 & 55.1 & 54.4 & 56.1 & 55.8 \\
SentimentAnalysisHindi & 50.3 & 53.8 & 57.5 & 49.4 & 55.7 & 58.0 & 49.5 & 56.1 & 58.8 \\
SinhalaNewsClassification & 48.0 & 47.0 & 47.7 & 51.1 & 52.4 & 56.1 & 54.5 & 57.2 & 61.1 \\
SiswatiNewsClassification & 71.9 & 70.0 & 71.0 & 71.1 & 71.4 & 72.1 & 72.4 & 71.9 & 72.2 \\
SwahiliNewsClassification & 52.3 & 55.6 & 55.6 & 59.4 & 59.1 & 60.5 & 60.5 & 61.7 & 61.9 \\
TswanaNewsClassification & 25.4 & 20.7 & 19.0 & 24.8 & 19.1 & 20.1 & 27.3 & 23.2 & 24.6 \\
TweetTopicSingleClassification & 66.6 & 55.5 & 57.7 & 65.6 & 56.3 & 60.1 & 66.4 & 58.4 & 61.7 \\
\bottomrule
\end{NiceTabular}
}
\end{table}

\begin{table}[!h]
\centering
\caption{\textbf{Full performance comparison on Llama3} between the probing baseline on the final-layer residual stream ($\ffnres$), residual-stream probing at $60\%$ depth ($\texttt{RC2@60}$), and the fixed strategy $\llm$. We report for each strategy the probing test accuracy ($\%$).}
\label{tab:llm_full_results_llama3}
\resizebox{\linewidth}{!}{
\begin{NiceTabular}{@{} l *{3}{w{c}{3.6em}@{\hspace{12pt}}}*{3}{w{c}{3.6em}@{\hspace{12pt}}}*{3}{w{c}{3.6em}@{\hspace{12pt}}}@{}}
\toprule
\multirow{2}{*}{Task} & \multicolumn{3}{c}{\textit{Llama3 1B}} & \multicolumn{3}{c}{\textit{Llama3 3B}} & \multicolumn{3}{c}{\textit{Llama3 8B}} \\
\cmidrule(l){2-4} \cmidrule(l){5-7} \cmidrule(l){8-10}
 & \texttt{Baseline} & \texttt{RC2@60} & $\llm$ & \texttt{Baseline} & \texttt{RC2@60} & $\llm$ & \texttt{Baseline} & \texttt{RC2@60} & $\llm$ \\
\midrule
BulgarianStoreReviewSentimentClassification & 46.3 & 51.9 & 54.6 & 51.4 & 57.5 & 58.3 & 53.1 & 58.9 & 60.0 \\
DalajClassification & 50.1 & 50.1 & 50.3 & 50.4 & 50.5 & 50.5 & 50.5 & 50.4 & 50.5 \\
DBpediaClassification & 90.3 & 91.8 & 92.9 & 91.1 & 93.1 & 93.6 & 91.3 & 93.3 & 93.8 \\
EstonianValenceClassification & 34.3 & 35.6 & 36.3 & 36.5 & 38.8 & 39.4 & 38.8 & 41.0 & 41.2 \\
FinancialPhrasebankClassification & 63.5 & 67.7 & 69.9 & 62.5 & 66.7 & 68.9 & 60.8 & 67.3 & 69.6 \\
GujaratiNewsClassification & 62.5 & 63.0 & 65.8 & 74.3 & 77.5 & 79.1 & 80.5 & 81.2 & 82.5 \\
IndonesianIdClickbaitClassification & 61.8 & 63.3 & 63.1 & 63.7 & 64.3 & 64.3 & 62.9 & 64.8 & 64.9 \\
IsiZuluNewsClassification & 23.9 & 24.0 & 24.4 & 24.5 & 25.0 & 25.1 & 24.7 & 25.6 & 25.6 \\
ItaCaseholdClassification & 56.5 & 42.0 & 53.3 & 59.0 & 52.3 & 58.7 & 60.4 & 55.0 & 61.4 \\
KorSarcasmClassification & 64.3 & 57.7 & 59.0 & 63.5 & 59.6 & 59.9 & 66.8 & 58.9 & 60.5 \\
KurdishSentimentClassification & 58.9 & 60.7 & 60.2 & 61.3 & 63.8 & 63.9 & 60.8 & 62.8 & 63.0 \\
NepaliNewsClassification & 87.9 & 89.2 & 91.0 & 89.4 & 90.8 & 92.3 & 90.8 & 92.1 & 92.9 \\
NusaX-senti & 55.2 & 58.1 & 57.6 & 58.7 & 62.0 & 61.4 & 58.5 & 63.1 & 62.0 \\
PAC & 67.9 & 67.2 & 68.7 & 69.1 & 68.5 & 68.8 & 67.7 & 67.7 & 67.7 \\
PoemSentimentClassification & 39.5 & 44.9 & 44.8 & 38.5 & 41.7 & 42.5 & 41.9 & 44.0 & 43.4 \\
PolEmo2.0-OUT & 42.0 & 47.5 & 47.9 & 43.7 & 52.8 & 53.1 & 46.0 & 56.0 & 57.2 \\
PunjabiNewsClassification & 68.5 & 67.6 & 75.5 & 71.3 & 73.6 & 76.9 & 72.2 & 76.0 & 76.8 \\
ScalaClassification & 52.8 & 52.4 & 53.3 & 54.8 & 54.5 & 54.1 & 56.0 & 55.1 & 55.2 \\
SentimentAnalysisHindi & 50.8 & 53.4 & 54.1 & 51.1 & 55.2 & 54.9 & 50.1 & 54.8 & 54.6 \\
SinhalaNewsClassification & 53.0 & 50.7 & 55.1 & 62.5 & 62.7 & 65.0 & 66.8 & 66.8 & 69.1 \\
SiswatiNewsClassification & 70.8 & 51.6 & 74.4 & 73.2 & 66.6 & 73.2 & 70.8 & 72.1 & 73.4 \\
SwahiliNewsClassification & 63.4 & 57.5 & 63.3 & 65.3 & 62.8 & 64.1 & 65.7 & 64.1 & 65.0 \\
TswanaNewsClassification & 28.3 & 19.5 & 22.9 & 32.1 & 27.6 & 30.7 & 34.6 & 33.7 & 37.5 \\
TweetTopicSingleClassification & 65.8 & 65.5 & 67.9 & 66.0 & 65.9 & 67.5 & 66.7 & 65.8 & 67.3 \\
\bottomrule
\end{NiceTabular}
}
\end{table}

\begin{table}[!h]
\centering
\caption{\textbf{Full performance comparison on Gemma3} between the probing baseline on the final-layer residual stream ($\ffnres$), residual-stream probing at $60\%$ depth ($\texttt{RC2@60}$), and the fixed strategy $\llm$. We report for each strategy the probing test accuracy ($\%$).}
\label{tab:llm_full_results_gemma3}
\resizebox{\linewidth}{!}{
\begin{NiceTabular}{@{} l *{3}{w{c}{3.6em}@{\hspace{12pt}}}*{3}{w{c}{3.6em}@{\hspace{12pt}}}*{3}{w{c}{3.6em}@{\hspace{12pt}}}@{}}
\toprule
\multirow{2}{*}{Task} & \multicolumn{3}{c}{\textit{Gemma3 1B}} & \multicolumn{3}{c}{\textit{Gemma3 4B}} & \multicolumn{3}{c}{\textit{Gemma3 12B}} \\
\cmidrule(l){2-4} \cmidrule(l){5-7} \cmidrule(l){8-10}
 & \texttt{Baseline} & \texttt{RC2@60} & $\llm$ & \texttt{Baseline} & \texttt{RC2@60} & $\llm$ & \texttt{Baseline} & \texttt{RC2@60} & $\llm$ \\
\midrule
BulgarianStoreReviewSentimentClassification & 51.1 & 58.0 & 59.0 & 54.9 & 58.4 & 61.8 & 56.4 & 58.5 & 61.5 \\
DalajClassification & 50.2 & 50.1 & 50.2 & 50.2 & 50.8 & 50.8 & 50.9 & 51.4 & 50.9 \\
DBpediaClassification & 91.7 & 87.5 & 92.6 & 92.0 & 90.7 & 92.9 & 91.9 & 91.7 & 93.3 \\
EstonianValenceClassification & 36.4 & 40.3 & 40.7 & 41.0 & 43.0 & 44.0 & 41.1 & 41.5 & 43.5 \\
FinancialPhrasebankClassification & 65.8 & 67.6 & 68.8 & 67.2 & 65.4 & 67.4 & 66.5 & 63.1 & 64.2 \\
GujaratiNewsClassification & 83.5 & 81.6 & 83.8 & 86.0 & 85.3 & 85.4 & 86.5 & 87.3 & 87.7 \\
IndonesianIdClickbaitClassification & 65.0 & 65.6 & 66.2 & 64.0 & 66.5 & 67.0 & 64.2 & 65.7 & 64.6 \\
IsiZuluNewsClassification & 23.0 & 23.8 & 24.8 & 26.5 & 25.9 & 28.5 & 28.0 & 28.3 & 30.7 \\
ItaCaseholdClassification & 51.4 & 50.4 & 58.6 & 54.3 & 49.4 & 57.6 & 57.6 & 56.6 & 64.4 \\
KorSarcasmClassification & 60.8 & 57.6 & 58.0 & 61.2 & 58.3 & 58.6 & 60.9 & 59.1 & 58.5 \\
KurdishSentimentClassification & 60.6 & 60.8 & 61.5 & 62.5 & 66.8 & 67.3 & 66.5 & 70.7 & 71.5 \\
NepaliNewsClassification & 93.3 & 90.7 & 92.0 & 94.1 & 91.9 & 92.1 & 94.0 & 93.8 & 94.3 \\
NusaX-senti & 58.4 & 58.0 & 60.6 & 58.7 & 61.6 & 63.3 & 60.7 & 65.1 & 68.0 \\
PAC & 67.8 & 69.5 & 71.4 & 68.7 & 71.2 & 70.9 & 67.6 & 69.5 & 71.0 \\
PoemSentimentClassification & 39.7 & 41.4 & 40.9 & 42.4 & 38.9 & 39.2 & 39.9 & 36.1 & 38.2 \\
PolEmo2.0-OUT & 41.6 & 49.6 & 53.1 & 46.8 & 58.2 & 59.5 & 49.9 & 56.6 & 54.8 \\
PunjabiNewsClassification & 76.6 & 74.3 & 75.2 & 76.1 & 74.9 & 76.2 & 75.8 & 75.5 & 75.0 \\
ScalaClassification & 52.6 & 54.7 & 54.6 & 54.6 & 57.1 & 56.8 & 55.5 & 58.6 & 57.9 \\
SentimentAnalysisHindi & 52.1 & 54.6 & 55.3 & 53.7 & 58.8 & 57.6 & 53.6 & 55.7 & 56.3 \\
SinhalaNewsClassification & 65.5 & 65.9 & 69.6 & 65.6 & 67.8 & 70.4 & 67.8 & 69.9 & 74.2 \\
SiswatiNewsClassification & 75.1 & 73.5 & 74.5 & 75.8 & 75.0 & 74.5 & 72.1 & 74.5 & 76.4 \\
SwahiliNewsClassification & 59.5 & 62.1 & 65.1 & 61.7 & 63.2 & 65.6 & 63.7 & 64.2 & 66.3 \\
TswanaNewsClassification & 27.9 & 22.9 & 24.6 & 33.6 & 37.0 & 42.2 & 38.7 & 43.3 & 48.3 \\
TweetTopicSingleClassification & 68.0 & 61.9 & 65.7 & 69.3 & 64.6 & 65.6 & 67.3 & 65.5 & 68.7 \\
\bottomrule
\end{NiceTabular}
}
\end{table}

\subsubsection{Comparison to embedding models}
\label[secinapp]{app:embedding_models}
We report in~\cref{tab:mteb_comparison} the performance of Gemma3 $4$B and $12$B, which are general-purpose, non-instruct models, alongside strong embedding models. We consider the $5$ best models reported in~\citet{enevoldsen2025mmteb} at the time of publication: multilingual E5 models~\citep{wang2024e5}, namely $\text{mE5}_\text{base}$, $\text{mE5}_\text{large}$, and $\text{mE5}_\text{large-instruct}$ GritLM-7B~\citep{muennighoff2025gritlm}, and $\text{E5}_\text{mistral-7b}$~\citep{wang2024e5mistral}, where the last two models are based on Mistral 7B~\citep{jiang2023mistral7b}. We also consider EmbeddingGemma~\citep{vera2025embeddinggemma}, the official embedding model built from Gemma3~\citep{kammath2025gemma3}, and Gemini Embedding~\citep{lee2025geminiembedding}, the industrial-scale closed embedding model derived from Gemini~\citep{geminiteam2024gemini15,geminiteam2025gemini}.\footnote{The exact version of Gemini used to obtain Gemini Embedding is not specified.} Performance is recovered on the $24$ classification tasks of \texttt{MTEB(Multilingual)} used in our LLM study, using the official live leaderboard (accessed at \url{https://huggingface.co/spaces/mteb/leaderboard} on September 9, 2026). Following~\citet{enevoldsen2025mmteb}, we rank models with a Borda count, where each task ranks the models, and each model scores points according to its position. Since all models are evaluated on all tasks, this is equivalent to ordering models by their average rank across tasks, which we report along with the average accuracy. With the probing baseline, Gemma3 models rank among the last and obtain the lowest average accuracy. Using $\llm$, without any finetuning, both move up the ranking: Gemma3 $12$B and $4$B reach the $3^\mathrm{rd}$ ($5.2$) and $4^\mathrm{th}$ ($5.9$) best average ranks, behind only Gemini Embedding and $\text{mE5}_\text{large-instruct}$. They notably rank ahead of GritLM-7B, the second-best model in~\citet{enevoldsen2025mmteb}, and of EmbeddingGemma, which remains slightly better in average accuracy than Gemma3 $12$B ($64.4$ vs. $64.2$).

\begin{table}[!h]
\centering
\caption{\textbf{Performance comparison with embedding models.} We compare Gemma3 $4$B and $12$B, probed on the final-layer residual stream (\texttt{Baseline}) or with the fixed strategy $\llm$, to the top-$5$ models of~\citet{enevoldsen2025mmteb}, EmbeddingGemma~\citep{vera2025embeddinggemma} and Gemini Embedding~\citep{lee2025geminiembedding}. Results for embedding models are taken from the \texttt{MTEB(Multilingual)} leaderboard at \url{https://huggingface.co/spaces/mteb/leaderboard}. We report the test accuracy ($\%$) on each task, the average accuracy, and the average rank across tasks (lower is better). Following \texttt{MMTEB}, models are ordered by average rank, which yields the same ordering as the Borda count used by the leaderboard since all models are evaluated on all tasks.}
\label{tab:mteb_comparison}
\resizebox{\linewidth}{!}{
\begin{NiceTabular}{@{} l *{11}{w{c}{8.2em}@{\hspace{4pt}}} @{}}
\toprule
\multirow{2}{*}{Task} & \multirow{2}{*}{\textit{Gemini Embedding}} & \multirow{2}{*}{$\textit{mE5}_\textit{large-instruct}$} & \textit{Gemma3 12B} & \textit{Gemma3 4B} & \multirow{2}{*}{\textit{EmbeddingGemma}} & \multirow{2}{*}{\textit{GritLM-7B}} & \multirow{2}{*}{$\textit{mE5}_\textit{large}$} & \multirow{2}{*}{$\textit{E5}_\textit{mistral-7b}$} & \textit{Gemma3 12B} & \multirow{2}{*}{$\textit{mE5}_\textit{base}$} & \textit{Gemma3 4B} \\
 &  &  & $\llm$ & $\llm$ &  & &  & & \texttt{Baseline} & & \texttt{Baseline} \\
\midrule
BulgarianStoreReviewSentimentClassification & 78.1 & 78.6 & 61.5 & 61.8 & 71.3 & 77.6 & 63.9 & 79.3 & 56.4 & 69.4 & 54.9 \\
DalajClassification & 50.5 & 50.0 & 50.9 & 50.8 & 50.3 & 50.6 & 50.0 & 50.9 & 50.9 & 49.8 & 50.2 \\
DBpediaClassification & 94.8 & 95.5 & 93.3 & 92.9 & 94.3 & 92.9 & 88.3 & 90.9 & 91.9 & 87.1 & 92.0 \\
EstonianValenceClassification & 53.5 & 49.1 & 43.5 & 44.0 & 38.3 & 38.7 & 42.9 & 42.3 & 41.1 & 41.2 & 41.0 \\
FinancialPhrasebankClassification & 88.6 & 84.4 & 64.2 & 67.4 & 86.5 & 84.4 & 83.9 & 79.2 & 66.5 & 82.1 & 67.2 \\
GujaratiNewsClassification & 92.0 & 87.5 & 87.7 & 85.4 & 82.8 & 69.9 & 76.7 & 73.1 & 86.5 & 74.9 & 86.0 \\
IndonesianIdClickbaitClassification & 67.0 & 61.3 & 64.6 & 67.0 & 60.9 & 60.4 & 61.2 & 57.9 & 64.2 & 60.1 & 64.0 \\
IsiZuluNewsClassification & 40.5 & 37.3 & 30.7 & 28.5 & 26.4 & 28.2 & 33.1 & 25.0 & 28.0 & 31.5 & 26.5 \\
ItaCaseholdClassification & 73.3 & 62.9 & 64.4 & 57.6 & 70.4 & 73.5 & 66.8 & 67.9 & 57.6 & 61.5 & 54.3 \\
KorSarcasmClassification & 60.5 & 58.2 & 58.5 & 58.6 & 58.1 & 59.7 & 56.8 & 57.0 & 60.9 & 54.7 & 61.2 \\
KurdishSentimentClassification & 86.4 & 80.9 & 71.5 & 67.3 & 60.0 & 67.2 & 77.1 & 69.1 & 66.5 & 76.4 & 62.5 \\
NepaliNewsClassification & 98.1 & 97.0 & 94.3 & 92.1 & 95.5 & 94.4 & 88.5 & 95.0 & 94.0 & 89.4 & 94.1 \\
NusaX-senti & 80.3 & 71.0 & 68.0 & 63.3 & 69.8 & 70.8 & 70.5 & 70.9 & 60.7 & 68.8 & 58.7 \\
PAC & 71.7 & 65.7 & 71.0 & 70.9 & 67.9 & 67.0 & 70.3 & 62.3 & 67.6 & 70.9 & 68.7 \\
PoemSentimentClassification & 59.7 & 57.1 & 38.2 & 39.2 & 58.9 & 51.6 & 50.7 & 59.1 & 39.9 & 46.6 & 42.4 \\
PolEmo2.0-OUT & 77.5 & 57.0 & 54.8 & 59.5 & 62.8 & 62.5 & 36.5 & 60.5 & 49.9 & 43.9 & 46.8 \\
PunjabiNewsClassification & 82.6 & 84.4 & 75.0 & 76.2 & 82.4 & 60.2 & 80.7 & 55.7 & 75.8 & 80.7 & 76.1 \\
ScalaClassification & 51.9 & 50.4 & 57.9 & 56.8 & 50.8 & 52.1 & 51.1 & 51.1 & 55.5 & 50.9 & 54.6 \\
SentimentAnalysisHindi & 76.1 & 59.5 & 56.3 & 57.6 & 65.5 & 57.1 & 64.2 & 62.1 & 53.6 & 64.6 & 53.7 \\
SinhalaNewsClassification & 82.3 & 78.0 & 74.2 & 70.4 & 65.7 & 58.0 & 66.8 & 56.3 & 67.8 & 65.8 & 65.6 \\
SiswatiNewsClassification & 62.4 & 47.5 & 76.4 & 74.5 & 57.0 & 58.9 & 54.4 & 43.1 & 72.1 & 47.0 & 75.8 \\
SwahiliNewsClassification & 66.0 & 59.1 & 66.3 & 65.6 & 66.0 & 63.0 & 59.8 & 58.3 & 63.7 & 62.6 & 61.7 \\
TswanaNewsClassification & 53.4 & 46.4 & 48.3 & 42.2 & 31.1 & 40.1 & 46.2 & 40.6 & 38.7 & 41.0 & 33.6 \\
TweetTopicSingleClassification & 71.1 & 74.9 & 68.7 & 65.6 & 73.0 & 75.0 & 65.3 & 76.6 & 67.3 & 67.4 & 69.3 \\
\midrule
Average accuracy & 71.6 & 66.4 & 64.2 & 63.1 & 64.4 & 63.1 & 62.7 & 61.8 & 61.5 & 62.0 & 60.9 \\
Average rank & 2.1 & 4.8 & 5.2 & 5.9 & 6.0 & 6.2 & 6.9 & 6.9 & 7.1 & 7.5 & 7.6 \\
\bottomrule
\end{NiceTabular}
}
\end{table}

\end{document}